\documentclass{article} 
\usepackage{iclr2025_conference,times}
\usepackage{etoc}  

\usepackage{amsmath,amsfonts,bm}

\def\eqref#1{equation~\ref{#1}}

\def\1{\bm{1}}

\DeclareMathAlphabet{\mathsfit}{\encodingdefault}{\sfdefault}{m}{sl}
\SetMathAlphabet{\mathsfit}{bold}{\encodingdefault}{\sfdefault}{bx}{n}

\usepackage{tabularx} 
\usepackage{adjustbox} 

\usepackage[normalem]{ulem}
\useunder{\uline}{\ul}{}

\usepackage{etoc}
\usepackage{titletoc}

\newcommand\DoToC{%
  \startcontents
  \printcontents{}{1}{\noindent \textbf{\Large{Table of Contents in Appendix}}\vskip3pt\vskip5pt}
  \vskip3pt\vskip5pt
}

\usepackage{hyperref}
\usepackage{url}
\usepackage{xspace}
\usepackage{hyperref}
\usepackage{url}
\usepackage{xspace}
\usepackage{booktabs}
\usepackage{graphicx}
\usepackage{MnSymbol,bbding,pifont}
\usepackage{colortbl}
\definecolor{LightCyan}{rgb}{0.88,1,1}
\usepackage{enumitem}
\usepackage{multirow}
\usepackage{xcolor}
\usepackage{wrapfig}

\definecolor{Pre-Algebra}{HTML}{F27970
}
\definecolor{Inter-Algebra}{HTML}{BB9727}
\definecolor{Algebra}{HTML}{54B345}
\definecolor{Probability}{HTML}{32B897}
\definecolor{NumTheory}{HTML}{05B9E2}
\definecolor{Calculus}{HTML}{8983BF}
\definecolor{Geometry}{HTML}{C76DA2}

\definecolor{darkblue}{RGB}{84, 112, 198}
\definecolor{lightgreen}{RGB}{145, 204, 117}
\definecolor{lightyellow}{RGB}{250, 200, 88}
\definecolor{lightred}{RGB}{238, 102, 102}
\definecolor{lightblue}{RGB}{115, 192, 222}

\usepackage{tcolorbox}
\usepackage{color-edits}
\addauthor{gn}{magenta}
\usepackage{pifont} 

\usepackage{fontawesome} 

\newtcolorbox{promptbox}[2][Prompt]{
colback=black!5!white,
arc=5pt, 
boxrule=0.5pt,
fonttitle=\bfseries,
title=#1, 
before upper={\small}, fontupper=\fontfamily{ptm}\selectfont,
colframe=#2,
}

\title{{\includegraphics[width=0.06\textwidth]{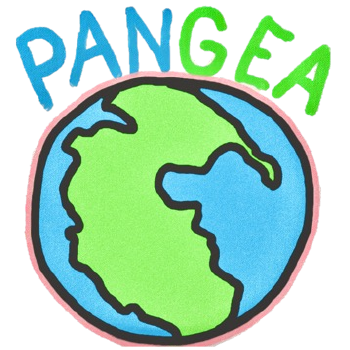}}Pangea: A Fully Open Multilingual Multimodal LLM for 39 Languages}

\newcommand{\model}{\textsc{Pangea}\xspace}
\newcommand{\traindata}{\textsc{PangeaIns}\xspace}
\newcommand{\evaldata}{\textsc{PangeaBench}\xspace}

\iclrfinalcopy 
\author{Xiang Yue\thanks{Equal Contributions.},\; Yueqi Song\footnotemark[1],\; Akari Asai, \\\textbf{Seungone Kim, Jean de Dieu Nyandwi, Simran Khanuja, Anjali Kantharuban, }\\
\textbf{Lintang Sutawika, Sathyanarayanan Ramamoorthy, Graham Neubig}\\
\texttt{\{xyue2,yueqis,gneubig\}@cs.cmu.edu}
\\[1em]
\makebox[\textwidth]{\fontsize{11}{11}\selectfont Carnegie Mellon University}
}

\begin{document}

\maketitle

\vspace{-1.0cm}
\begin{center}
    \url{https://neulab.github.io/Pangea/}
\end{center}
\vspace{0cm}

\begin{abstract}
Despite recent advances in multimodal large language models (MLLMs), their development has predominantly focused on English- and western-centric datasets and tasks, leaving most of the world's languages and diverse cultural contexts underrepresented.  
This paper introduces \model, a multilingual multimodal LLM trained on \traindata, a diverse 6M instruction dataset spanning 39 languages. \traindata features: 1) high-quality English instructions, 2) carefully machine-translated instructions, and 3) culturally relevant multimodal tasks to ensure cross-cultural coverage. 
To rigorously assess models' capabilities, we introduce \evaldata, a holistic evaluation suite encompassing 14 datasets covering 47 languages. 
Results show that \model significantly outperforms existing open-source models in multilingual settings and diverse cultural contexts. Ablation studies further reveal the importance of English data proportions, language popularity, and the number of multimodal training samples on overall performance.  We fully open-source our data, code, and trained checkpoints, to facilitate the development of inclusive and robust multilingual MLLMs, promoting equity and accessibility across a broader linguistic and cultural spectrum.
\end{abstract}

\begin{figure}[ht]
  \centering
    \vspace{-5pt}
\includegraphics[width=\textwidth]{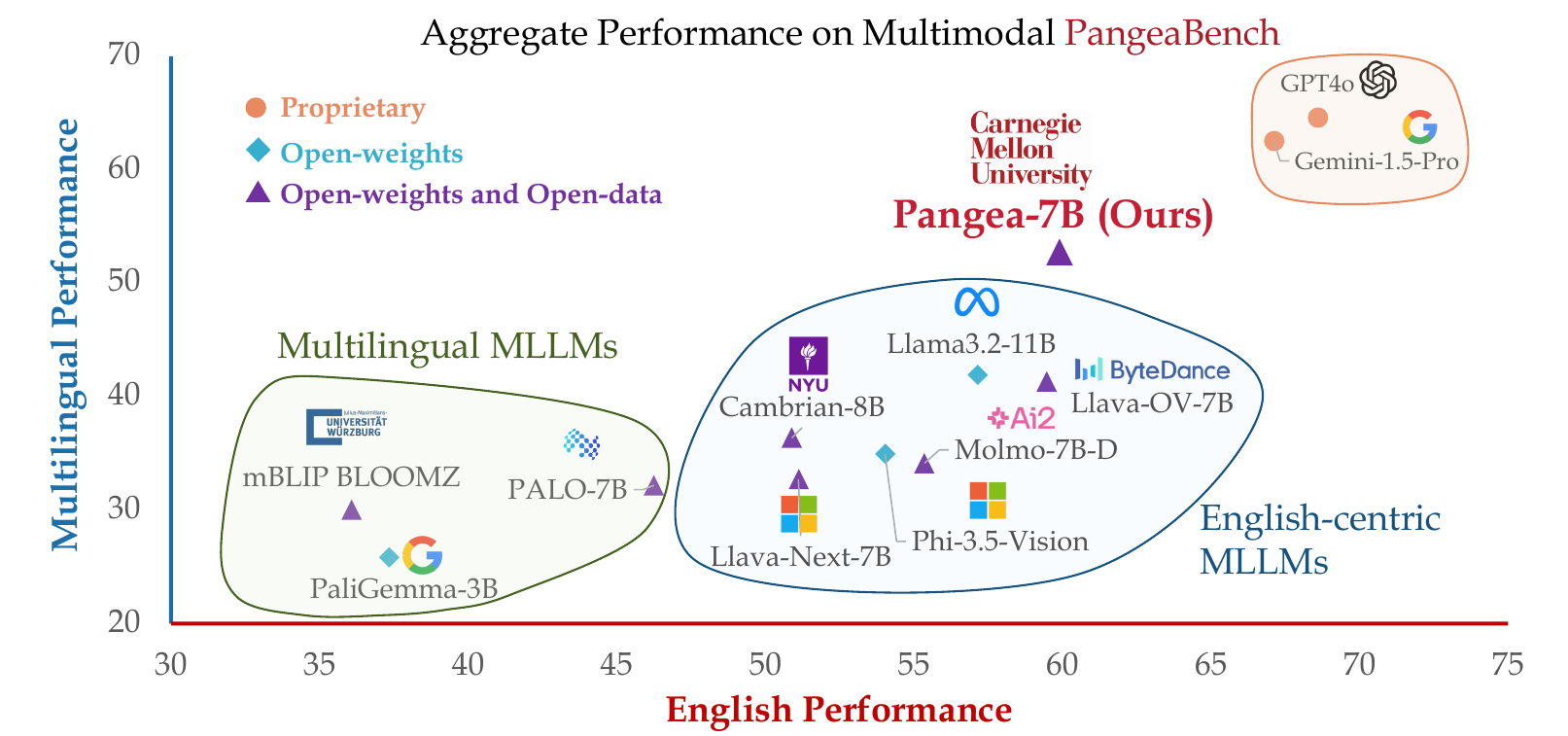}
\caption{Overview of the aggregate performance of various multimodal LLMs on \evaldata. Our \model-7B demonstrates comparable performance to SoTA open-source models in English settings, while significantly outperforming them in multilingual scenarios.}
  \label{fig:teaser}
  \vspace{-10pt}
\end{figure}

\section{Introduction}
Multimodal large language models (MLLMs)~\citep{liu2023llava,liu2024llavanext,dubey2024llama,deitke2024molmo,gpt4o,deepmind_gemini_report} have demonstrated impressive capabilities in tasks such as image captioning, visual question answering, and complex reasoning~\citep{yue2024mmmu,yue2024mmmupro}. 
Despite this rapid progress in their reasoning ability, a critical flaw persists: \emph{the overwhelming focus on English- and western-centric training and evaluation datasets} \citep{liu2021visually,song-etal-2023-globalbench}. 

This homogeneity results in a lack of representation for the vast majority of the world's languages and diverse cultural contexts~\citep{yu-etal-2022-beyond}. 
Consequently, models predominately trained on such data exhibit: (a) diminished performance in multilingual settings~\citep{blasi-etal-2022-systematic} with poor tokenization further leading to higher inference costs~\citep{ahia-etal-2023-languages}; (b) generate outputs misaligned with the socio-cultural norms of underrepresented languages~\citep{alkhamissi-etal-2024-investigating}; and (c) lack the ability to recognize objects from geographically diverse regions~\citep{ramaswamy2024geode} or rare objects belonging to the long-tail~\citep{gupta2019lvis}.
With the increased adoption of these models into real-world applications across the globe, there's an urgent need to develop multilingual MLLMs that equitably serve a diverse set of users. Few efforts have been made to develop multilingual MLLMs~\citep{geigle_etal_2024_mblip, PALO}, however, their performance still exhibits inequalities across languages and lacks evaluation of cultural understanding.

In this paper, we address how to train and evaluate culturally inclusive multilingual MLLMs, using limited open-source resources, tackling four major challenges~\citep{yu-etal-2022-beyond}: {\bf 1) Data scarcity:} high-quality multilingual multimodal data is scarce, especially in low-resource languages, making it difficult to create large-scale training data; {\bf 2) Cultural nuances:} visual interpretations are context-dependent and vary across cultures~\citep{NEURIPS2023_d08b6801,khanuja2024image}; {\bf 3) Catastrophic forgetting:} training on many languages or modalities often results in suboptimal performance on some subsets and require careful balancing; {\bf 4) Evaluation complexity:} substantial resources and expertise are required to accurately measure performance across languages and cultures. 

\begin{figure}[!t]
    \centering
    \includegraphics[width=\linewidth]{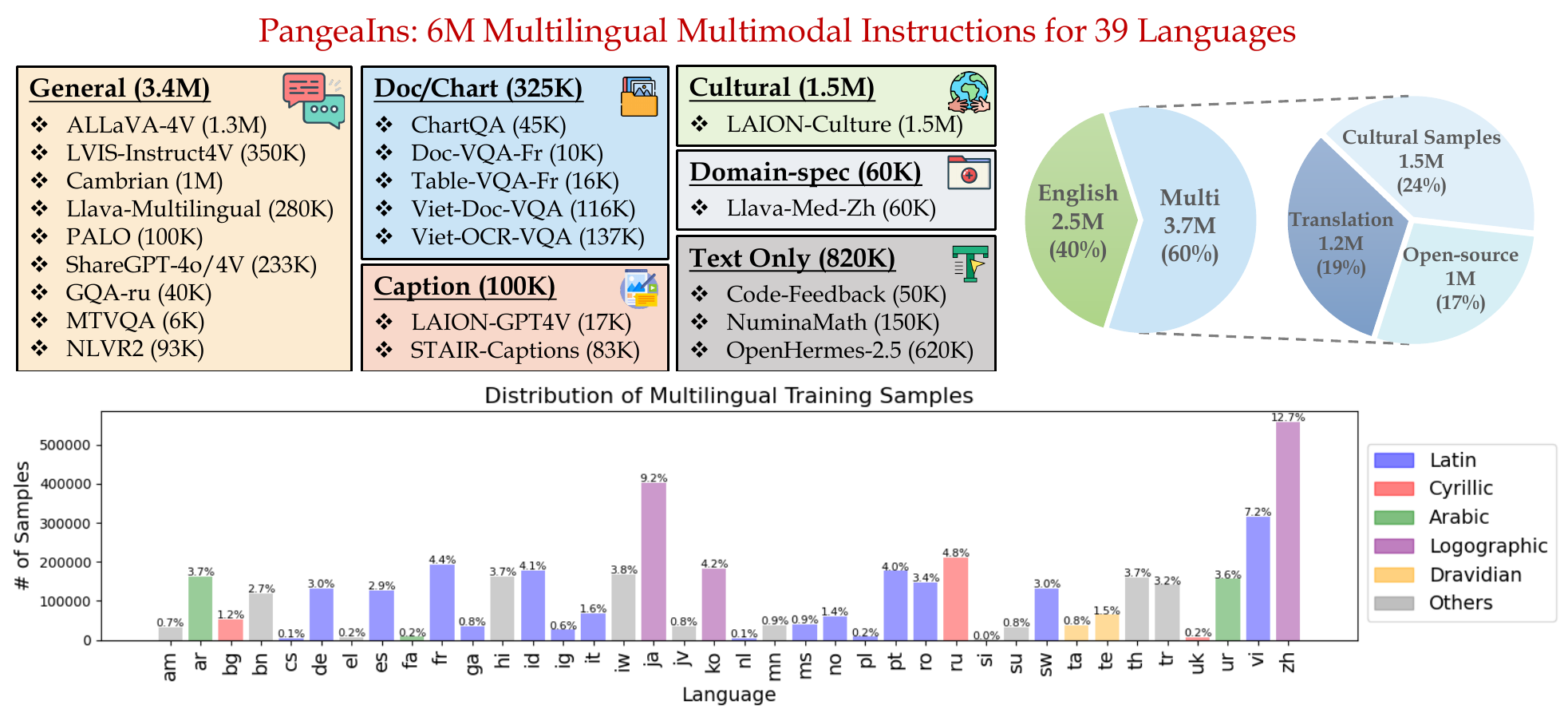}
    \vspace{-20pt}
\caption{Statistics of \traindata, comprising 6M multimodal instructions in 39 languages. The distribution of multilingual training data shows the percent of instances for each language among the \textit{multilingual} instructions. \traindata includes general instructions, document and chart question answering, captioning, domain-specific, culturally relevant, and text-only instructions. 
}
    \vspace{-10pt}

    \label{fig:train_data_distribution}
\end{figure}

To tackle these challenges, we introduce \model, an open-source multilingual MLLM designed to bridge linguistic and cultural gaps in visual understanding tasks. \model is trained on \traindata (\autoref{fig:train_data_distribution}), a high-quality multilingual multimodal instruction tuning dataset comprising 6 million samples in 39 typologically diverse languages.
\traindata combines existing open-source resources with newly created instructions focused on multicultural understanding. We curate high-quality English instructions, carefully translate and adapt them for multilingual contexts.  To address Western-centric biases in visual representations, we source images from LAION-Multi~\citep{schuhmann2022laion}, which includes images from various countries and captions in multiple languages. However, LAION-Multi contains images that are not culturally representative of the country's speaking population, and the associated alt text is often short, noisy, and lacks sufficient detail. To combat these issues, we develop a multicultural multilingual multimodal instruction generation pipeline, leveraging an LLM~\citep{dubey2024llama} to score and filter images based on cultural informativeness. We then enhance the remaining data by generating detailed descriptions and creating complex instructions that combine culturally relevant tasks with general multilingual scenarios. This approach improves the model's cultural understanding while maintaining robust multilingual performance.

To evaluate \model's capabilities, we present \evaldata, a comprehensive multilingual and multimodal evaluation suite comprising five multimodal and three text-based tasks across 14 datasets in 47 languages. \evaldata assesses MLLMs' performance on open-domain multimodal chat, image captioning, cultural understanding, multimodal reasoning, and text-only tasks including question answering and complex math reasoning. 
A key highlight of \evaldata is the introduction of xChat, a human-crafted benchmark designed to evaluate open-ended, information-seeking multimodal conversations. xChat employs a fine-grained evaluation pipeline where human annotators annotate both reference answers and scoring rubrics for each query. An LLM then uses these rubrics to score the model's predictions on a 1-5 scale. This approach offers a more precise assessment of MLLM performance, addressing limitations of coarse LLM-as-Judge methods~\citep{zheng2023judging}. Additionally, we introduce xMMMU, a translated version of MMMU~\citep{yue2024mmmu},  testing college-level multimodal reasoning across seven languages. Together, these components provide a robust, nuanced evaluation of \model's cross-lingual and cross-cultural capabilities.

Our results demonstrate \model's abilities in both English and multilingual scenarios, significantly outperforming existing open-source MLLMs on \evaldata, surpassing the best open MLLMs by 0.4\% on English tasks and 10.9\% on multilingual tasks on average. Notably, \model excels in multilingual and multicultural understanding, evidenced by its performance on xChat, CVQA, and MaRVL benchmarks. \model also matches or outperforms state-of-the-art proprietary LLMs, namely Gemini-1.5-Pro and GPT4o, on several tasks such as XGQA. However, some performance gaps remain in multimodal chat and complex reasoning, shedding light on the need for further improvements in open MLLMs. We discuss key insights from  training\model, including the scaling effect of instructions, the role of English data, the impact of language-specific training proportions, and preliminary methods to improve multilingual OCR. We fully open-source \traindata, \evaldata, \model-7B, and code, to advance culturally inclusive MLLMs across diverse languages.

\section{\traindata: Multilingual Multimodal Instruction Tuning}
Creating a truly multilingual, multicultural MLLM presents unique challenges. 
We developed \traindata, a diverse and high-quality instruction tuning dataset. Comprising 6 million samples in 39 languages, \traindata was curated with a focus on linguistic and cultural diversity. 
We implemented three key strategies to ensure comprehensive coverage, each addressing the specific hurdles encountered in multilingual multimodal learning. \autoref{fig:train_data_distribution} shows the distribution of \traindata. 

\subsection{Machine Translated Instructions}
\label{sec:machine_translation}
To address the scarcity of human-annotated multilingual multimodal data, we primarily adopt machine translation as a practical and scalable solution to extend data beyond English. While human annotation is ideal, it is resource-intensive and impractical to cover a wide range of languages. 

\textbf{Constructing a High-quality Pool of English Instructions from Existing Sources.}
We first collect a high-quality set of English multimodal instructions, which serve as the foundation for translation into other languages. These instructions span a wide range of visual understanding tasks, including general visual instructions and conversations~\citep{tong2024cambrian,liu2024llavanext}, visual reasoning, captioning, and chart question answering~\citep{masry2022chartqa}. Besides, we also added text-only high-quality English instructions, covering general instructions~\citep{OpenHermes25}, code~\citep{zheng2024opencodeinterpreter}, and math~\citep{numina_math_datasets}.  \autoref{fig:train_data_distribution} shows the statistics of our translated datasets. By leveraging existing  English instructions, we ensured comprehensive coverage of visual interpretation and text instruction following tasks in English, preparing a pool of high-quality data for translation.

\textbf{Translation Model Selection.} 
To expand the English instructions to other languages, we initially experimented with strong open-source machine translation models, such as NLLB-3B~\citep{nllb2024scaling}. However, we found that these models struggled with complex instruction-following scenarios and context-switching tasks, particularly in specialized domains like code generation and mathematical reasoning. For example, in code-related tasks, the model failed to recognize and correctly translate programming language keywords, significantly reducing the quality of the instructions. Based on these limitations, we shifted to using the proprietary Gemini 1.5 Pro model, which shows slightly better performance in small-scale human evaluations compared with GPT4o. 

\noindent\textbf{Post-Processing Translated Data.} Even with high-quality translations, inconsistencies arose. To resolve issues such as mismatched conversation turns or missing candidates in multiple-choice questions, we developed a post-processing pipeline. This pipeline automatically corrected these errors or directly dropped the examples, ensuring that all translated instructions remained consistent.

Overall, Gemini 1.5 Pro's translation seems satisfactory, providing a fast, cost-effective alternative to human annotation, especially for scaling across languages. However, we acknowledge that machine translation still has limitations, particularly in handling nuanced contexts and cultural subtleties.

\begin{figure}
    \centering
    \includegraphics[width=1.0\linewidth]{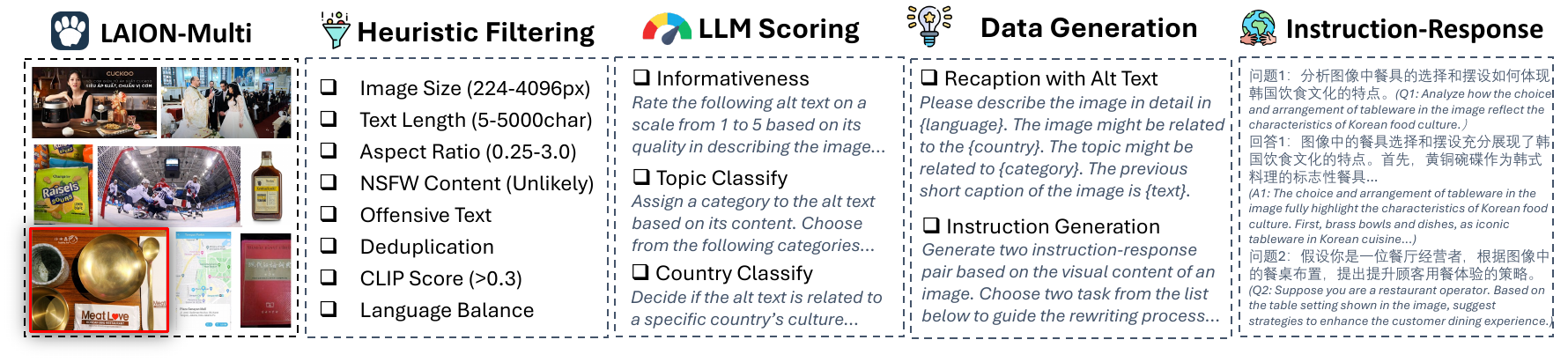}
    \caption{Overview of multicultural understanding instructions data generation pipeline.}
    \label{fig:cultural_understanding_pipeline}
    \vspace{-15pt}
\end{figure}

\subsection{Multicultural Understanding Instructions}
\label{sec:multicultural_data}
While machine translation enables scaling across multiple languages, data translated from English is still Anglo-centric in coverage of cultural concepts~\citep{yu-etal-2022-beyond}. To address this, we developed a pipeline focused on creating instructions for multicultural understanding. 
Both visual and textual elements can convey deep cultural significance, and our goal is to design a dataset that allows models to not only recognize these nuances but also respond appropriately across various cultural contexts. The pipeline of creating multicultural understanding instructions is shown in \autoref{fig:cultural_understanding_pipeline}.

\noindent\textbf{Curation of Culturally Diverse Images.} To ensure that our dataset captures a wide array of cultural contexts, we began by sampling 10 million images from the LAION-Multi dataset~\citep{schuhmann2022laion}, which includes images and short alt texts from diverse languages and regions. A filtering process was proposed to guarantee both the quality and cultural relevance of the images.
\begin{itemize}[leftmargin=*]
    \item \textit{Heuristic Filtering:} We implemented automatic filtering based on several key criteria: Image Size, Aspect Ratio, Text Length, NSFW content, Offensive Text, Deduplication, and CLIP Score (used to assess the alignment between the image and its textual description). This helped remove low-quality or inappropriate images and ensured the remaining dataset adhered to quality standards.
    \item \textit{LLM Scoring:} To further refine the dataset, we employed the Llama-3.1-8B-Instruct model~\citep{dubey2024llama} to evaluate the quality, subjects, and cultural relevance of the accompanying text descriptions (alt text) for each image. The model was instructed to perform the following tasks: 1) \textbf{Evaluate Text Quality:} The alt text was rated on a scale from 1 to 5 based on how well it described the corresponding image, assuming the model could not access the image itself. Alt text scoring below 4 was removed. 2) \textbf{Subject Classification:} The model assigned a subject or category to the alt text based on its content. 3) \textbf{Country/Region Classification:} The model determined whether the alt text was closely related to a specific country’s culture. Images classified as “no specific country” (approximately 60\% of the dataset) were excluded to ensure we focused on culturally identifiable content. The full LLM scoring prompt is included in Appendix \ref{llm_scoring}.
    \item \textit{Avoiding Overrepresentation:} To maintain a balanced representation, we downsampled images from frequently occurring subjects, such as objects, materials, and clothing, to avoid skewing the dataset toward specific topics or regions. Then, we conducted an accessibility check, removing 30\% of the remaining samples due to image download or other issues. Ultimately, we curated a final set of 1M high-quality, culturally specific images, forming the foundation of \traindata.
\end{itemize}


\noindent\textbf{Captioning Images with Different Languages.} To provide context and enhance the model's ability to interpret images, we regenerated more detailed captions using Gemini 1.5 Pro based on high-quality alt texts. In this step, each image was accompanied by a caption written in the language corresponding to its cultural origin. However, our approach was not just about using a capable model. The alt text played a critical role in enriching the data, as it often contained culturally specific and contextually important information that would otherwise be absent from the images alone. For example, in \autoref{fig:caption_example}, with high-quality alt text, models can incorporate details such as \textit{``President and CEO of The Walt Disney Company''} and \textit{``a model of Shanghai Disneyland,''} adding significant context that may not be immediately evident from the image. This additional layer of information helps the model generate captions that better capture the cultural and contextual nuances. 

\noindent\textbf{Generating Multilingual and Cross-Cultural Instructions.} 

After recaptioning, we generated multilingual instructions based on the detailed captions with Gemini 1.5 Pro. Instead of only prompting the model to generate random instructions, we did a careful prompt engineering where we first came up with 13 task types (e.g., Information Seeking, Coding \& Debugging, Critical Reasoning, Cultural Interpretation, etc.). Then for each image, up to two QA pairs were created, representing different instruction types to ensure a diverse set of interactions. This approach ensures that the model not only recognizes these visual elements but also responds appropriately across varied linguistic and different instruction contexts. The captioning and instruction generation prompts are included in \autoref{llm_scoring}.

\subsection{Curating Existing Multilingual Instructions
}
To further enrich \traindata, we conducted an extensive survey of available multilingual multimodal literature and datasets, including those hosted on HuggingFace. As a result, we incorporated several high-quality, open-source datasets into \traindata. These include Chinese ALLaVA-4V~\citep{chen2024allava}, Viet Document and OCR QA~\citep{doan2024vintern}, Llava Chinese~\citep{ChineseLLaVA}, Llava Medical Chinese Instruction~\citep{ChineseLLaVA_Med}, LLaVA-Japanese-Instruct~\citep{LLaVA_JP_Instruct_108K}, MTVQA~\citep{tang2024mtvqa}, Japanese STAIR Captions~\citep{yoshikawa2017stair}, Russian GQA~\citep{deepvk2024gqa_ru}, French Doc-VQA~\citep{SoSoDocvqa}, and French Table-VQA~\citep{AgDeTQA}. Each of these datasets brings unique linguistic and cultural perspectives to the mix, covering a wide range of languages and task types. 

\subsection{Dataset Statistics}
By combining these three methods, we created \traindata, a comprehensive dataset addressing major challenges in building multilingual MLLMs: data scarcity, linguistic diversity, and cultural nuance. Its balanced language and task distribution supports the development of more sophisticated LLMs that can handle complex visual and textual content in a multilingual, multicultural context.

\textbf{Language and Task Distribution:} \traindata features an extensive and balanced distribution of languages, tasks, and cultural contexts (as shown in \autoref{fig:train_data_distribution}). We empirically keep the final language ratio of English to Multilingual as 40\%:60\% as we found a significant portion of English data plays an important role in cross-lingual transfer. See more discussions about the ratio in \autoref{sec:discussion} and \autoref{fig:english_ratio}. The inclusion of diverse multimodal instructions ensures that the model develops a deeper understanding of varied linguistic and cultural environments. Examples of training samples from different languages and categories are provided in \autoref{sec:train_examples}.
The comprehensive nature of \traindata lays a solid foundation for training \model, enabling it to become a truly multilingual, multicultural multimodal LLM, capable of understanding and interacting effectively with users from diverse linguistic and cultural backgrounds.

\begin{figure}[!t]
    \centering
    \includegraphics[width=\linewidth]{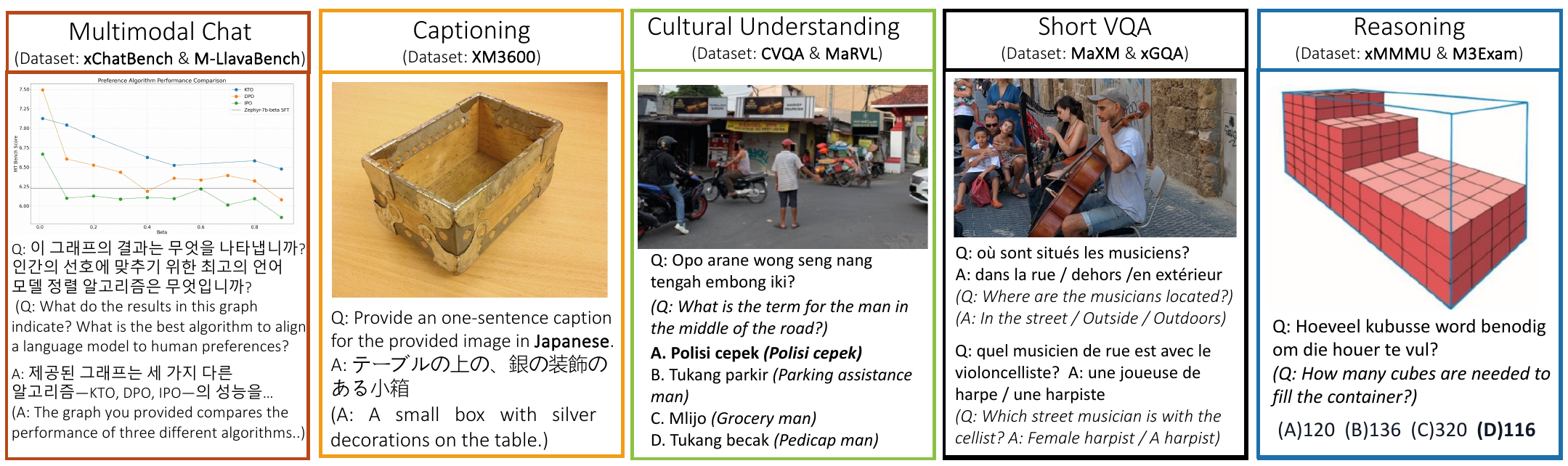}
    \resizebox{\columnwidth}{!}{%
    \begin{tabular}{@{}lllllll@{}}
    \toprule
    Category & Tasks & Datasets & Forms & Size & Languages & Metric \\ \midrule
    \multirow{9}{*}{Multimodal} & \multirow{2}{*}{\begin{tabular}[c]{@{}l@{}}Multimodal  Chat\end{tabular}} & xChatBench & Long & 400 & zh,en,hi,id,ja,rw,ko,es & LLM-as-Judge\\
     &  & M-LlavaBench & Long & 600 & ar,bn,zh,fr,hi,ja,ru,es,ur,en & LLM-as-Judge \\ \cmidrule(l){2-7} 
     & Captioning & XM100 & Long & 3.6K & 36 languages & ROUGE-L\\ \cmidrule(l){2-7} 
     & \multirow{2}{*}{\begin{tabular}[c]{@{}l@{}}Cultural \\ Understanding\end{tabular}} & CVQA & MC & 21K & en,zh,ko,mn,ja,id,jv,min,su & Accuracy \\
     &  & MaRVL & Short & 6K & id,sw,ta,tr,zh & Accuracy\\ \cmidrule(l){2-7} 
     & \multirow{2}{*}{\begin{tabular}[c]{@{}l@{}}Multilingual\\ VQA\end{tabular}} & xGQA & Short & 77K & en,de,pt,ru,id,bn,ko,zh  & Accuracy\\
     &  & MaXM & MC & 2K & hi,th,zh,fr,en,iw,ro  & Accuracy\\ \cmidrule(l){2-7} 
     & \multirow{2}{*}{\begin{tabular}[c]{@{}l@{}}Reasoning\\ (Multi-subject)\end{tabular}} & xMMMU & Short/MC & 3K & en,ar,fr,hi,id,ja,pt  & Accuracy\\
     &  & M3Exam & MC & 3K & en,zh,it,pt,vi,th,af  & Accuracy\\ \midrule
    \multirow{5}{*}{Text-only} & QA & TyDiQA & Short & 5.1K & ar,ru,bn,te,fi,sw,ko,id,en  & Accuracy\\ \cmidrule(l){2-7} 
     & Translation & FLORES-Sub & Long & 18K & ar,en,fr,de,hi,id,iw,ja,pt,ro,tr & ChrF\\ \cmidrule(l){2-7} 
     & \multirow{3}{*}{\begin{tabular}[c]{@{}l@{}}Reasoning\\ (Multi-subject,\\ Commonsense, Math)\end{tabular}} & MMMLU & MC & 197K & ar,bn,de,es,fr,hi,id,it,ja,ko,pt,sw,yo,zh & Accuracy\\
     &  & XStoryCloze & MC & 21K & en,ar,es,eu,hi,id,my,ru,sw,te,zh & Accuracy\\
     &  & MGSM & Open & 3K & bn,de,en,es,fr,ja,ru,sw,te,th,zh & Accuracy\\ \bottomrule
    \end{tabular}%
    }
    
    \caption{Overview of \evaldata, which contains 5 multimodal and 3 text tasks covering 14 datasets (including two newly curated xChatBench and xMMMU datasets). The table provides details about the datasets, while the figure shows evaluation examples from five different multimodal eval tasks in our \evaldata. }
    \vspace{-5pt}
    \label{fig:eval_data}
\end{figure}

\section{\evaldata: Evaluation of Multilingual Multimodal Models}

\subsection{Overview of \evaldata} To assess the capabilities of \model across a variety of languages, cultures, and task types, we have developed \evaldata, a comprehensive multilingual and multimodal evaluation suite. \evaldata integrates diverse benchmarks that encompass both multimodal and text-only tasks, enabling a holistic evaluation of \model's performance in cross-lingual, cross-cultural, and multimodal contexts. Each task within \evaldata is designed to probe specific aspects of \model's proficiency, ensuring robust testing across a wide range of scenarios. All tasks in \evaldata are evaluated under a zero-shot setting.

\subsection{Multimodal Tasks} The multimodal tasks in \evaldata are categorized as follows: Multimodal Chat, Captioning, Cultural Understanding, Multilingual Visual Question Answering (VQA), and Multi-Subject Reasoning. We incorporate these in \evaldata to ensure comprehensive testing of \model's multimodal capabilities. The overview and examples of \evaldata are shown in \autoref{fig:eval_data}.

\noindent\textbf{Multimodal Chat.} This task tests the model's ability to engage in natural and dynamic real-world conversations involving both text and images. Multilingual LlavaBench~\citep{PALO} (M-LlavaBench for short) stands as the only benchmark for evaluating multilingual long-form generation capabilities from MLLMs. Following the evaluation pipeline from \citet{zheng2023judging} and \citet{liu2023llava}, M-LlavaBench uses a coarse-grained evaluation criteria (e.g., ``Please rate the helpfulness, relevance, accuracy, level of details of their responses.''). Previous works suggest that employing such coarse-grained evaluation criteria may lead to automatic evaluation results that diverge from how humans would evaluate them~\citep{ye2023flask,kim2023prometheus,lee2024prometheusvision,kim2024biggen,kim2024prometheus}. To assess baselines with a more accurate evaluation pipeline with \textit{fine-grained evaluation criteria} on diverse scenarios, we additionally annotate a new multilingual multimodal generation benchmark called the \textbf{xChatBench}, included in the multimodal chat category of \evaldata. A more detailed explanation of the annotation process of xChatBench is included in~\autoref{appendix:xchat}. 

\noindent\textbf{Captioning.}  The XM3600~\citep{thapliyal2022crossmodal} dataset was developed to evaluate models' capability in multilingual image captioning. It contains images paired with captions in 36 different languages. However, it includes many similar images and captions. To address this, we clustered the images based on captions and manually selected 100 representative images (denoted as XM100). This approach enhances the diversity of the samples and accelerates the evaluation process. 

\noindent\textbf{Cultural Understanding.} To assess the model's ability to reason about and understand culturally diverse visual content, we use the CVQA~\citep{romero2024cvqa} and MaRVL~\citep{liu2021visually} datasets. These datasets are designed to test the model's performance in reasoning tasks involving culturally relevant imagery and concepts across multiple languages.

\noindent\textbf{Multilingual VQA.} This task measures the model's proficiency in answering questions about images across multiple languages. The xGQA~\citep{pfeiffer2022xgqa} and MaXM~\citep{changpinyo2022maxm} datasets provide a diverse range of visual question-answering challenges in several languages and scripts, addressing cross-lingual visual understanding.

\noindent\textbf{Multi-Subject Reasoning.} The xMMMU and M3Exam~\citep{zhang2023m3exam} datasets are used to evaluate the model's reasoning abilities across different academic subjects. xMMMU is a machine-translated version of MMMU validation questions, which focuses on multimodal reasoning in multiple subjects. We randomly sample 300 questions from MMMU~\citep{yue2024mmmu} validation set and employ GPT-4o for the six languages translation. M3Exam challenges the model with real-world educational questions requiring both textual and visual comprehension. Details on how we ensure the translation quality, as well as detailed descriptions of other datasets, can be found in \autoref{sec:eval_dataset}.

\subsection{Text-Only Multilingual Datasets}

While multimodal tasks are critical for evaluating the holistic capabilities of models like \model, text-only multilingual tasks provide an equally essential dimension to assess. Most existing multimodal evaluations tend to overlook the importance of text-only evaluation, especially across diverse languages. Including text-only tasks in \evaldata allows us to examine whether the model can perform well in scenarios that require deep linguistic understanding without the aid of visual context, highlighting its performance as a foundation model. We include three tasks QA, Translation, and Reasoning covering five datasets for the text-only evaluations in \evaldata. 

Specifically, we include TydiQA~\citep{clark2020tydi} to test the model's ability to answer questions across 11 typologically diverse languages. We adopt the FLORES~\citep{nllb2024scaling} dataset to assess machine translation performance.
We sample 11 languages (denoted as FLORES-Sub). We use MMMLU~\citep{MMMLU}, a human-translated version of MMLU to test the general language understanding.
We use XStoryCloze~\citep{lin2022fewshotlearningmultilinguallanguage} and MGSM~\citep{shi2022mgsm} to test the model's commonsense and mathematical reasoning ability in multilingual contexts respectively.


\section{Experiments}

\subsection{Experimental Setup}

\begin{wraptable}[12]{r}{0.55\linewidth}
\centering
\small
\vspace{-70pt}
\begin{tabular}{@{}lcc@{}}
\toprule
\textbf{Stages} & \textbf{Pretraining} & \textbf{Finetuning}\\ \midrule
\multicolumn{3}{c}{\textbf{Training Data}} \\ \midrule
Dataset & LLaVA LCS-558K & \traindata \\
\#Samples & 558K & 6M\\ \midrule
\multicolumn{3}{c}{\textbf{Model}} \\ \midrule
Trainable & Projector (20M) & Full Model (8B)
\\ \midrule
\multicolumn{3}{c}{\textbf{Training}} \\ \midrule
Batch Size & 128 & 128\\
LR: $\psi_{\text{vision}}$ & 1 $\times$ $10^{-3}$ & 2 $\times$ $10^{-6}$ \\
LR: $\{\theta_{\text{proj}}, \phi_{\text{LLM}}\}$ & 1 $\times$ $10^{-3}$ & 2 $\times$ $10^{-5}$\\
Epoch & 1 & 1\\ 
GPU Hours (H100) & 32 & 1344\\\bottomrule
\end{tabular}
\caption{\model's training configurations.}
\label{tab:stages}
\end{wraptable}

We train \model on \traindata, our multilingual multimodal dataset comprising 6 million samples across 39 languages. 
The model uses LLaVA-Next as architecture~\citep{liu2024llavanext}, Qwen2-7B-Instruct~\citep{yang2024qwen2} as the language model backbone and clip-vit-large-patch14-336~\citep{radford2021learning} as the vision encoder. 
The training consists of two stages. First, we pretrain the vision-language connector that aligns the outputs of vision encoder to backbone, with the LLaVA LCS-558K\footnote{\url{https://huggingface.co/datasets/liuhaotian/LLaVA-Pretrain}}~\citep{liu2023llava,liu2023improvedllava}. 
Then, we perform finetuning on \traindata, where we employ a learning rate of 2e-5, a batch size of 512, coupled with a cosine decay schedule with 0.03 warmup steps. 
We pretrain and finetune the model for 1 epoch, where pretraining took 4 hours with 8 H100 (32 GPU hours), and finetuning took 168 hours with 8 H100 (1344 GPU hours).

For evaluation, we compare \model against several state-of-the-art open source baselines, including English-centric models Llava-1.5-7B~\citep{liu2023improvedllava}, Llava-Next-7B~\citep{liu2024llavanext}, Phi-3.5-Vision~\citep{abdin2024phi3technicalreporthighly}, Cambrian-8B~\citep{tong2024cambrian}, Llava-OV-7B~\citep{li2024llava}, Molmo-7B-D~\citep{deitke2024molmo} Llama3.2-11B~\citep{dubey2024llama} and multilingual models PaliGemma-3B~\citep{beyer2024paligemma}, PALO-7B~\citep{PALO}, mBLIP mT0-XL and mBLIP BLOOMZ~\citep{geigle_etal_2024_mblip}. We also consider two text-only LLMs baselines Vicuna-1.5-7B~\citep{zheng2023judging} and Qwen2-7B-Instruct~\citep{yang2024qwen2}, which are the backbones of Llava-Next and our \model respectively. We integrate our multimodal tasks in \evaldata into \texttt{lmms-eval}~\citep{lmms_eval2024},
a multimodal evaluation package that supports many English multimodal benchmarks. We use \texttt{lm-evaluation-harness}~\citep{biderman2024lessons}
to evaluate text-only tasks. We follow the original paper for their best models' prompts in different tasks, and mostly reproduce their original numbers on datasets reported in the original papers. 

\begin{table}[!t]
\small
\centering
\resizebox{\columnwidth}{!}{%
\begin{tabular}{@{}lcccccccccc@{}}
\toprule
\multirow{3}{*}{Models} & \multicolumn{2}{c}{\multirow{2}{*}{AVG (all)}} & \multicolumn{4}{c}{Multimodal Chat} & \multicolumn{4}{c}{Cultural Understanding} \\ \cmidrule(l){4-11} 
 & \multicolumn{2}{c}{} & \multicolumn{2}{c}{xChatBench} & \multicolumn{2}{c}{M-LlavaBench} & \multicolumn{2}{c}{CVQA} & \multicolumn{2}{c}{MaRVL} \\ \cmidrule(l){2-11} 
 & en & mul & en & mul & en & mul & en & mul & en & mul \\ \midrule
Gemini-1.5-Pro & 67.1 & 62.5 & 67.0 & 54.4 & 103.4 & 106.6 & 75.9 & 75.7 & 76.4 & 72.0 \\
GPT4o & 68.6 & 64.6 & 71.0 & 64.4 & 104.6 & 100.4 & 79.1 & 79.4 & 81.4 & 82.1 \\ \midrule
Llava-1.5-7B & 45.4 & 28.4 & 28.5 & 11.8 & 66.1 & 40.8 & 48.9 & 36.5 & 56.2 & 53.7 \\
Llava-Next-7B & 51.1 & 32.7 & 40.5 & 18.9 & 78.9 & 50.7 & 55.7 & 42.6 & 62.8 & 50.9 \\
Phi-3.5-Vision & 54.0 & 35.0 & 38.5 & 13.2 & 70.8 & 58.0 & 56.3 & 42.3 & 72.1 & 56.5 \\
Cambrian-8B & 50.9 & 36.4 & 27.5 & 11.3 & 78.4 & 61.8 & 59.7 & 47.5 & {\ul 75.4} & 61.8 \\
Llava-OV-7B & {\ul 59.5} & 41.3 & 51.0& {\ul 28.5} & 89.7 & 55.3 & {\ul 65.2} & 53.7 & 72.7 & 57.5 \\
Molmo-7B-D & 55.4 & 34.1 & \textbf{49.5} & 21.1 & \textbf{95.9} & 13.8 & 59.4 & 48.3 & 65.3 & 54.9 \\
Llama3.2-11B & 57.2 & 41.9 & {\ul 49.0} & 27.8 & {\ul 93.9} & 58.2 & \textbf{70.2} & \textbf{61.4} & 64.5 & 58.1 \\ \midrule
PaliGemma-3B & 37.3 & 25.8 & 6.0 & 3.5 & 32.1 & 31.9 & 52.9 & 42.9 & 56.5 & 52.2 \\
PALO-7B & 46.3 & 32.2 & 27.0 & 11.8 & 68.9 & 71.2 & 50.9 & 39.2 & 63.3 & 54.2 \\
mBLIP mT0-XL & 35.1 & 29.8 & 2.5 & 0.5 & 32.7 & 28.2 & 40.5 & 37.5 & 67.3 & {\ul 66.7} \\
mBLIP BLOOMZ & 36.1 & 30.0 & 4.0 & 1.6 & 43.5 & 41.0 & 44.9 & 36.9 & 62.3 & 58.6 \\ \midrule
\model-7B (Ours) & \textbf{59.9} & \textbf{52.8} & 46.0 & \textbf{35.8} & 84.2 & 89.5 & 64.4 & {\ul 57.2} & \textbf{87.0} & 79.0 \\
\rowcolor{LightCyan} 
$\Delta$ over SoTA Open & {+0.4} & {+10.9} & -3.5 & {+7.3} & -11.7 & +18.3 & -5.8 & -4.2 & {+11.6} & {+12.3} \\ \midrule
\multirow{3}{*}{Models} & \multicolumn{2}{c}{Captioning} & \multicolumn{4}{c}{Short VQA} & \multicolumn{4}{c}{Multi-subject Reasoning} \\ \cmidrule(l){2-11} 
 & \multicolumn{2}{c}{XM100} & \multicolumn{2}{c}{xGQA} & \multicolumn{2}{c}{MaXM} & \multicolumn{2}{c}{xMMMU} & \multicolumn{2}{c}{M3Exam} \\ \cmidrule(l){2-11} 
 & en & mul & en & mul & en & mul & en & mul & en & mul \\ \midrule
Gemini-1.5-Pro & 27.6 & 19.1 & 54.2 & 48.7 & 56.4 & 63.5 & 65.8 & 57.7 & 77.4 & 64.7 \\
GPT4o & 27.7 & 19.1 & 55.8 & 51.0& 60.7 & 65.4 & 69.1 & 58.3 & 68.0 & 61.0 \\ \midrule
Llava-1.5-7B & 28.6 & 1.1 & 62.0 & 30.6 & 49.8 & 20.4 & 36.2 & 31.5 & 32.3 & 29 \\
Llava-Next-7B & 29.3 & 9.4 & \textbf{64.8} & 37.8 & {\ul 54.9} & 21.4 & 36.7 & 34.3 & 36.5 & 28.4 \\
Phi-3.5-Vision & 30.2 & 5.2 & {\ul 64.7} & 38.4 & \textbf{55.3} & 25.0 & 42.6 & 38.8 & 55.8 & 37.2 \\
Cambrian-8B & 20.6 & 9.9 & 64.6 & 39.8 & \textbf{55.3} & 28.7 & 41.8 & 33.2 & 34.7 & 33.4 \\
Llava-OV-7B & \textbf{30.6} & 7.0 & 64.4 & {\ul 48.2} & 54.9 & 34.8 & {\ul 46.3} & 41.0 & {\ul 60.4} & \textbf{45.8} \\
Molmo-7B-D & 22.1 & 9.1 & 51.5 & 43.0 & 52.9 & 37.5 & 44.5 & 40.4 & 57.1 & 39.1 \\
Llama3.2-11B & 27.6 & 4.5 & 55.6 & 45.4 & \textbf{55.3} & {\ul 43.9} & \textbf{46.5} & 41.4 & 51.8 & 36.6 \\\midrule
PaliGemma-3B & 18.7 & 0.8 & 59.7 & 30.5 & 47.9 & 19.9 & 26.3 & 25.2 & 36.0 & 25.6 \\
PALO-7B & {\ul 30.4} & 0.8 & 60.5 & 37.8 & 51.4 & 16.3 & 33.1 & 30.5 & 30.8 & 27.8 \\
mBLIP mT0-XL & 31.9 & 3.1 & 44.2 & 39.9 & 44.7 & 36.8 & 29.3 & 30.4 & 22.8 & 25 \\
mBLIP BLOOMZ & 22.5 & {\ul 10.3} & 43.3 & 36.9 & 44.7 & 24.8 & 29.2 & 30.8 & 30.3 & 29.5 \\ \midrule
\model-7B (Ours) & {\ul 30.4} & \textbf{14.2} & {\ul 64.7} & \textbf{60.2} & \textbf{55.3} & \textbf{53.3} & \textbf{45.7} & \textbf{43.7} & \textbf{61.4} & {\ul 42.1} \\\rowcolor{LightCyan} 
$\Delta$ over Best Open Model & -0.2 & {+3.9} & -0.1 & {+12.0} & 0.0 & {+9.4} & -0.8 & {+2.3} & {+1.0} & -3.7 \\ \bottomrule
\end{tabular}%
}
\caption{Overall performance on the multilingual multimodal benchmarks in \evaldata. The best-performing open model on each dataset is in \textbf{bold} and the second best is {\ul underlined}. }
\vspace{-15pt}
\label{tab:overall_results}
\end{table}


\subsection{Multilingual Multimodal Results}

The results in \autoref{tab:overall_results} provide clear insights into the strengths and remaining challenges of \model-7B in multilingual and multimodal tasks. Key observations from the evaluation include:

\noindent\textbf{Superior English and Multilingual Performance:} \model-7B outperforms existing open-source models across both English and multilingual tasks.  While concurrent multimodal models such as Molmo~\citep{deitke2024molmo} or Llama 3.2 show strong performance on English datasets, they struggle in multilingual evaluation settings. 
Particularly in multilingual subsets like xChatBench, M-LlavaBench, and MaRVL, it has achieved substantial gains, highlighting its effectiveness in both cross-lingual and cross-cultural contexts.

\noindent\textbf{Balanced Cross-Language Capabilities:} Unlike many models that exhibit a significant drop in performance when moving from English to multilingual tasks, \model-7B is relatively consistent. For instance, in Multimodal Chat tasks, the performance gap between English and multilingual remains relatively small, indicating its ability to handle multiple languages effectively.

\noindent\textbf{Challenges Compared to Proprietary Models:} While \model-7B leads in open-source models, some gaps remain when compared to closed-source models like GPT4o. Additionally, though \model-7B narrows the gap between English and multilingual performance, there is still room for improvement in fully closing this divide across all tasks.

\subsection{Multilingual Text-only Results}
We further evaluate our model in text-only scenarios in \autoref{tab:text_only_result}. Interesting findings include:

\noindent\textbf{Best Text Performance Among Multimodal LLMs:} \model-7B demonstrates the strongest performance among all multimodal LLMs in the text-only tasks consistently outperforming baselines like Llava-Next-7B. This highlights that, despite being trained as a multimodal model, \model-7B maintains superior text understanding and reasoning capabilities compared to other MLLMs.

\noindent\textbf{Maintained Performance from its Text Backbone.}  \model-7B generally maintains or sees slight drops in performance on most text-only benchmarks compared with its text backbone Qwen2-7B-Instruct. Notably, the model shows a significant improvement in MGSM. This improvement is directly attributable to the inclusion of math-related instructions in \traindata, which enhances the model's capability to handle complex multilingual reasoning and mathematical tasks.

\begin{table}[!t]
\small
\centering
\resizebox{\columnwidth}{!}{%
\begin{tabular}{@{}lcccccccccccc@{}}
\toprule
\multirow{2}{*}{Models} & \multicolumn{2}{c}{AVG (all)} & \multicolumn{2}{c}{FLORES-Sub} & \multicolumn{2}{c}{TyDiQA} & \multicolumn{2}{c}{XStoryCloze} & \multicolumn{2}{c}{MGSM} & \multicolumn{2}{c}{MMMLU} \\ \cmidrule(l){2-13} 
 & en & mul & x→en & en→x & en & mul & en & mul & en & mul & en & mul \\ \midrule
Vicuna-1.5-7B & 52.1 & 38.7 & 55.6 & 42.4 & 59.7 & 52.7 & 78.1 & 57.4 & 17.6 & 6.4 & 49.5 & 34.7 \\
Qwen2-7B-Instruct & {\ul 66.6} & \textbf{54.5} & \textbf{61.8} & \textbf{46.0} & 72.2 & \textbf{71.2} & \textbf{80.3} & \textbf{61.9} & 48.8 & {\ul 40.4} & \textbf{70.1} & \textbf{53.1} \\ \midrule
Llava-1.5-7B & 53.1 & 39.0 & 54.7 & 41.5 & 66.8 & 52.8 & 79.1 & 57.6 & 14.8 & 7.6 & 50.2 & 35.7 \\
Llava-Next-7B & 54.0 & 38.9 & 54.8 & 41.4 & 68.3 & 52.1 & 79.1 & 57.1 & 15.6 & 7.5 & 52.1 & 36.5 \\
Phi-3.5-Vision & 60.7 & 41.7 & 28.5 & 32.5 & \textbf{75.9} & 51.3 & 77.9 & 54.8 & \textbf{59.2} & 33.1 & 62.0 & 36.7 \\
PALO-7B & 52.0 & 37.5 & 52.9 & 40.4 & 69.4 & 50.8 & 77.4 & 57.2 & 13.6 & 5.8 & 46.7 & 33.4 \\ \midrule
\model-7B (Ours) & \textbf{72.8} & {\ul 54.3} & {\ul 60.7} & {\ul 44.9} & {\ul 73.7} & {\ul 66.0} & {\ul 79.1} & {\ul 61.2} & \textbf{82.0} & \textbf{47.4} & {\ul 68.4} & {\ul 52.2} \\ \bottomrule
\end{tabular}
}
\caption{Overall performance on text-only multilingual benchmarks in \evaldata.}

\label{tab:text_only_result}
\end{table}



\section{Discussion}
\label{sec:discussion}

Finally, we explore implications of our findings and their potential impact on future developments in the field. We examine the scaling effects of instruction quantity, the persistent role of English data, the relationship between training sample proportions and performance. Through this discussion, we aim to provide a comprehensive understanding of our model and chart a course for future advancements. More discussion on qualitative examples of model behavior on multilingual multimodal chat and challenges in multilingual OCR can be found in \autoref{appendix:xchat} and \autoref{sec:multingual_ocr_details}.

\begin{figure}[!h]
    \centering
        \vspace{-10pt}

\includegraphics[width=1.0\linewidth]{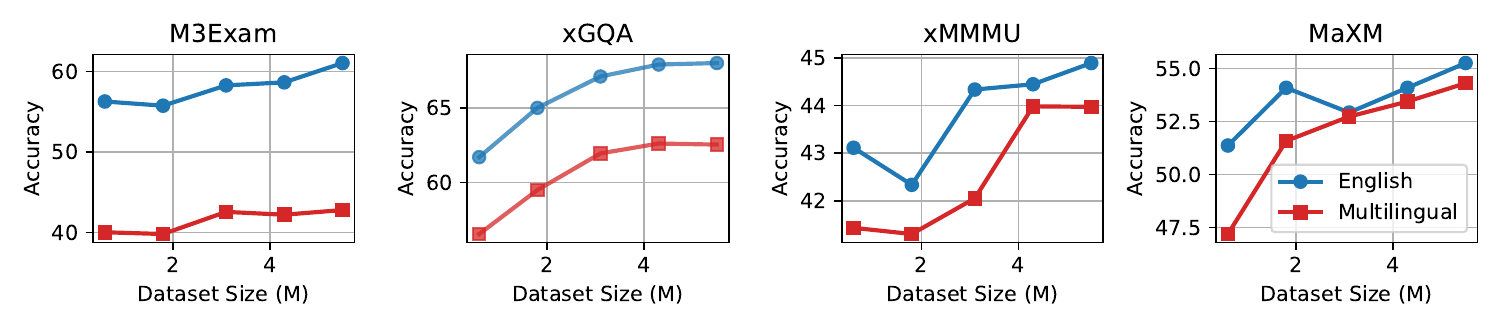}
    \vspace{-20pt}
    \caption{Scaling effect of training samples on English and multilingual scores across datasets.}
    \label{fig:scaling_effect}
\end{figure}

\noindent\textbf{Scaling Effect of Number of Instructions.} Understanding how the quantity of instructions affects model performance is crucial for optimizing training strategies and resource allocation. \autoref{fig:scaling_effect} reveals a clear scaling effect related to the number of instructions used during training. Performance improvements were consistent as we increased the number of multilingual instructions in \traindata, for both English and multilingual performance. This demonstrates the necessity of scaling multilingual multimodal instruction tuning. 

\noindent\textbf{Role of English Data.} In multilingual scenarios, English data plays a pivotal role in cross-lingual transfer. To investigate this, we sampled 500K examples from the translated data described in \autoref{sec:machine_translation}, ensuring a consistent data distribution. We varied the ratio of English data while keeping the total number of training samples fixed at 500K. For the 17 multilingual languages in the translated subset, we evenly distributed the number of samples across languages. 

\begin{wrapfigure}{r}{0.6\linewidth}
    \centering
    \vspace{-10pt}
    \includegraphics[width=\linewidth]{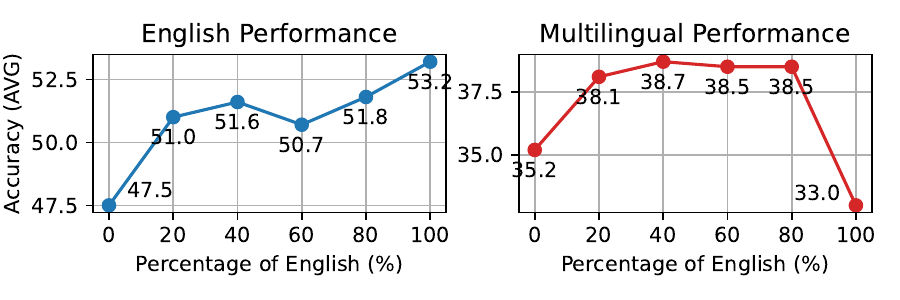}
    \vspace{-25pt}
    \caption{Impact of English training data proportion on English vs. multilingual performance.}
        \vspace{-10pt}
\label{fig:english_ratio}
\end{wrapfigure}

As shown in \autoref{fig:english_ratio}
, English performance generally improves as the percent of English data increases. Surprisingly, using only multilingual data results in relatively lower multilingual performance. As we introduce more English data, multilingual performance improves, peaking at 38.7\% with 40\% English. However, performance drops sharply when English data reaches 100\%. This suggests that English data aids cross-lingual transfer, however, over-reliance on it harms multilingual performance.

\begin{figure}[!h]
    \centering
    \vspace{-10pt}
\includegraphics[width=1.0\linewidth]{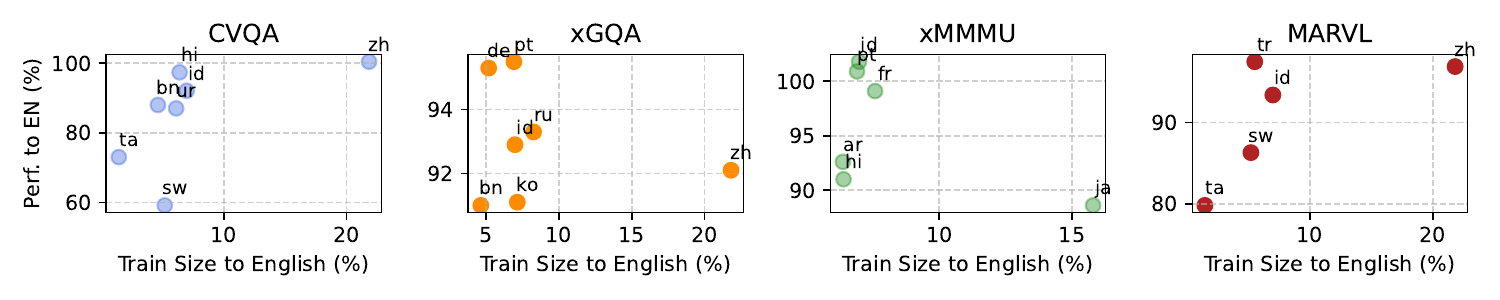}
    \vspace{-20pt}
    \caption{The relationship between training sample size (relative to English) and performance (relative to English) of different languages across four datasets.}
\label{fig:language_portion_downstream_performance}
\end{figure}

\noindent\textbf{How does the proportion of training samples in a language affect downstream performance?} Is downstream task performance correlated with the number of training samples? 
Our analysis in \autoref{fig:language_portion_downstream_performance} revealed the relationship between training sample proportion and downstream performance. While there is a general positive correlation, the impact varies significantly across languages and tasks. For widely spoken languages with rich resources, we observed a near-linear relationship. However, for low-resource languages, even a small increase in proportion yielded disproportionately large performance gains. Interestingly, we also noted instances of positive transfer between typologically similar languages. These findings suggest that strategic allocation of training samples, considering both language prevalence and linguistic similarities, can optimize model performance.

\section{Conclusion}
In this paper, we introduced \model, a  multilingual  MLLM designed to bridge linguistic and cultural gaps in visual understanding tasks. By leveraging \traindata, our newly curated 6M multilingual multimodal instruction data samples, we demonstrated significant improvements in cross-lingual and cross-cultural understanding across 39 typologically diverse languages. Our comprehensive evaluation using \evaldata revealed \model's superior performance compared to existing open-source models. We also highlight ongoing challenges in areas such as low-resource language support and multilingual OCR. We fully open-source \model, \traindata, and \evaldata to facilitate future research to build open and inclusive MLLMs.

\section*{Acknowledgments}

This work was supported in part by a Carnegie Bosch Institute Fellowship to Xiang Yue, as well as grants from DSTA Singapore, and the Programs for Bridging the gap between R\&D and the IDeal society (society 5.0) and Generating Economic and social value (BRIDGE)/Practical Global Research in the AI × Robotics Services, implemented by the Cabinet Office, Government of Japan. The training is supported by the CMU FLAME Center. The authors would like to thank Google Gemini credits for data construction and evaluation. The authors would also like to thank CMU NeuLab colleagues for their constructive comments.

\bibliography{iclr2025_conference}
\bibliographystyle{iclr2025_conference}

\newpage
\appendix  

\DoToC
\clearpage

\newpage

\section{Related Work}

\paragraph{Visual Instruction Tuning.}
Visual instruction tuning is a key technique for enhancing multimodal large language models by aligning visual inputs with textual instructions to improve understanding and generation tasks \citep{liu2023llava}. Traditionally, these instructions are built using English-language data from visual question answering and other datasets \citep{liu2023llava,liu2023improvedllava,xu-etal-2023-multiinstruct,liu2024llavanext,tong2024cambrian,beyer2024paligemma,zhu2024minigpt,dai2023instructblip,tong2024cambrian,li2024omnicorpus}. Researchers often supplement this with synthetic instruction tuning data, generating large volumes of instructional pairs to possibly cover multiple languages too \citep{geigle_etal_2024_mblip,li2023m3it,li2024omnicorpus}. However, these instruction-tuning datasets have mostly been task-focused and lack conversational capabilities. Further, while translation gives lends to multilingual capabilities, the data remains to be culturally homogeneous. By curating multilingual and multicultural instruction tuning data across various task types, our model is designed to intuitively understand and engage with users from diverse demographics.

\begin{table}[h!]
\centering
\resizebox{\columnwidth}{!}{
\begin{tabular}{lccccc}
\toprule
\textbf{Dataset} & \textbf{\# Languages} & \textbf{\# of Instances} & \textbf{Multicultural} & \textbf{\# of Task Types} & \textbf{Open-Sourced} \\
\midrule
MultiInstruct \citep{xu-etal-2023-multiinstruct} & 1 & $\sim$235.0K & \ding{55} & 310 & \ding{55} \\
MiniGPT4 \citep{zhu2024minigpt} & 1 & 5.0K & \ding{55} & 149 & $\checkmark$ \\
LLAVA \citep{liu2023llava} & 1 & 1.2M & \ding{55} & $>$100K & $\checkmark$ \\
InstructBLIP 
\citep{dai2023instructblip} & 1 & $\sim$1.6M & \ding{55} & $>$100K & \ding{55} \\
M$^3$IT \citep{li2023m3it} & 80 & 2.4M & \ding{55} & 400 & $\checkmark$ \\
mBLIP \citep{geigle-etal-2024-mblip} & 95 & 5.1M & \ding{55} & 68 & $\checkmark$ \\ 
PALO \citep{PALO} & 10 & 2.1M & \ding{55} & 22 & $\checkmark$\\
Cambrian \citep{tong2024cambrian} & 1 & 7.1M & \ding{55} & $>$1M & $\checkmark$ \\
\traindata (Ours) & 39 & 6.2M & $\checkmark$ & $>$1M & $\checkmark$ \\
\bottomrule
\end{tabular}}
\caption{Comparison of datasets in terms of number of languages, number of instances, whether the dataset is multicultural, number of task types, and open-sourced. }
\label{tab:comparison}
\end{table}


\paragraph{Multilingual Multimodal LLMs.}

Multilingual MLLMs have evolved from dual-encoder-based models, only capable of understanding and reasoning \citep{ni2021m3p, zeng2022cross, jain2021mural}, to encoder-decoder models capable of multilingual text generation as well \citep{shan2022ernie, chen2023pali, geigle_etal_2024_mblip}. Despite their advancements, these models have remained focused on conventional tasks such as VQA and image captioning. Moreover, most efforts have centered around training with multilingual text, while little attention has been given to curating culturally diverse image datasets. Even for text, despite the focus on multilinguality, few attempts have been made to reflect cultural diversity in instructions and captions. As a result, these models tend to reflect a Western-centric bias. By selecting culturally diverse images from LAION and intentionally integrating this diversity into our instructions and captions, our model aims to serve a wide range of users in an inclusive and equitable manner.
\newpage

\section{Prompts used in the data construction}
In this appendix, we will list the detailed prompts we used when constructing cultural understanding instruction tuning data described in \autoref{sec:multicultural_data}.

\begin{promptbox}[Cultural Images LLM Scoring Prompt]{lightgreen}
\label{llm_scoring}
You are given an [Alt Text] associated with an image from the web.\\

\textbf{[Alt Text]}: \{Alt Text\}\\

Your goal is to:
\\
\begin{enumerate}[leftmargin=*]
    \item \textbf{Evaluate Text Quality:} Rate the following alt text on a scale from 1 to 5 based on its quality in describing the image, assuming the model does not have access to the image:
    \begin{itemize}[leftmargin=10pt]
        \item 1 (Very Low Quality): Alt text is vague, irrelevant, misleading, or uses placeholders (e.g., file names).
        \item 2 (Low Quality): Alt text is overly simplistic, generic, or provides minimal useful information.
        \item 3 (Moderate Quality): Alt text is somewhat descriptive but lacks detail or relevance, with possible redundancy or ambiguity.
        \item 4 (High Quality): Alt text is descriptive, clear, concise, and provides sufficient information to understand the image's content.
        \item 5 (Very High Quality): Alt text is highly specific, detailed, and relevant, with a clear description that conveys all key aspects of the image.
    \end{itemize}
    
    \item \textbf{Subject Classification:} Assign a subject/category to the alt text based on its content. Choose from the following categories:
    \begin{itemize}[leftmargin=10pt]
        \item Vehicles and Transportation
        \item Cooking and Food
        \item People and Everyday Life
        \item Sports and Recreation
        \item Plants and Animals
        \item Objects, Materials, and Clothing
        \item Brands and Products
        \item Geography, Buildings, and Landmarks
        \item Tradition, Art, and History
        \item Public Figure and Pop-Culture
        \item Others
    \end{itemize}

    \item \textbf{Country/Region Classification:} Decide if the alt text is closely related to a specific country's culture. For example, if the alt text says, "Tokyo Skytree Photo in March with beautiful cherry blossoms", it's strongly related to Japan. If the alt text is not specifically about a certain culture or country, you can say "No specific country". Even if the alt text is written in their official language, it doesn't mean the caption is specifically about the country (e.g., a product page caption is often unlikely to be country-specific).
\end{enumerate}

\textbf{Output:}  
Provide the final result in the following structured format:
\begin{enumerate}[leftmargin=*]
    \item \textbf{Text Quality Score (1-5):}
    \item \textbf{Subject Category:}
    \item \textbf{Country/Region:}
\end{enumerate}

Only generate the final result without any additional descriptions or explanations.

\end{promptbox}

\begin{promptbox}[Image Recaption Prompts]{lightred}
We randomly select one recaption prompt from the following:\\

\textbf{PROMPT 1:}\\
Please describe the image in detail in \{language\}. The image might be related to the country: "\{country\}". The topic might be related to: "\{category\}". The previous short caption of the image is \{text\}.\\

\textbf{PROMPT 2:}\\
Analyze this image and provide a comprehensive description in "\{language\}". Consider that it may be associated with "\{country\}" and the theme could be related to "\{category\}". If there is cultural significance, please include it. A brief previous description was: \{text\}.\\

\textbf{PROMPT 3:}\\
In "\{language\}", give a detailed description of what you see in this image. Keep in mind it might be connected to "\{country\}" and the subject could be about "\{category\}". If there are culturally relevant details, please include them. An earlier short description stated: \{text\}.\\

\textbf{PROMPT 4:}\\
Examine this image closely and describe its contents in "\{language\}" in a more structured way. The image might have a connection to "\{country\}" and could be about "\{category\}". A previous concise caption mentioned: \{text\}.\\

\textbf{PROMPT 5:}\\
Using "\{language\}", provide an in-depth and structured description of this image. It may be related to "\{country\}" and the topic could be associated with "\{category\}". A prior brief description was given as: \{text\}.

\end{promptbox}

\begin{promptbox}[Instruction Generation Prompt]{darkblue}
\label{ins_gen_prompt}

\textbf{Task}: Generate two \textbf{instruction-response pair} based on the visual content of an image. Choose two task types from the list below to guide the rewriting process:

\begin{itemize}[leftmargin=20pt]
    \item Coding \& Debugging
    \item Information Seeking
    \item Creative Writing
    \item Critical Reasoning
    \item Planning \& Strategy
    \item Mathematical Thinking
    \item Text Revision \& Editing
    \item Data Analysis
    \item Role Playing \& Scenarios
    \item Brainstorming \& Ideation
    \item Advice Seeking \& Problem-Solving
    \item Learning \& Understanding
    \item Cultural Interpretation
\end{itemize}

\textbf{Guidelines}:\\

\textbf{Instruction}:
\begin{itemize}[leftmargin=20pt]
    \item Select two different task types from the list above.
    \item Make sure the instruction prompts an interpretation or analysis \textbf{directly tied to what can be visually observed in the image}, not just general reasoning.
    \item The instruction should require a response that \textbf{uses details from the image}. Avoid generic instructions that can be answered without visual information.
\end{itemize}

\textbf{Response}:
\begin{itemize}[leftmargin=20pt]
    \item Provide a \textbf{very detailed and structured} response that reflects a clear understanding of the implied visual information.
    \item Offer multiple perspectives, deep analysis, or step-by-step explanations where applicable.
    \item Avoid general responses that could be inferred without observing the image. Responses must rely on interpreting the visual content.
\end{itemize}

\textbf{Content}:
\begin{itemize}[leftmargin=20pt]
    \item Instructions should be varied, challenging, and explore different advanced aspects of the visual scene.
    \item Responses must showcase a deep understanding of the image's visual context, using thoughtful insights where applicable.
\end{itemize}

\textbf{Output}:
\begin{itemize}[leftmargin=20pt]
    \item Provide the output in JSON format with three keys: ``task\_type'', ``instruction'' and ``response''.
    \item Ensure the instruction and response \textbf{do not mention ``based on caption''} but instead, refer to the \textbf{image} or simply avoid reference to any external description.
    \item Do not include additional text or explanations beyond what is required.
    \item Provide both the ``instruction'' and ``response'' in \{language\} but ``task\_type'' in English.
\end{itemize}

Caption: \{caption\}

\end{promptbox}

\newpage
\section{Recaptioning Example from LAION-Cultural}
\begin{figure}[!h]
    \centering
    \includegraphics[width=0.8\linewidth]{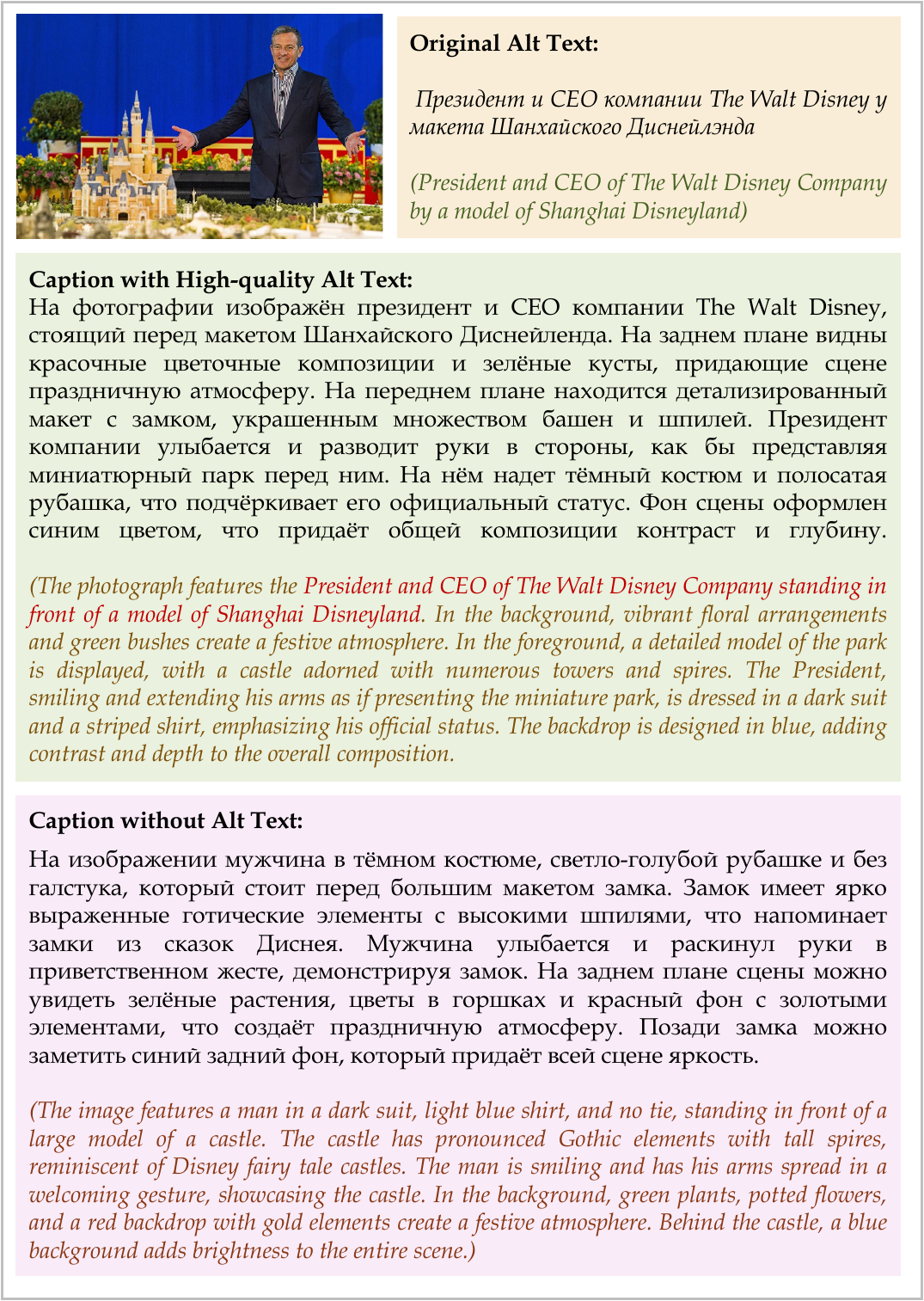}
    \caption{An example from LAION-Cultural illustrating why the filtered informative alt text helps generate a more informative caption. With the high-quality alt text, the model incorporates important details like \textit{``President and CEO of The Walt Disney Company standing in front of a model of Shanghai Disneyland''} into the generated caption. }
    \label{fig:caption_example}
\end{figure}
\newpage
\section{Datasets used in \evaldata}
\label{sec:eval_dataset}
To comprehensively assess the capabilities of \model across diverse languages, cultures, and task types, we developed \evaldata. We list the details of each dataset included in the \evaldata.

\subsection{Multimodal Datasets}

\begin{itemize}[leftmargin=*]
    \item \textbf{xGQA} \citep{pfeiffer2021xgqa}: A cross-lingual visual question-answering dataset featuring 9,666 questions in eight languages covering five scripts. The dataset includes 300 unique images from Visual Genome \citep{krishna2017visual}. xGQA tests the model's ability to understand and reason about visual content across multiple languages.

    \item \textbf{MaXM} \citep{changpinyo2022maxm}: A VQA dataset in seven languages and five scripts, with questions and answers in the same language. Images are culturally matched to the target language regions. MaXM specifically addresses the challenge of cultural diversity in multimodal understanding.

    \item \textbf{MaRVL} \citep{liu2021visually}: A Multicultural Reasoning over Vision and Language dataset in five languages and three scripts, featuring 4,914 culturally diverse images matched to respective languages. MaRVL focuses on evaluating models' ability to reason about culturally diverse visual concepts.

    \item \textbf{XM100} \citep{thapliyal2022crossmodal}: We create a subset of 3600 instances (100 instances per language) from the original XM100 dataset, a large multilingual image captioning dataset comprising 36 languages, with 261,375 captions for 100 unique images per language, culturally matched to each language. XM100 evaluates a model's ability to generate culturally appropriate captions across a wide range of languages. For sampling, we select 100 instances per language, ensuring that all languages share the same set of images for their respective 100 instances. To ensure diversity within our sample, we use Sentence-BERT~\citep{reimers2019sentence} to cluster the 3600 English instances from the original dataset into 100 groups, and then select one instance from each group. This method ensures that the sampled instances are as diverse as possible. We evaluate models on this new sample of 3600 instances, which allows for a more time-efficient evaluation while still accurately reflecting the multilingual capabilities of models in diverse contexts.

    \item \textbf{M3Exam}~\citep{zhang2023m3exam}: A novel benchmark sourced from real and official human exam questions, featuring 12,317 questions in 9 languages across three educational levels. Approximately 23\% of the questions require image processing. M3Exam tests the model's ability to handle complex, multi-step reasoning tasks in an educational context.

    \item \textbf{xMMMU}: MMMU contains multimodal questions from college-level materials across six disciplines and 30 subjects. The dataset features 183 subfields and 30 diverse image types, including charts, diagrams, and chemical structures. We sample 300 questions from the original MMMU validation set and translate them using GPT-4o into xx languages. To ensure the quality, we translated each sampled question multiple times and then back-translated it to English. We select the translation with the highest BLEU score.  xMMMU evaluates the model's capacity to understand and reason about specialized academic content across languages and modalities.
\end{itemize}

\subsection{Text-Only Multilingual Datasets}

\begin{itemize}[leftmargin=*]
    \item \textbf{TyDiQA} \citep{clark2020tydi}: A question answering dataset covering 11 typologically diverse languages with 204K question-answer pairs. Questions are written by native speakers without seeing the answers, ensuring a realistic information-seeking task. TyDiQA is designed to test linguistic diversity and avoid translation artifacts.

    \item \textbf{FLORES} \citep{nllb2024scaling}: A machine translation benchmark for 200 languages, including many low-resource languages. It consists of 3,001 sentences from 842 web articles, divided into dev, devtest, and test splits. FLORES-200 includes translations from multiple pivot languages and provides script alternatives for some languages, making it a comprehensive test of translation capabilities.

    \item \textbf{MMMLU}~\citep{MMMLU}: A human-translated version of MMLU \citep{hendrycksmeasuring2021}, covering 57 subjects across STEM, humanities, social sciences, and more. It ranges in difficulty from elementary to advanced professional levels, testing both world knowledge and problem-solving ability in a zero-shot and few-shot setting across multiple languages.

    \item \textbf{MGSM} \citep{shi2022mgsm}: Multilingual Grade School Math Benchmark, featuring 250 grade-school math problems translated into 10 languages. Based on GSM8K, it requires multi-step reasoning and tests the model's ability to solve complex mathematical word problems across languages.
\end{itemize}

This diverse set of datasets in \evaldata allows for a comprehensive evaluation of \model's capabilities across various languages, cultures, modalities, and task types, providing a holistic assessment of its performance in multilingual and multimodal contexts.

\newpage
\section{Explanation of xChatBench}
\label{appendix:xchat}

\paragraph{Task Category} We first divide into 10 task categories, namely \textit{art\_explanation}, \textit{bar\_chart\_interpretation}, \textit{defeasible\_reasoning}, \textit{figurative\_speech\_explanation}, \textit{iq\_test}, \textit{ocr}, \textit{graph\_interpretation}, \textit{image\_humor\_understanding}, \textit{science\_figure\_explanation}, \textit{unusual\_images}. The task categories are inspired by existing papers that do not use a free-form generation format~\citep{lumathvista,yue2024mmmu,han2023reading,hessel2023androids,kim2022donut}.

\paragraph{Construction Procedure} To annotate the instances, we mainly follow the procedure of \citet{kim2024biggen}. Two human annotators first hand-crafted the instances by searching through appropriate images for the task and then hand-crafting each component of the instance. As our motivation for fine-grained evaluation, each instance consists of not only an \textbf{instruction}, \textbf{reference answer}, but also a unique \textbf{evaluation criteria} tailored to each instance (\textit{e.g.}, Does the response effectively explain the humor in the image based on the juxtaposition of a character's portrayal in different scenarios?) and a \textbf{description for each score} between 1 and 5 (\textit{e.g.}, score4\_description: The response understands the juxtaposition and relates it to the humor involving machine learning models, but may miss some nuances or the related aspect of the humor). During the annotation process, we asked the annotators to not copy-and-paste results from LLM services like ChatGPT or directly from the web. Then, we hire four additional annotators to assess the quality of the instances. Each participant to asked to grade if each instance (1) fits into the devised task category, (2) if the quality of the reference answer is good enough, and (3) if the score rubric is suitable to assess the response. We iteratively ask the annotators who made the instances to revise them if the instance does not satisfy all three criteria. The resulting dataset consists of 50 instructions, reference answers, and evaluation criteria with a corresponding score rubric.

\paragraph{Translation Procedure} To assess the multilingual generation capabilities of MLLMs, we translate the hand-crafted 50 instances into 6 different languages, namely Chinese, Hindi, Indonesian, Japanese, Korean, and Spanish.  We first use GPT-4o-2024-08-06 to translate the instruction and reference answer of each instance with a naive prompt, ``Translate the following sentences into \{target\_language\}. Sentences: \{sentences\}''. Then, the coauthors who are native speakers of each language reviewed the instances and made adjustments if the translated results were unnatural.

\paragraph{Evaluation Pipeline} Similar to prior works employing LLM-as-a-Judge, we use GPT-4o-2024-08-06 as the judge model and prompt it in a direct assessment manner. As input, the judge model is given the instruction, the model's response, the reference answer, the evaluation criteria, and the descriptions for each score. As output, the judge generates verbal feedback and an integer score between 1 and 5. For this procedure, we use the \texttt{prometheus-eval} library~\citep{kim2024prometheus} and employ their default hyperparameter setting for evaluation. Lastly, the final score is acquired by averaging the results across the 50 instances for each language. Note that in the main result and breakdown result tables, we normalize the score from 1-5 to 0-100 by $(\text{score}-1)\times 25$. For the multimodal chat scenarios, we found that many English-centric models tend to respond in English regardless of the query language. This behavior is problematic, as it undermines the fundamental capability of a multilingual model, which should ideally respond in the language of the query. To address this, we implemented a strict evaluation criterion where such responses were penalized and assigned a score of 0. We believe this is crucial, as users may not understand English, and failing to respond in the appropriate language can hinder effective communication and user experience. Thus, for postprocessing, we use langdetect~\footnote{\href{https://pypi.org/project/langdetect/}{https://pypi.org/project/langdetect/}} to identify whether the response is written in the given language and change the score to 1 when it is written in a different language, a phenomenon called \textit{language hallucination}~\citep{xue2020mt5,pfeiffer2023mmt5}.







\clearpage
\section{Qualitative Examples from xChatBench}

\begin{figure}[!t]
    \centering
    \includegraphics[width=\linewidth]{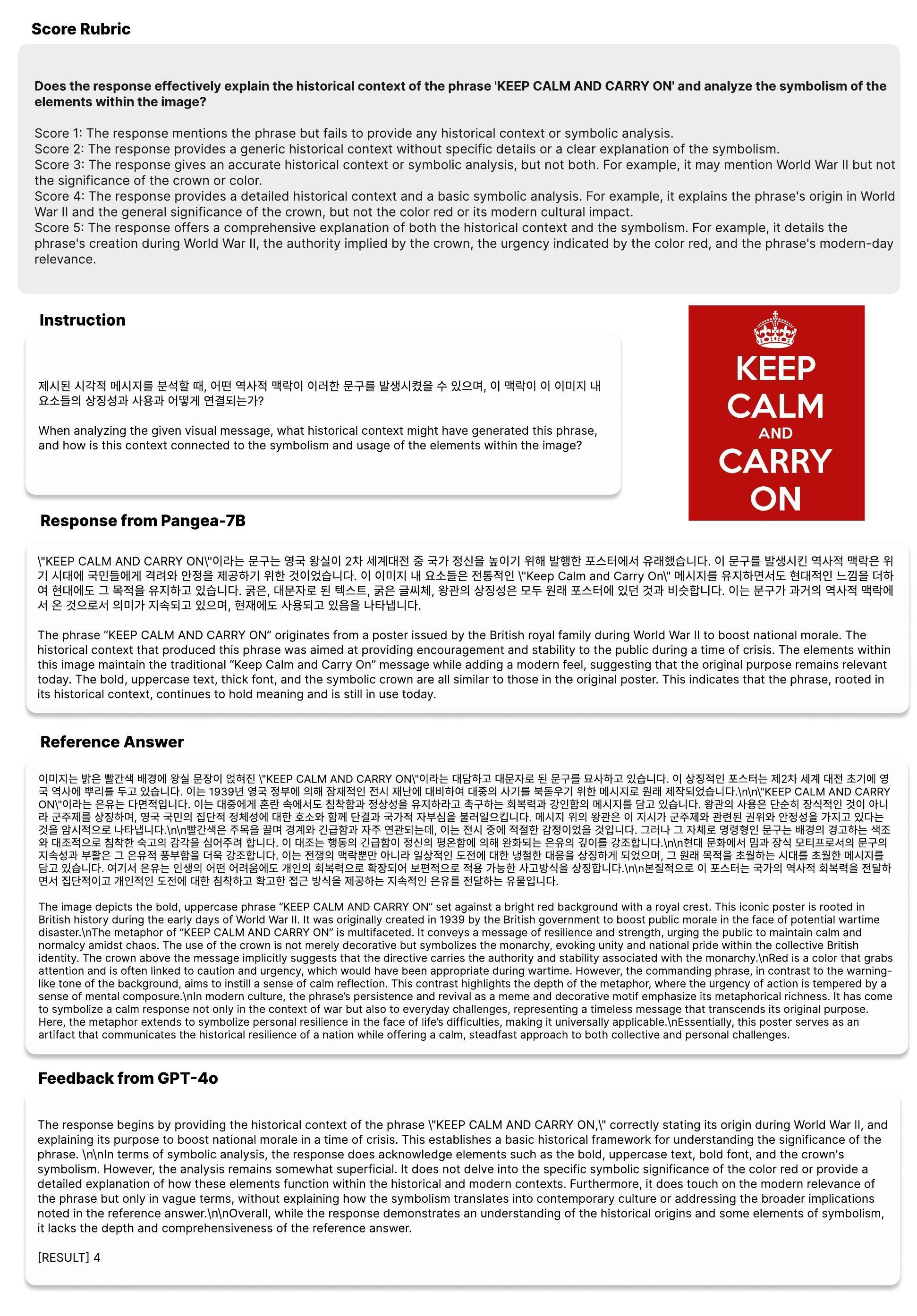}
    \caption{An xChat example from the Korean subset and our model \model's response. The English part of the instruction, response, and reference answer is additionally added only on the figure to help to understand and was not given nor generated by \model.}
    \label{fig:xchat_korean}
\end{figure}

\begin{figure}[!t]
    \centering
    \includegraphics[width=\linewidth]{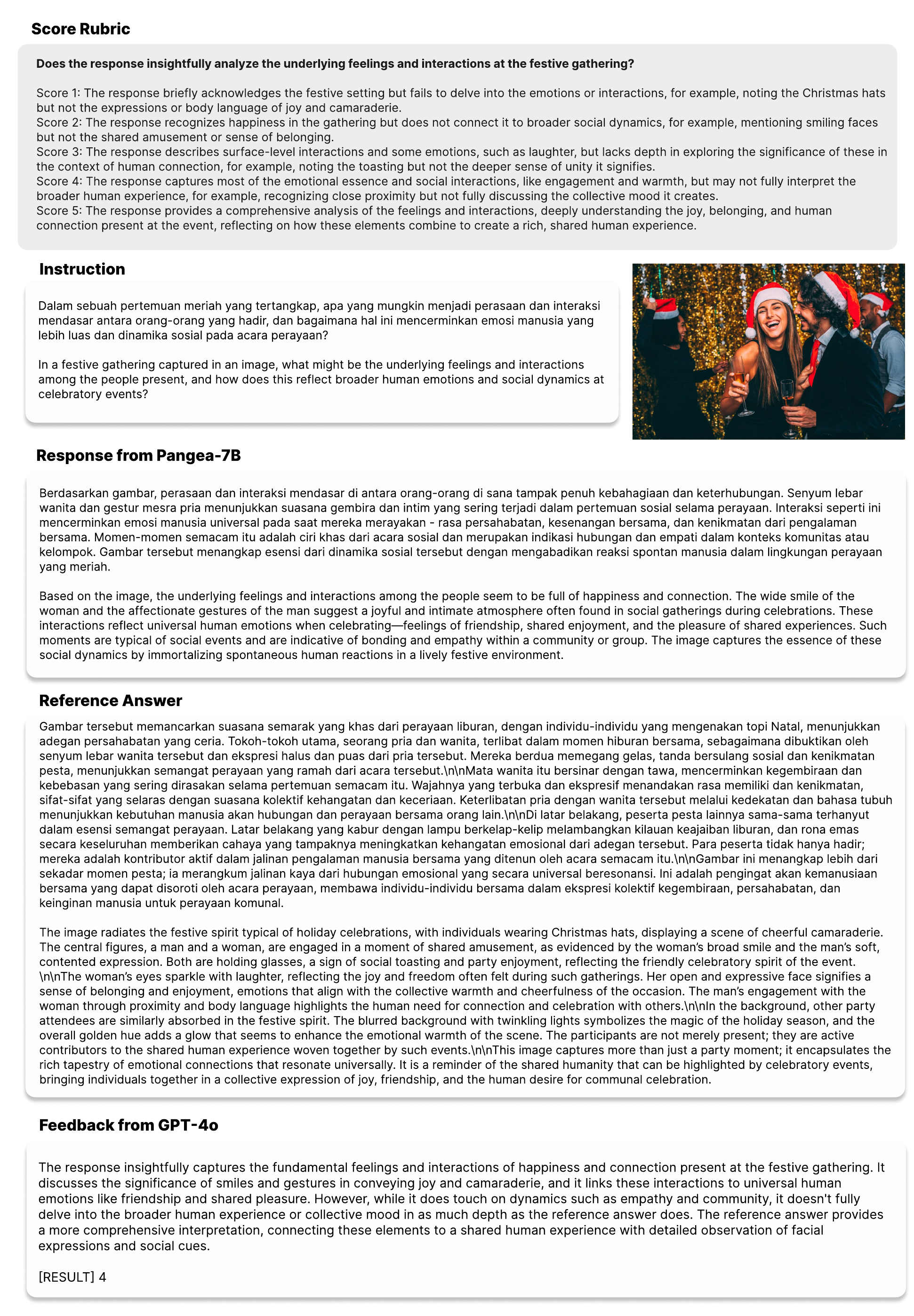}
    \caption{An xChat example from the Indonesian subset and our model \model's response. The English part of the instruction, response, and reference answer is additionally added only on the figure to help to understand and was not given nor generated by \model.}
    \label{fig:xchat_indonesian}
\end{figure}

\begin{figure}[!t]
    \centering
    \includegraphics[width=\linewidth]{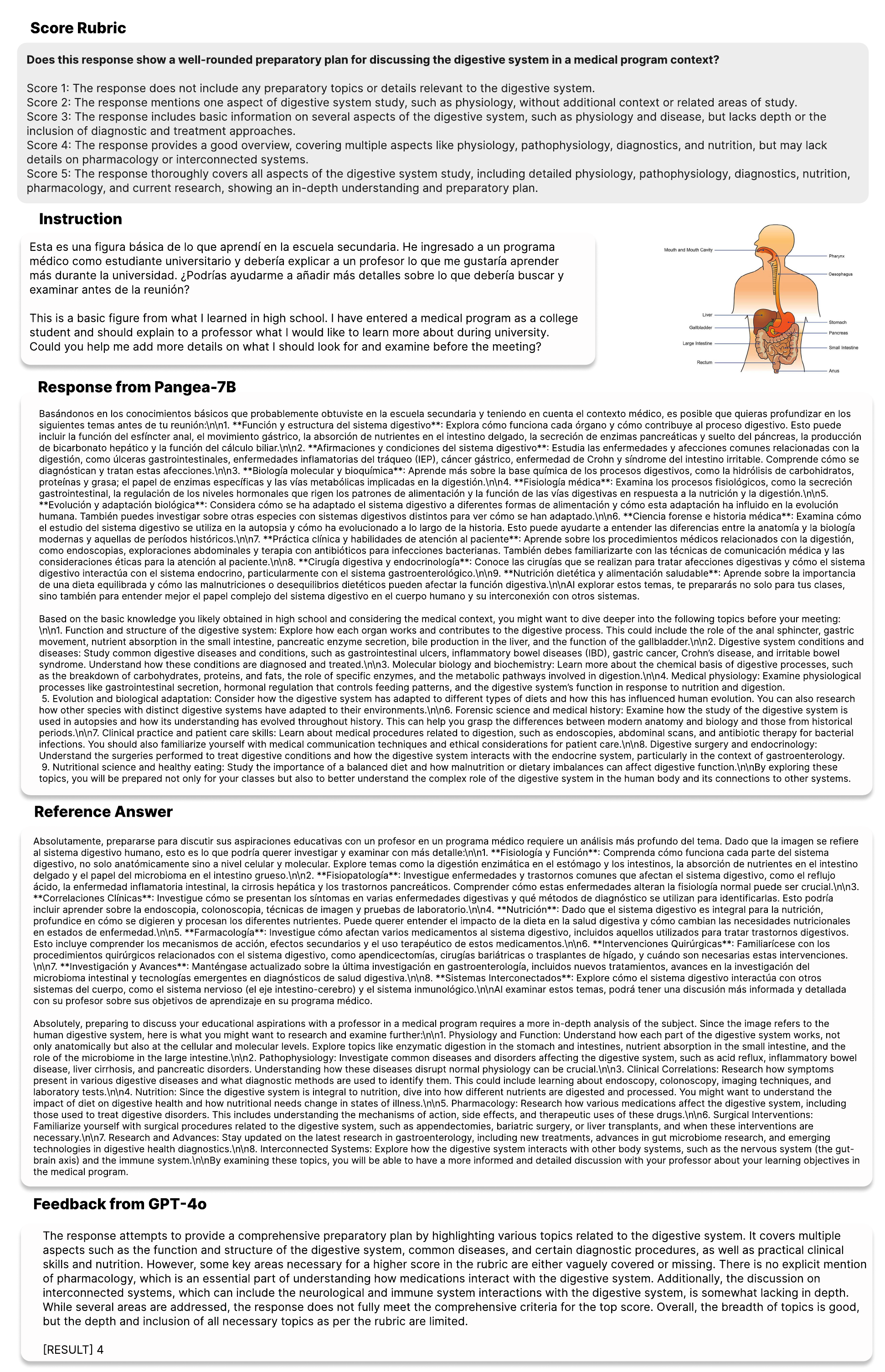}
    \caption{An xChat example from Spanish subset and our model \model's response. The English part of the instruction, response, and reference answer is additionally added only on the figure to help to understand and was not given nor generated by \model.}
    \label{fig:xchat_spanish}
\end{figure}

\begin{figure}[!t]
    \centering
    \includegraphics[width=\linewidth]{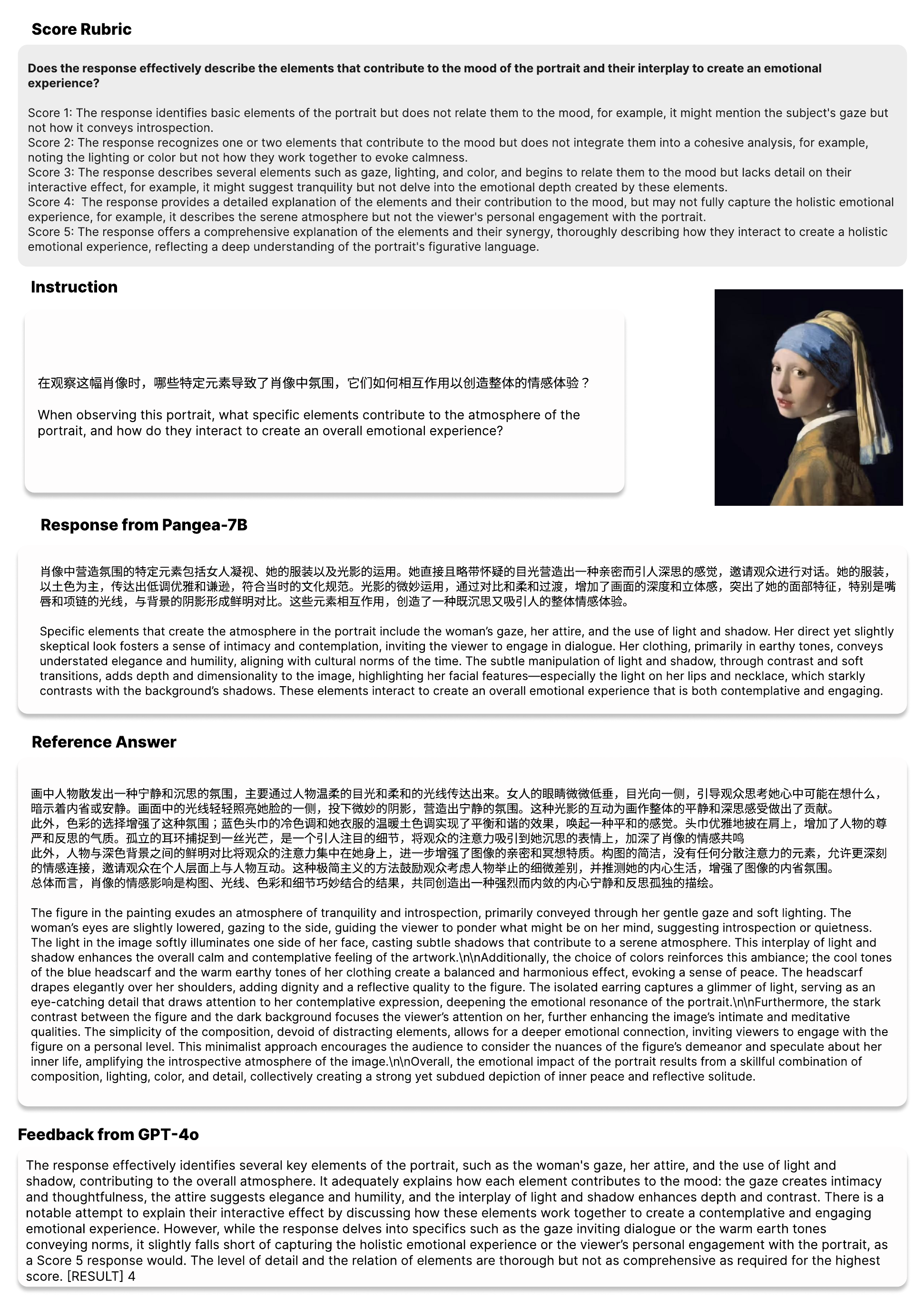}
    \caption{An xChat example from the Chinese subset and our model \model's response. The English part of the instruction, response, and reference answer is additionally added only on the figure to help to understand and was not given nor generated by \model.}
    \label{fig:xchat_chinese}
\end{figure}

\begin{figure}[!t]
    \centering
    \includegraphics[width=\linewidth]{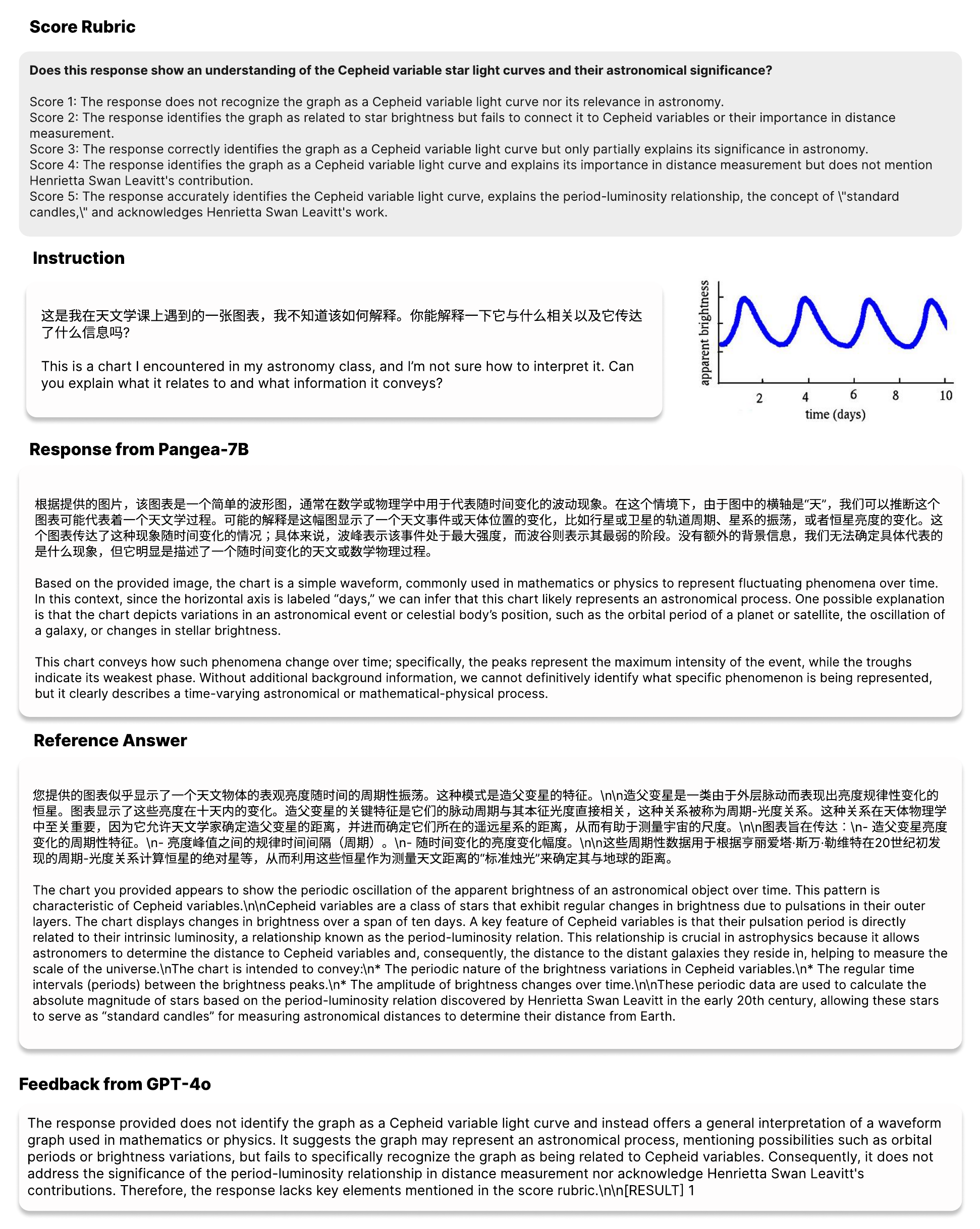}
    \caption{An xChat example from the Chinese subset and our model \model's response. The English part of the instruction, response, and reference answer is additionally added only on the figure to help to understand and was not given nor generated by \model.}
    \label{fig:xchat_chinese_negative}
\end{figure}

\begin{figure}[!t]
    \centering
    \includegraphics[width=\linewidth]{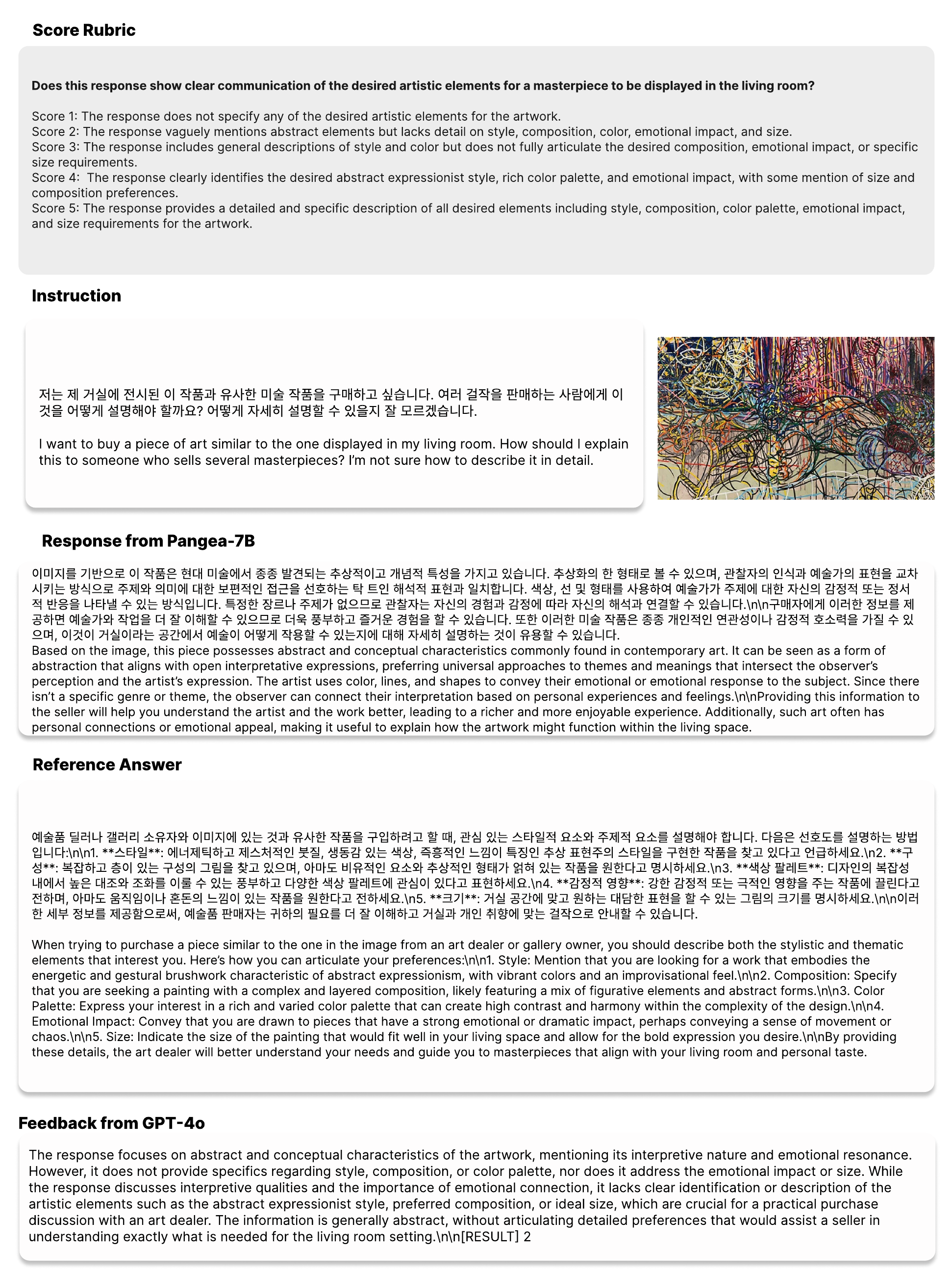}
    \caption{An xChat example from the Korean subset and our model \model's response. The English part of the instruction, response, and reference answer is additionally added only on the figure to help to understand and was not given nor generated by \model.}
    \label{fig:xchat_korean_negative}
\end{figure}

One important application of MLLMs is to answer users' queries in the wild. Here, we show the outputs of \model for the multimodal chat queries from our xChatBench. The examples included the scoring rubric, query, response from our \model, reference answer, and LLM-as-Judge feedback. As shown in Appendix \autoref{fig:xchat_japanese}, \ref{fig:xchat_hindi}, \ref{fig:xchat_korean}, \ref{fig:xchat_indonesian}, \ref{fig:xchat_spanish}, \ref{fig:xchat_chinese}, \model successfully interprets the figures in different tasks and generates fluent and readable in certain languages. These qualitative examples further demonstrate the remarkable visual understanding ability of \model in multilingual contexts. On the other hand, we also identified a few bad cases shown in \autoref{fig:xchat_chinese_negative}, \ref{fig:xchat_korean_negative}. Despite generating relevant responses to the queries, the model does not capture the key details of the images due to the lack of knowledge, which points out potential improvement directions in the future.


\begin{figure}[!h]
    \centering
    \includegraphics[width=\linewidth]{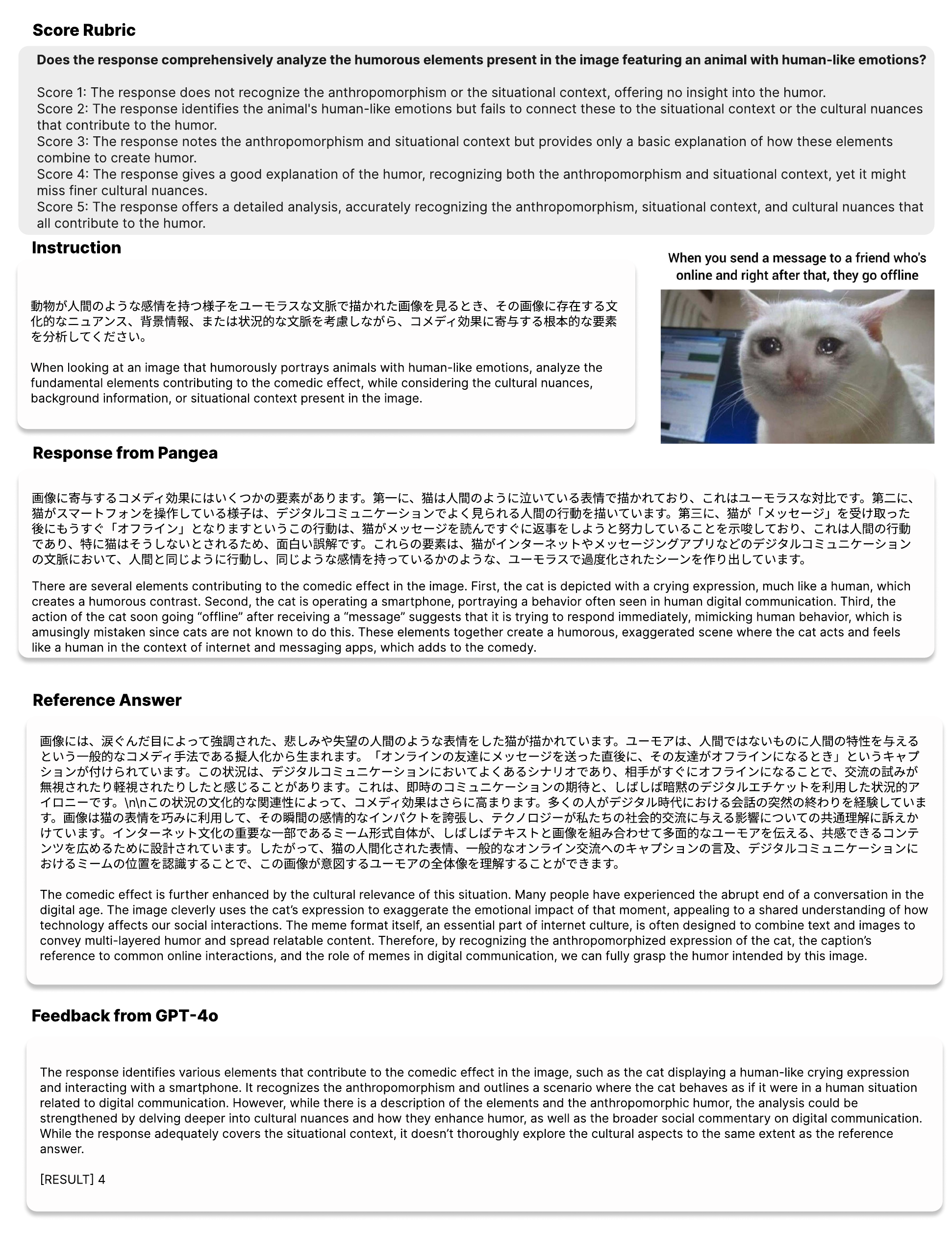}
    \caption{An xChat example from Japanese subset and our model \model's response. The English part of the instruction, response, and reference answer is additionally added only on the figure to help to understand and was not given nor generated by \model.}
    \label{fig:xchat_japanese}
\end{figure}

\clearpage

\begin{figure}[!t]
    \centering
    \includegraphics[width=\linewidth]{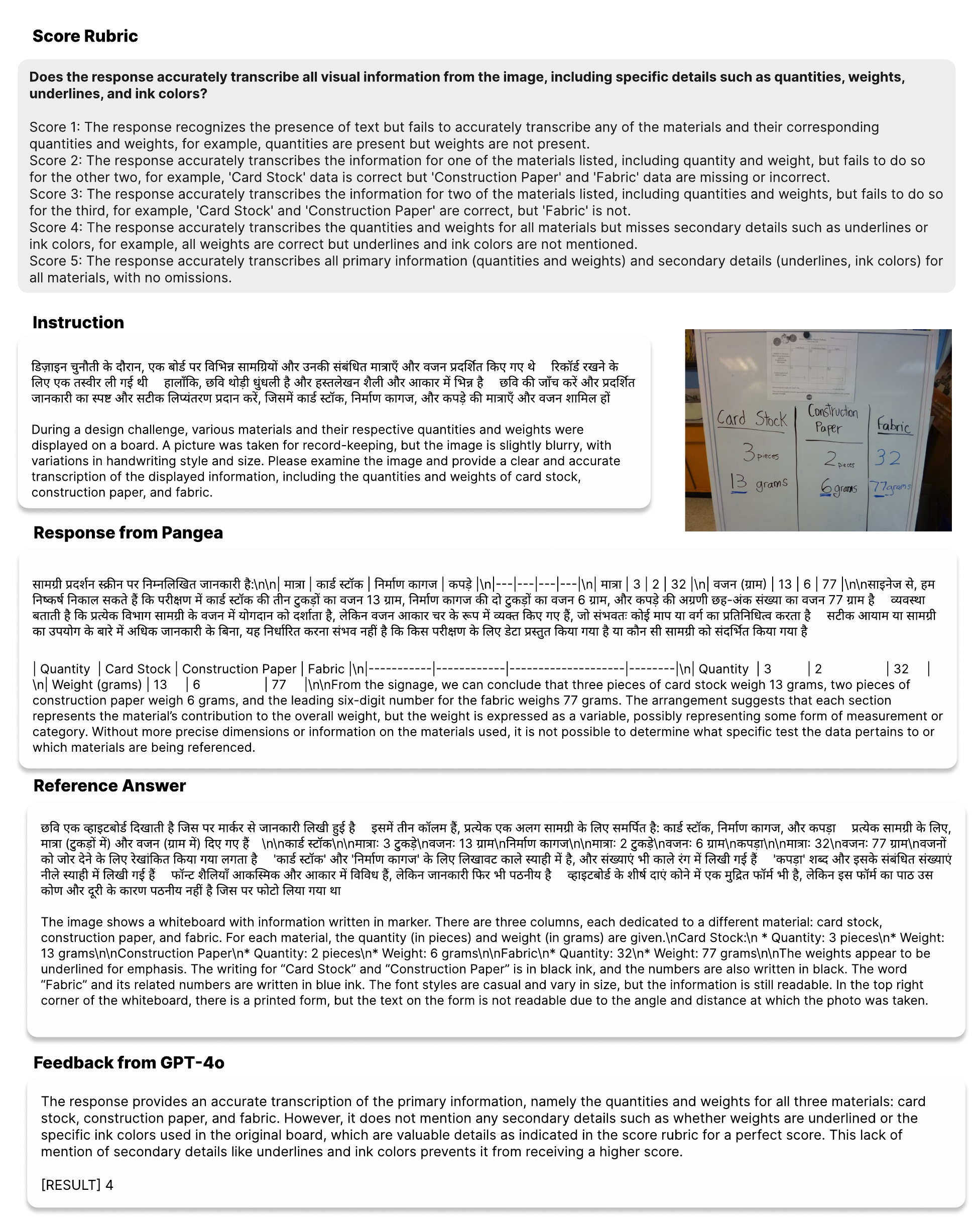}
    \caption{An xChat example from Hindi subset and our model \model's response. The English part of the instruction, response, and reference answer is additionally added only on the figure to help to understand and was not given nor generated by \model.}
    \label{fig:xchat_hindi}
\end{figure}

\clearpage

\clearpage

\clearpage

\clearpage

\clearpage

\clearpage

\clearpage
\section{Languages in \traindata}

\autoref{tab:languages} demonstrates the number of instances for each language that we include in \traindata.

\begin{table}[!h]
\centering
\resizebox{\textwidth}{!}{%
\begin{tabular}{l|cccccccccccccccccccc}
\toprule
\textbf{Languages} & \textbf{en} & \textbf{multi} & \textbf{am} & \textbf{ar} & \textbf{bg} & \textbf{bn} & \textbf{cs} & \textbf{de} & \textbf{el} & \textbf{es}\\
\midrule
\textbf{Count} & 2554.6 & 4389.5 & 31.7 & 162.8 & 52.7 & 118.4 & 4.6 & 132.2 & 7.3 & 126.6\\
\textbf{Percentage} (\%) & 36.8 & 63.2 & 0.5 & 2.3 & 0.8 & 1.7 & 0.1 & 1.9 & 0.1 & 1.8\\
\midrule
\textbf{Languages} & \textbf{fa} & \textbf{fr} & \textbf{ga} & \textbf{hi} & \textbf{id} & \textbf{ig} & \textbf{it} & \textbf{iw} & \textbf{ja} & \textbf{jv}\\
\midrule
\textbf{Count} & 8.4 & 193.8 & 34.7 & 163.5 & 178.4 & 26.1 & 68.8 & 168.3 & 403.3 & 35.2\\
\textbf{Percentage (\%)} & 0.1 & 2.8 & 0.5 & 2.4 & 2.6 & 0.4 & 1.0 & 2.4 & 5.8 & 0.5\\
\midrule
\textbf{Languages} & \textbf{ko} & \textbf{nl} & \textbf{mn} & \textbf{ms} & \textbf{no} & \textbf{pl} & \textbf{pt} & \textbf{ro} & \textbf{ru} & \textbf{si}\\
\midrule
\textbf{Count} & 182.5 & 4.5 & 37.6 & 39.2 & 60.7 & 8.2 & 176.5 & 147.2 & 211.0 & 0.6\\
\textbf{Percentage (\%)} & 2.6 & 0.1 & 0.5 & 0.6 & 0.9 & 0.1 & 2.5 & 2.1 & 3.0 & 0.1\\
\midrule
\textbf{Languages} & \textbf{su} & \textbf{sw} & \textbf{ta} & \textbf{te} & \textbf{th} & \textbf{tr} & \textbf{uk} & \textbf{ur} & \textbf{vi} & \textbf{zh}\\
\midrule
\textbf{Count} & 33.0 & 132.7 & 37.1 & 64.6 & 160.4 & 140.7 & 7.4 & 156.4 & 315.5 & 557.4\\
\textbf{Percentage (\%)} & 0.5 & 1.9 & 0.5 & 0.9 & 2.3 & 2.0 & 0.1 & 2.3 & 4.5 & 8.0\\
\bottomrule
\end{tabular}%
}
\caption{Language distribution of \traindata. We demonstrate the number of instances (in thousands) for each language, and the percentage of instances for each language in \traindata}
\label{tab:languages}
\end{table}
\newpage
\section{Training Examples}
\label{sec:train_examples}

\subsection{Machine Translated Instructions}
\begin{figure}[!h]
    \centering
    \includegraphics[width=1.0\linewidth]{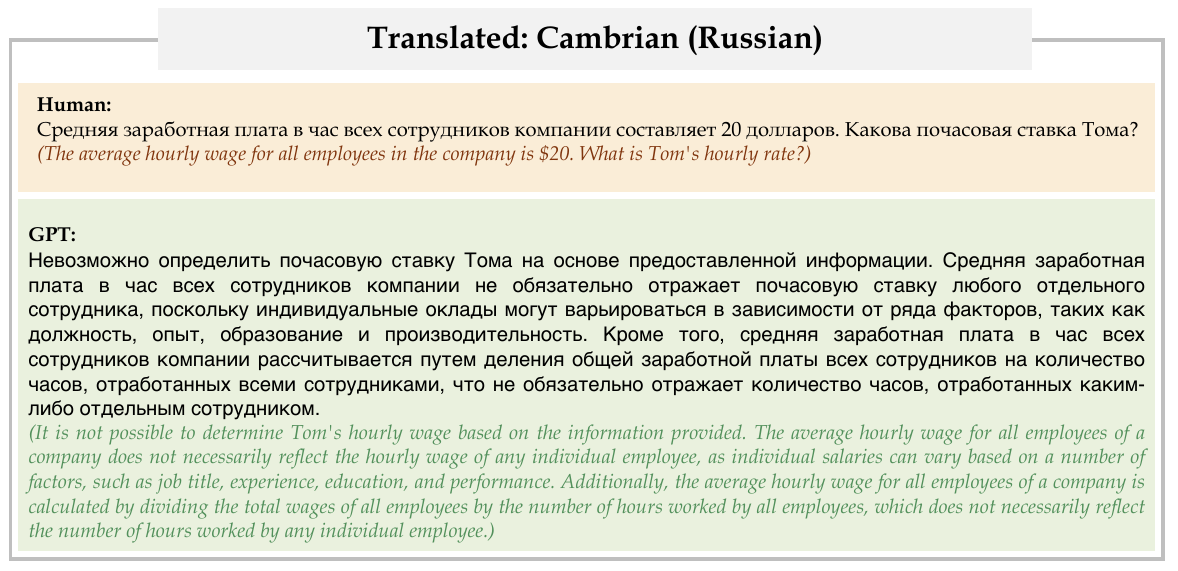}
    \caption{
        \textbf{\textit{Translated Task}}: An example from the Cambrian dataset where it discusses the concept of hourly wages based on a given prompt in Russian.
    }
    \label{fig:translated_task_russian}
\end{figure}

\begin{figure}[!h]
    \centering
    \includegraphics[width=1.0\linewidth]{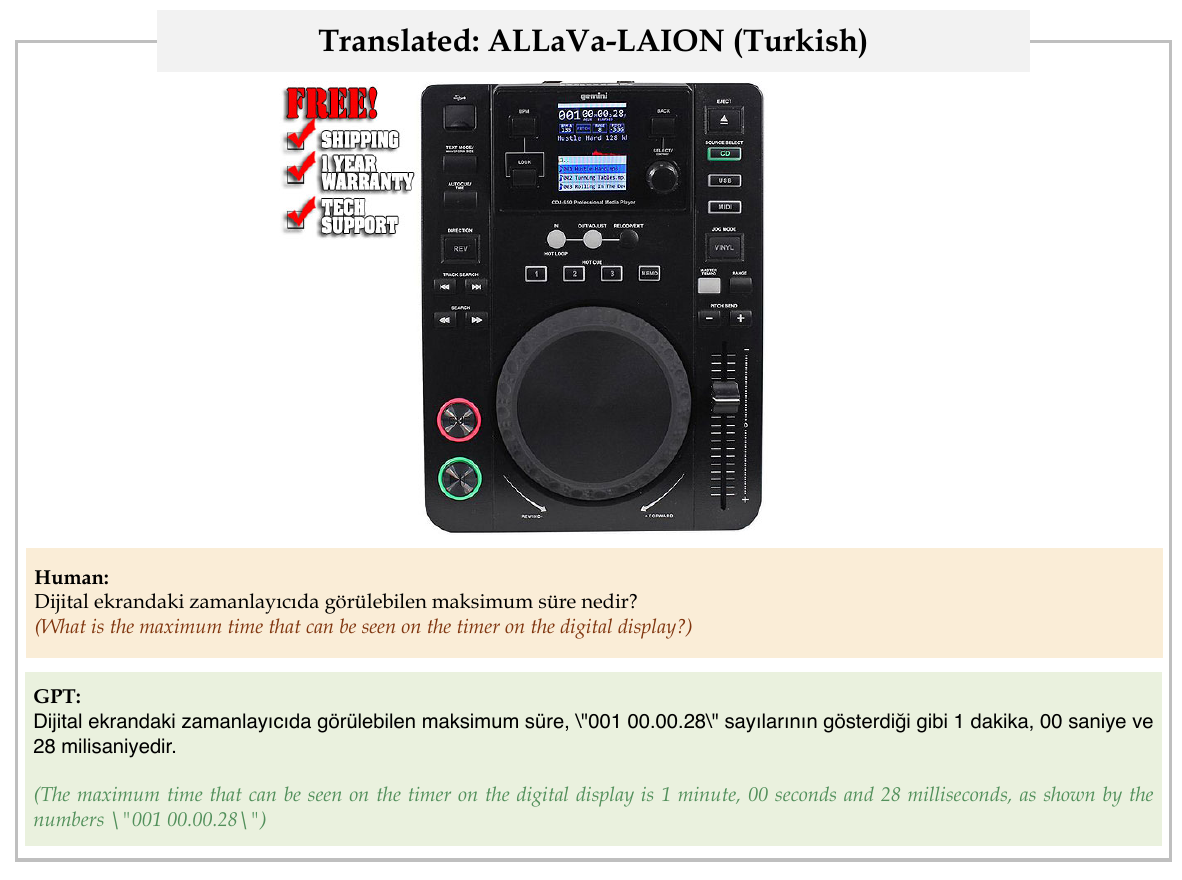}
    \caption{
        \textbf{\textit{Multimodal Translated Task}}: An example from the ALLaVa-LAION dataset where the GPT model answers a prompt in Turkish regarding the maximum time displayed on a digital timer.
    }
    \label{fig:translated_task_turkish}
\end{figure}

\begin{figure}[!h]
    \centering
    \includegraphics[width=1.0\linewidth]{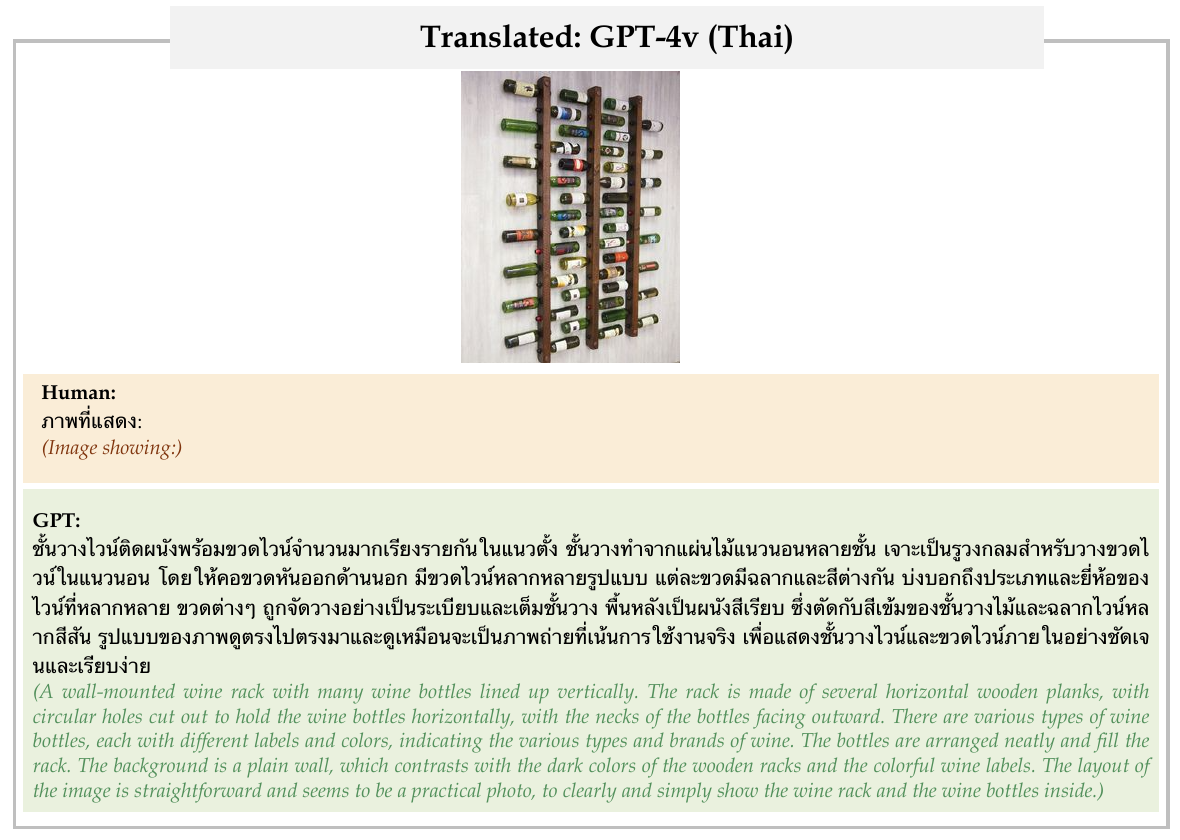}
    \caption{
        \textbf{\textit{Multimodal Translated Task}}: An example from the ShareGPT-4v dataset where the model describes an image of a wine rack in Thai, detailing its structure and the arrangement of wine bottles.
    }
    \label{fig:translated_task_thai}
\end{figure}

\begin{figure}[!h]
    \centering
    \includegraphics[width=\linewidth]{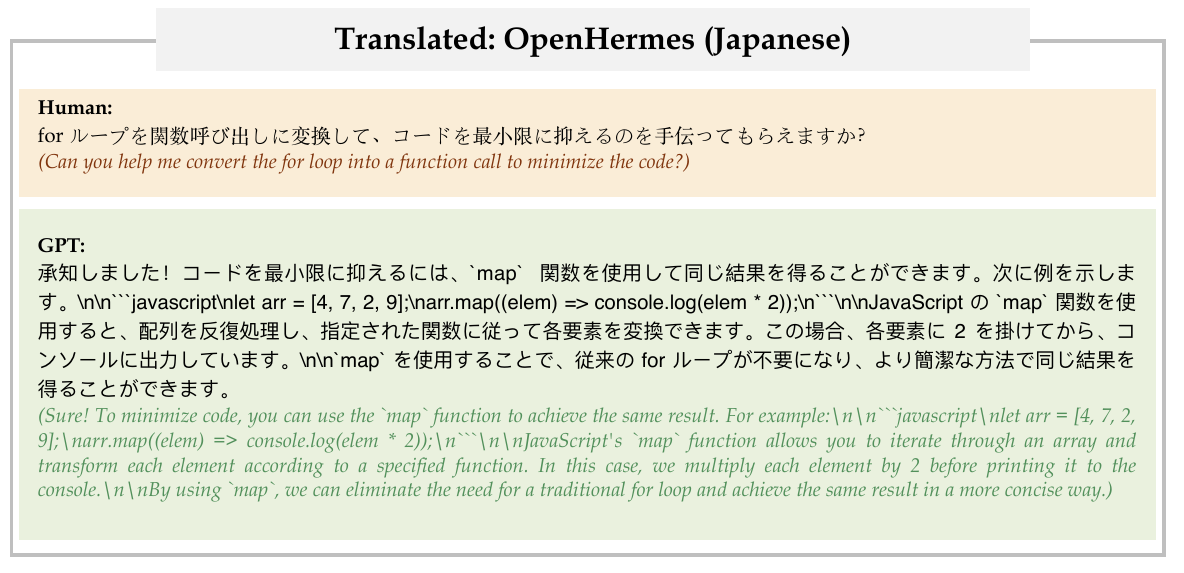}
    \caption{
        \textbf{\textit{Text-only Translated Task}}: An example from the OpenHermes2.5 dataset translated into Japanese.
    }
    \label{fig:translated_task_jp}
\end{figure}

\begin{figure}[!h]
    \centering
    \includegraphics[width=\linewidth]{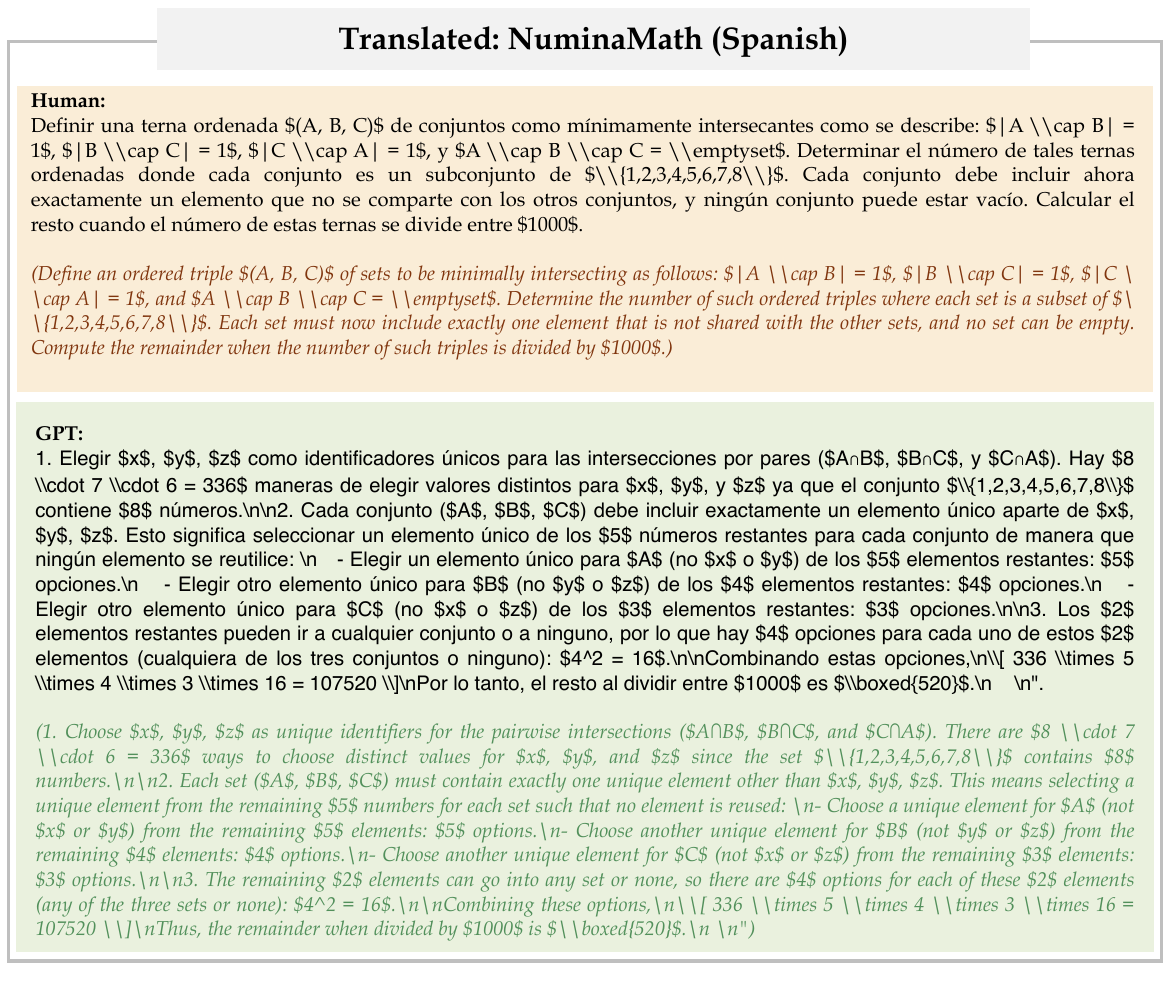}
    \caption{
        \textbf{\textit{Text-only Translated Task}}: An example from the NumininaMath dataset translated into Spanish.
    }
    \label{fig:translated_task_spanish}
\end{figure}

\begin{figure}[!h]
    \centering
    \includegraphics[width=\linewidth]{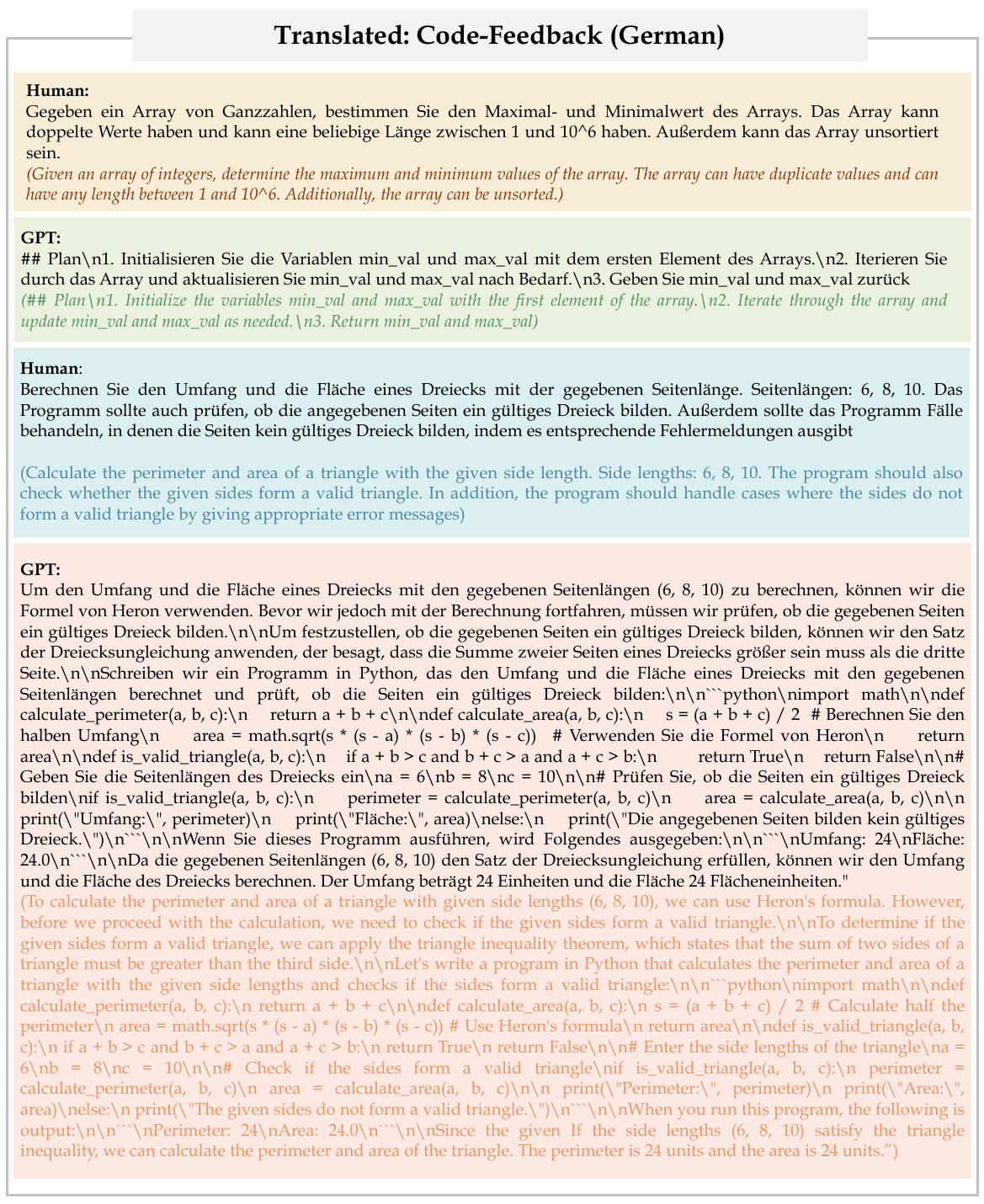}
    \caption{
        \textbf{\textit{Text-only Translated Task}}: An example from the Code-Feedback dataset translated into German.
    }
    \label{fig:translated_task_german}
\end{figure}

We include a few machine-translated training examples from \traindata in \autoref{fig:translated_task_russian},   \autoref{fig:translated_task_turkish},   \autoref{fig:translated_task_thai},  
\autoref{fig:translated_task_jp},  
\autoref{fig:translated_task_spanish},   \autoref{fig:translated_task_german}.

\subsection{Multicultural Understanding Instructions}
\begin{figure}[!h]
    \centering
    \includegraphics[width=1.0\linewidth]{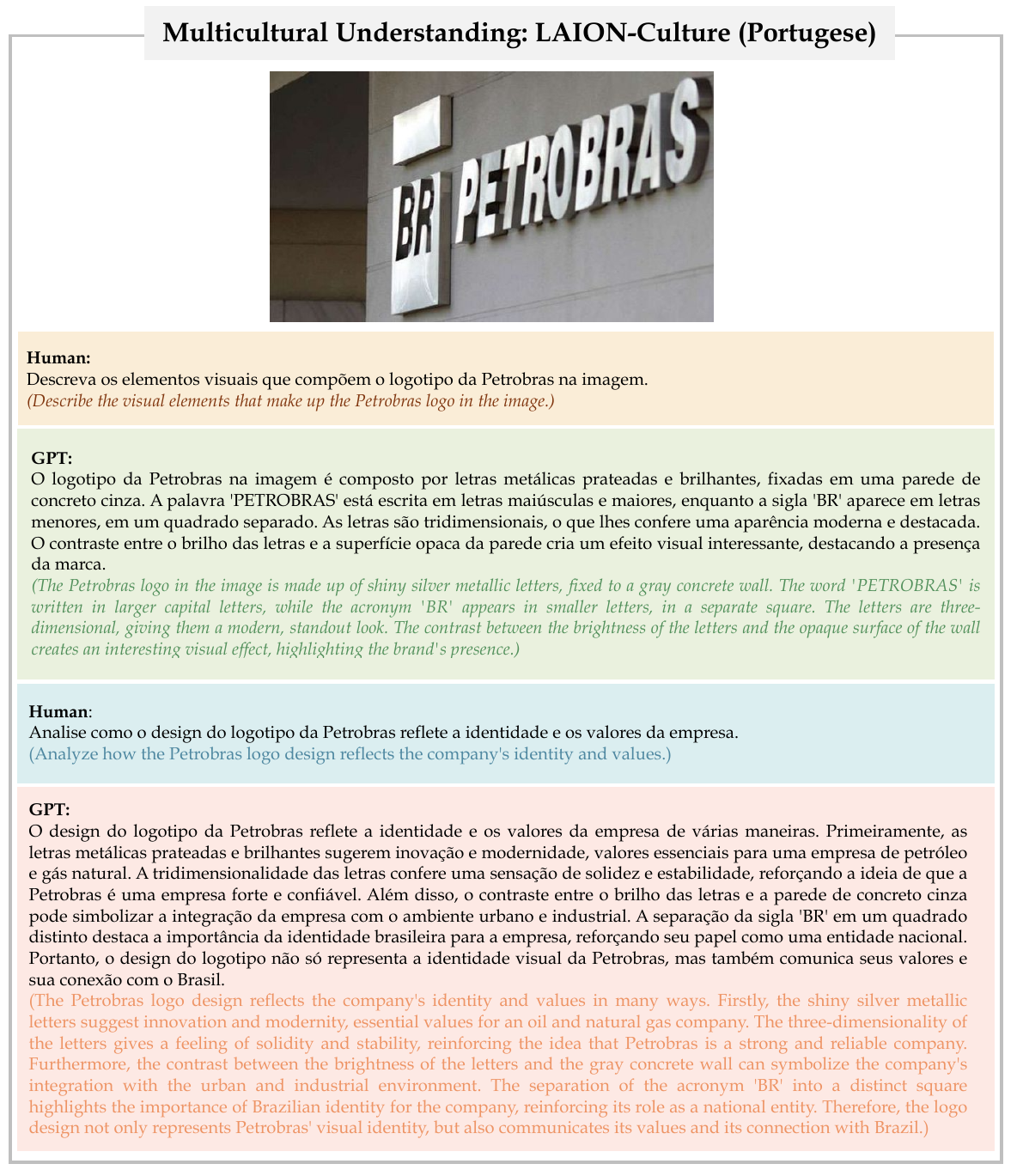}
    \caption{\textbf{\textit{Multicultural Understanding}}: An example from the LAION-Culture dataset where the GPT model describes and analyzes the Petrobras logo in Portuguese.}
    \label{fig:multilingual_task_portugese}
\end{figure}

\begin{figure}[!h]
    \centering
    \includegraphics[width=1.0\linewidth]{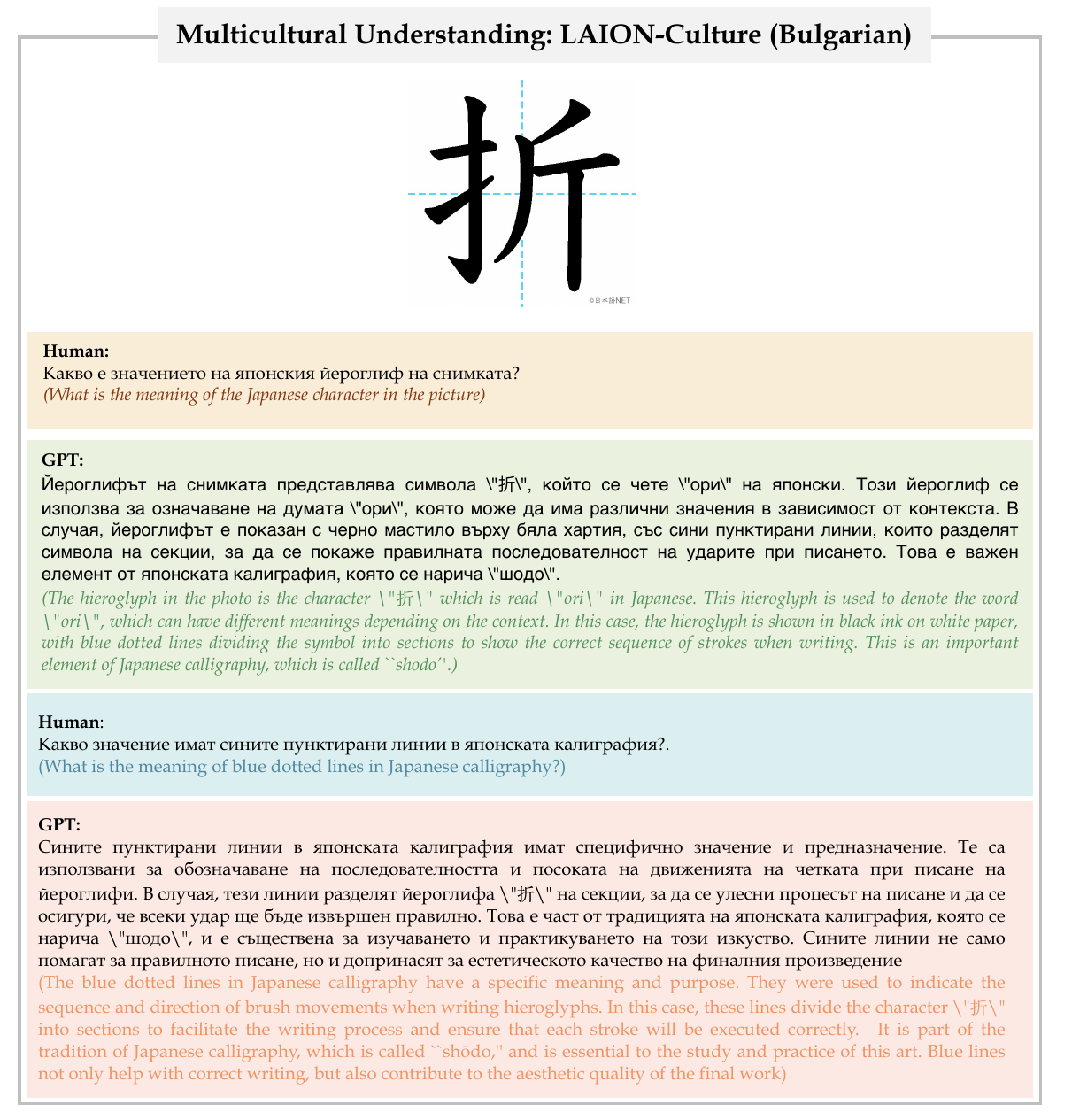}
    \caption{
        \textbf{\textit{Multicultural Understanding}}: Example from the LAION-Culture dataset where the model interprets the meaning of a Japanese character and its representation in calligraphy, described in Bulgarian.
    }
    \label{fig:multilingual_task_bulgarian}
\end{figure}

\begin{figure}[!h]
    \centering
    \includegraphics[width=1.0\linewidth]{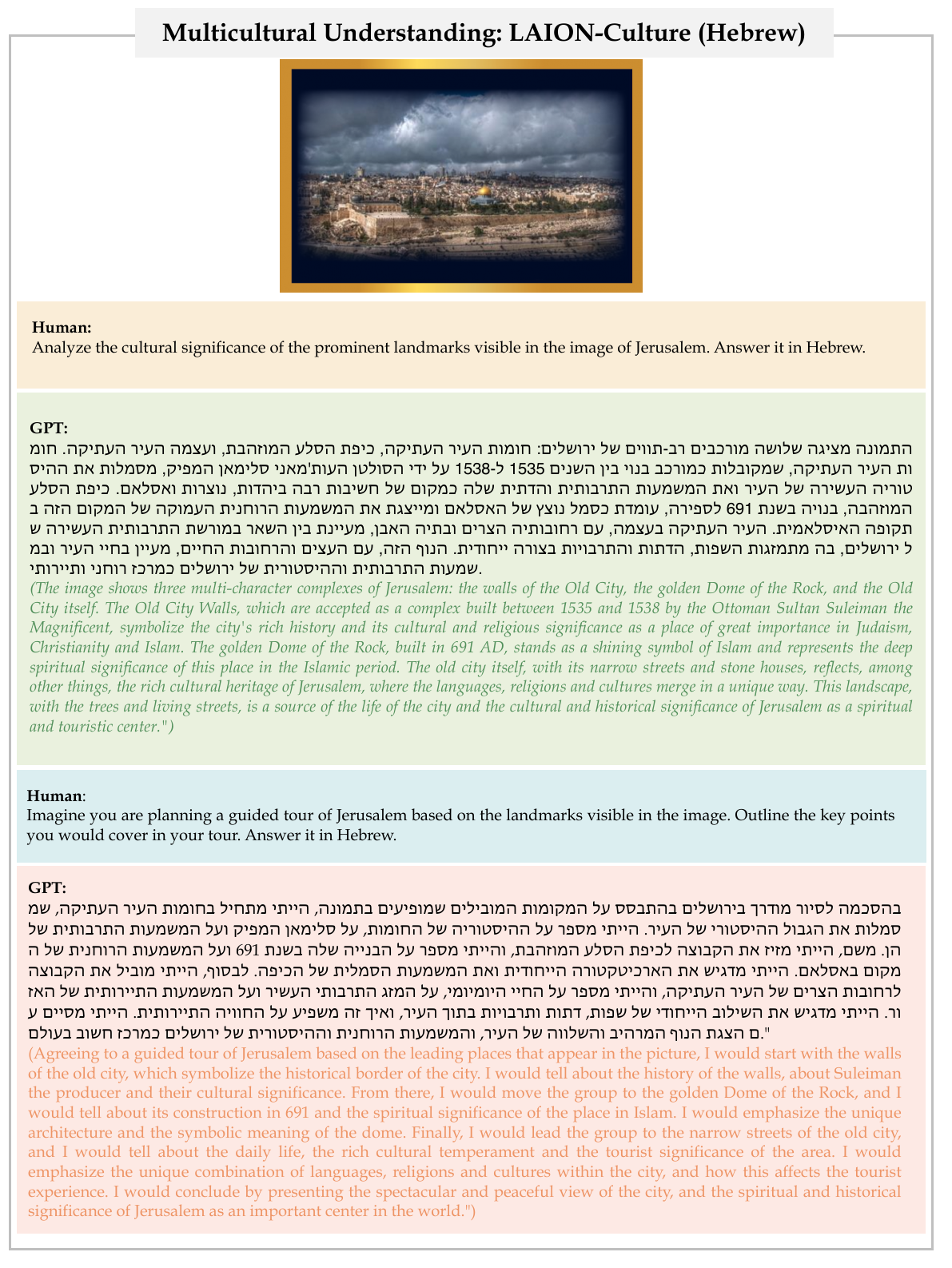}
    \caption{
     \textbf{\textit{Multicultural Understanding}}: LAION-Culture sample in Hebrew.
    }
    \label{fig:multicultural_task_hebrew}
\end{figure}

\begin{figure}[!h]
    \centering
    \includegraphics[width=1.0\linewidth]{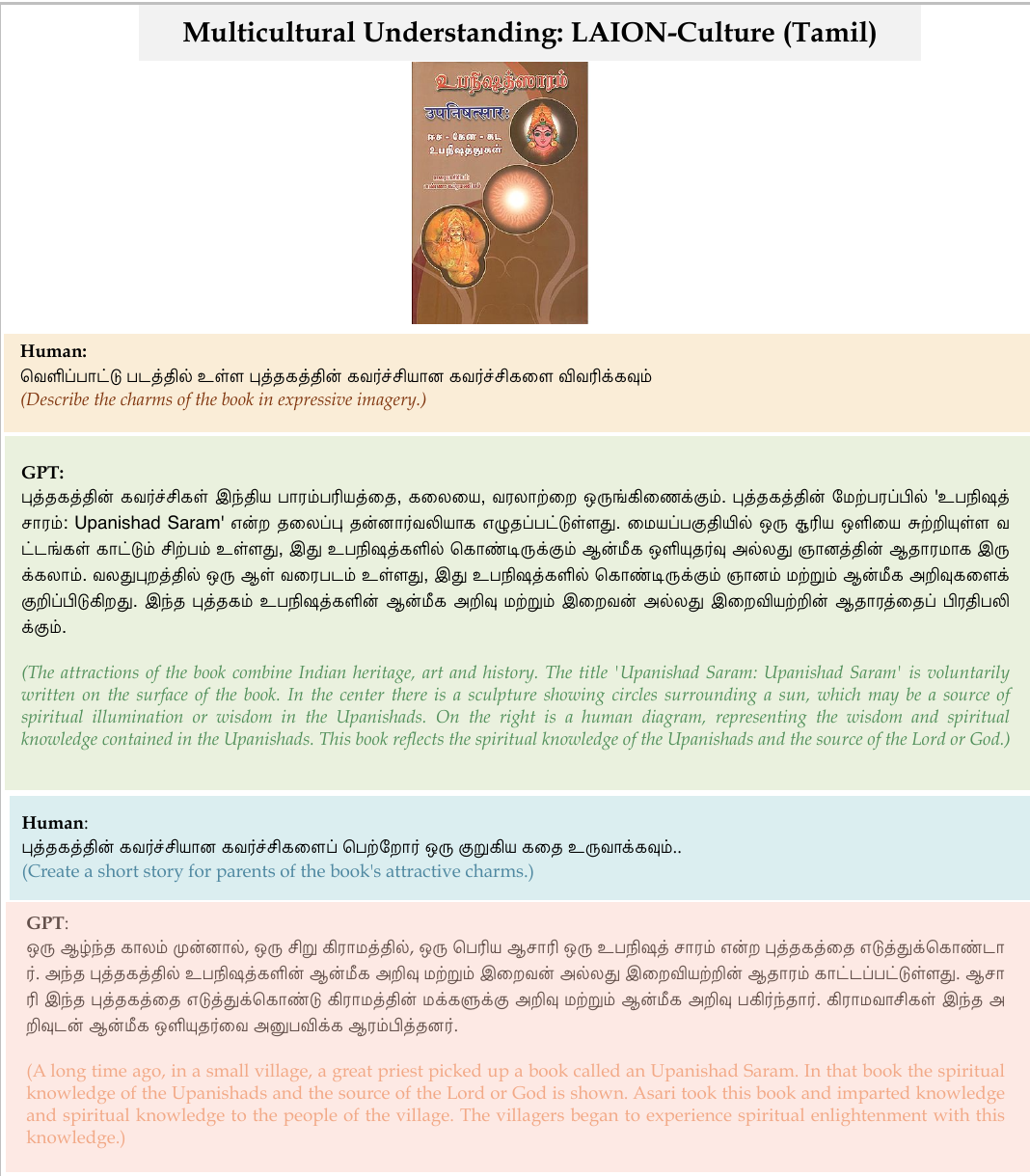}
    \caption{
     \textbf{\textit{Multicultural Understanding}}: LAION-Culture sample in Tamil.
    }
    \label{fig:multicultural_task_tamil}
\end{figure}

We include a few multicultural understanding instructions from \traindata in \autoref{fig:multilingual_task_portugese}, \autoref{fig:multilingual_task_bulgarian}, \autoref{fig:multicultural_task_hebrew}, \autoref{fig:multicultural_task_tamil}.

\clearpage

\clearpage

\clearpage

\clearpage

\clearpage


\clearpage

\newpage
\section{Breakdown Results of Different Languages on \evaldata}

\subsection{xChat}
We show the performance of different models on the xChat benchmark in \autoref{tab:breakdown:xChat}.
\begin{table}[!h]
\centering
\resizebox{\textwidth}{!}{%
\begin{tabular}{l|rr|rrrrrr}
\toprule
\textbf{Models} & \textbf{English} & \textbf{Multi} & \textbf{Spanish} & \textbf{Hindi} & \textbf{Indonesian} & \textbf{Japanese} & \textbf{Korean} & \textbf{Chinese} \\
\midrule
Gemini-1.5-Pro & 71.0 & 65.6 & 66.0 & 62.0 & 65.5 & 68.0 & 66.5 & 65.5 \\
GPT4o & 67.0 & 65.1 & 66.0 & 64.0 & 65.0 & 66.5 & 67.5 & 61.5 \\\midrule
Llava-1.5-7B & 22.5 & 16.7 & 22.5 & 3.5 & 18.0 & 23.0 & 12.0 & 21.0 \\
Llava-Next-7B & 40.5 & 20.4 & 33.0 & 1.5 & 19.0 & 25.0 & 15.0 & 29.0 \\
Phi-3.5-Vision & 38.5 & 21.1 & 37.0 & 11.5 & 10.5 & 31.0 & 12.5 & 24.0 \\
Cambrian-8B & 27.5 & 15.8 & 22.5 & 4.0 & 20.0 & 20.0 & 10.5 & 18.0 \\
Llava-OV-7B & 51.0 & 33.1 & 45.5 & 6.5 & 42.0 & 36.5 & 26.0 & 42.0 \\
Molmo-7B-D & 49.5 & 34.7 & 45.0 & 19.5 & 36.5 & 36.0 & 35.0 & 46.0 \\
Llama3.2-11B & 49.0 & 31.3 & 42.5 & 19.5 & 45.0 & 26.0 & 21.0 & 43.0 \\
PaliGemma-3B & 6.0 & 3.8 & 4.5 & 0.5 & 6.5 & 6.5 & 2.0 & 3.0 \\
PALO-7B & 27.0 & 16.2 & 23.0 & 3.0 & 19.0 & 20.0 & 13.5 & 18.5 \\
mBLIP mT0-XL & 2.5 & 0.5 & 0.0 & 0.0 & 0.5 & 2.0 & 0.5 & 0.0 \\
mBLIP BLOOMZ-7B & 4.0 & 1.7 & 2.0 & 2.5 & 2.5 & 0.0 & 0.0 & 3.0 \\\midrule
\model-7B (Ours) & 46.0 & 35.8 & 43.5 & 23.5 & 34.5 & 39.0 & 33.5 & 40.5 \\
\bottomrule
\end{tabular}%
}
\caption{Comparison of models on the xChat dataset across different languages.}
\label{tab:breakdown:xChat}
\end{table}

\subsection{Multilingual LLaVABench}
We show the performance of different models on the Multilingual LLaVABench benchmark in \autoref{tab:breakdown:MM_LMM}.
\begin{table}[!h]
\centering
\resizebox{\textwidth}{!}{%
\begin{tabular}{l|rr|rrrrrrrrr}
\toprule
\textbf{Models} & \textbf{English} & \textbf{Multi} & \textbf{Arabic} & \textbf{Bengali} & \textbf{Chinese} & \textbf{French} & \textbf{Hindi} & \textbf{Japanese} & \textbf{Russian} & \textbf{Spanish} & \textbf{Urdu} \\
\midrule
Gemini-1.5-Pro & 103.4 & 106.6 & 112.9 & 117.1 & 104.1 & 115.5 & 106.2 & 118.1 & 95.7 & 88.2 & 101.6 \\
GPT4o & 104.6 & 100.4 & 98.3 & 111.9 & 96.5 & 101.1 & 99.7 & 104.0 & 88.5 & 100.9 & 102.5  \\\midrule
Llava-1.5-7B & 66.1 & 40.8 & 26.4 & 11.9 & 50.7 & 63.8 & 23.2 & 70.0 & 46.5 & 59.2 & 15.4 \\
Llava-Next-7B & 78.9 & 50.7 & 24.9 & 11.2 & 72.8 & 91.4 & 18.0 & 70.1 & 71.8 & 82.9 & 13.4 \\
Phi-3.5-Vision & 70.8 & 58.0 & 50.1 & 35.1 & 69.2 & 86.0 & 35.9 & 63.0 & 67.6 & 75.6 & 39.3 \\
Cambrian-8B & 78.4 & 61.8 & 54.1 & 35.4 & 80.9 & 87.3 & 44.2 & 64.4 & 76.4 & 90.3 & 23.3 \\
Llava-OV-7B & 89.7 & 55.3 & 45.5 & 33.8 & 90.0 & 89.4 & 35.3 & 70.3 & 44.7 & 75.5 & 13.3 \\
Molmo-7B-D & 95.9 & 13.8 & 10.1 & 4.2 & 0.3 & 59.6 & 5.5 & 6.0 & 8.7 & 29.5 & 0.0 \\
Llama3.2-11B & 93.9 & 58.2 & 39.4 & 48.1 & 47.2 & 85.6 & 67.8 & 53.7 & 68.5 & 77.8 & 35.3 \\
PaliGemma-3B & 32.1 & 31.9 & 37.3 & 38.2 & 29.1 & 30.0 & 35.8 & 33.4 & 26.1 & 32.3 & 25.1 \\
PALO-7B & 68.9 & 71.2 & 79.1 & 54.6 & 71.5 & 83.9 & 61.9 & 66.6 & 80.9 & 74.4 & 68.2 \\
mBLIP mT0-XL & 32.7 & 28.2 & 33.7 & 26.2 & 3.6 & 39.8 & 26.9 & 26.8 & 34.1 & 36.9 & 26.0 \\
mBLIP BLOOMZ-7B & 43.5 & 41.0 & 48.1 & 44.1 & 30.6 & 53.3 & 39.1 & 29.8 & 38.1 & 51.5 & 34.0 \\
\midrule
\model-7B (Ours) & 84.2 & 89.5 & 91.0 & 94.9 & 94.4 & 93.8 & 84.9 & 92.8 & 91.2 & 87.4 & 75.5 \\
\bottomrule
\end{tabular}%
}
\caption{Comparison of models on the Multilingual LLaVABench benchmark across different languages.}
\label{tab:breakdown:MM_LMM}
\end{table}

\subsection{CVQA}
We show the performance of different models on the CVQA benchmark in \autoref{tab:breakdown:CVQA} and \autoref{tab:cvqa_en}.
\begin{table}[!h]
\centering
\resizebox{\textwidth}{!}{%
\begin{tabular}{l|cccccccccc}
\toprule
\textbf{Models} & \textbf{ar-es} & \textbf{br-pt} & \textbf{bu-bg} & \textbf{ch-es} & \textbf{ch-zh} & \textbf{co-es} & \textbf{ec-es} & \textbf{eg-ar} & \textbf{et-am} & \textbf{et-or} \\
\midrule
Llava-1.5-7B & 37.8 & 51.1 & 35.6 & 42.4 & 44.4 & 50.6 & 48.6 & 31.5 & 27.8 & 31.8\\
Llava-Next-7B & 52.5 & 62.3 & 41.5 & 59.0 & 51.1 & 54.8 & 50.8 & 33.5 & 29.5 & 36.9\\
Phi-3.5-Vision & 54.0 & 57.2 & 36.9 & 57.7 & 51.1 & 52.3 & 50.1 & 38.4 & 27.8 & 32.2\\
Cambrian-8B & 59.6 & 60.6 & 42.0 & 64.5 & 59.5 & 57.7 & 56.1 & 40.9 & 27.8 & 25.7\\
Llava-OV-7B & 64.5 & 69.7 & 49.6 & 67.1 & 69.1 & 66.8 & 65.5 & 47.8 & 32.5 & 41.1\\
Molmo-7B-D & 61.1 & 69.0 & 54.9 & 60.7 & 66.2 & 58.5 & 54.9 & 56.7 & 58.1 & 60.7\\
Llama3.2-11B & 69.1 & 74.6 & 64.2 & 70.5 & 73.6 & 69.3 & 66.9 & 68.5 & 68.4 & 63.1\\
PaliGemma-3B & 48.7 & 53.9 & 39.1 & 53.4 & 53.7 & 50.6 & 45.3 & 40.4 & 24.8 & 28.0\\
PALO-7B & 50.9 & 56.7 & 36.7 & 55.1 & 45.3 & 48.5 & 46.4 & 28.6 & 19.2 & 32.7\\
mBLIP BLOOMZ-7B & 45.3 & 51.4 & 30.5 & 45.3 & 51.1 & 46.9 & 44.8 & 35.9 & 23.9 & 25.7\\
mBLIP mT0-XL & 40.8 & 44.4 & 38.0 & 44.9 & 39.9 & 41.9 & 42.5 & 31.0 & 35.9 & 26.6\\
\model-7B (Ours) & 68.3 & 72.9 & 53.9 & 70.5 & 74.0 & 64.7 & 63.5 & 49.3 & 36.3 & 35.5\\
\midrule
\textbf{Models} & \textbf{fr-br} & \textbf{in-bn} & \textbf{in-ta} & \textbf{in-te} & \textbf{ind-id} & \textbf{ind-jv} & \textbf{ind-mi} & \textbf{ind-sv} & \textbf{ir-ir} & \textbf{ja-jp} \\
\midrule
Llava-1.5-7B & 29.4 & 31.1 & 29.8 & 28.0 & 41.7 & 32.0 & 32.7 & 33.5 & 42.6 & 37.4\\
Llava-Next-7B & 27.4 & 31.1 & 28.8 & 28.0 & 42.2 & 38.7 & 40.2 & 35.5 & 42.6 & 32.5\\
Phi-3.5-Vision & 29.3 & 39.0 & 40.0 & 36.8 & 45.0 & 38.2 & 38.2 & 30.8 & 39.6 & 39.7 \\
Cambrian-8B & 31.6 & 47.2 & 38.1 & 44.0 & 50.2 & 43.8 & 39.4 & 45.5 & 47.9 & 40.9\\
Llava-OV-7B & 34.3 & 56.3 & 43.9 & 46.5 & 58.0 & 45.8 & 45.4 & 40.5 & 50.6 & 49.8\\
Molmo-7B-D & 44.2 & 61.9 & 61.2 & 58.5 & 52.9 & 53.9 & 54.6 & 55.0 & 64.4 & 42.9\\
Llama3.2-11B & 49.4 & 76.9 & 80.4 & 80.5 & 65.8 & 60.6 & 68.9 & 64.0 & 76.4 & 54.2\\
PaliGemma-3B & 29.9 & 46.2 & 46.0 & 43.5 & 45.4 & 41.4 & 39.8 & 33.0 & 34.4 & 43.3\\
PALO-7B & 29.1 & 37.8 & 31.2 & 25.0 & 41.3 & 32.3 & 32.3 & 32.0 & 42.9 & 30.5\\
mBLIP BLOOMZ-7B & 26.7 & 41.9 & 40.0 & 42.0 & 41.9 & 35.4 & 35.1 & 32.0 & 29.4 & 31.0\\
mBLIP mT0-XL & 23.5 & 36.4 & 44.2 & 39.0 & 37.4 & 37.4 & 34.7 & 31.0 & 35.3 & 30.0\\
\model-7B (Ours) & 34.6 & 59.1 & 51.9 & 54.5 & 62.1 & 49.5 & 47.8 & 53.0 & 56.4 & 48.3\\
\midrule
\textbf{Models} & \textbf{ke-sw} & \textbf{ma-my} & \textbf{me-es} & \textbf{mo-mg} & \textbf{ni-ig} & \textbf{no-ng} & \textbf{pk-ur} & \textbf{ph-fi} & \textbf{ro-ro} & \textbf{ru-ru} \\
\midrule
Llava-1.5-7B & 34.4 & 42.2 & 42.4 & 26.9 & 34.5 & 47.5 & 26.4 & 43.8 & 47.0 & 51.0\\
Llava-Next-7B & 46.2 & 45.7 & 51.4 & 33.3 & 35.0 & 56.9 & 36.6 & 46.8 & 52.3 & 53.5\\
Phi-3.5-Vision & 46.0 & 45.1 & 46.3 & 31.9 & 33.3 & 50.0 & 35.2 & 41.4 & 47.4 & 50.5\\
Cambrian-8B & 50.5 & 52.1 & 56.7 & 34.6 & 36.0 & 53.5 & 48.6 & 47.3 & 52.0 & 61.5\\
Llava-OV-7B & 46.5 & 55.6 & 59.4 & 35.9 & 33.5 & 62.5 & 58.3 & 56.2 & 60.3 & 75.5\\
Molmo-7B-D & 73.3 & 54.6 & 53.6 & 51.9 & 53.0 & 54.8 & 67.1 & 57.6 & 63.6 & 61.5\\
Llama3.2-11B & 79.1 & 72.1 & 66.6 & 54.5 & 61.5 & 66.9 & 78.7 & 70.0 & 76.8 & 74.5\\
PaliGemma-3B & 44.0 & 44.1 & 47.4 & 29.2 & 32.0 & 52.2 & 44.9 & 39.9 & 50.3 & 53.5\\
PALO-7B & 35.9 & 42.5 & 44.3 & 28.8 & 29.5 & 49.2 & 44.4 & 39.4 & 46.0 & 47.0\\
mBLIP BLOOMZ-7B & 37.0 & 42.5 & 44.8 & 28.8 & 33.0 & 49.2 & 47.7 & 31.5 & 46.0 & 34.0\\
mBLIP mT0-XL & 45.1 & 40.6 & 44.9 & 29.2 & 30.5 & 42.8 & 40.3 & 32.0 & 43.7 & 42.0\\
\model-7B (Ours) & 64.1 & 59.7 & 62.2 & 42.3 & 46.0 & 64.5 & 66.2 & 58.6 & 64.6 & 74.0\\
\midrule
\textbf{Models} & \textbf{rw-ki} & \textbf{sg-zh} & \textbf{sk-ko} & \textbf{sp-es} & \textbf{sr-si} & \textbf{ur-es} & \textbf{macro} & & & \\
\midrule
Llava-1.5-7B & 31.1 & 44.3 & 44.5 & 56.9 & 24.9 & 37.8 & 38.7 & & &\\
Llava-Next-7B & 34.5 & 44.8 & 43.4 & 63.5 & 29.8 & 41.0 & 42.6 & & &\\
Phi-3.5-Vision & 31.1 & 43.9 & 55.2 & 62.4 & 28.0 & 43.3 & 42.4 & & &\\
Cambrian-8B & 31.9 & 54.7 & 54.5 & 70.4 & 36.4 & 45.7 & 47.5 & & &\\
Llava-OV-7B & 35.3 & 70.3 & 65.2 & 79.9 & 31.6 & 47.3 & 53.8\\
Molmo-7B-D & 57.4 & 69.3 & 65.2 & 70.1 & 68.0 & 50.8 & 59.4 & & & \\
Llama3.2-11B & 57.9 & 80.7 & 73.8 & 81.4 & 72.4 & 52.4 & 70.1 & & & \\
PaliGemma-3B & 27.2 & 48.6 & 61.0 & 60.1 & 31.6 & 39.4 & 43.0 & & &\\
PALO-7B & 28.9 & 45.8 & 44.5 & 64.8 & 28.0 & 39.4 & 39.3 & & &\\
mBLIP BLOOMZ-7B & 29.4 & 47.6 & 33.1 & 56.6 & 28.0 & 39.4 & 36.9 & & &\\
mBLIP mT0-XL & 33.2 & 36.8 & 38.3 & 53.5 & 31.1 & 39.1 & 37.6 & & &\\
\model-7B (Ours) & 35.7 & 65.6 & 70.7 & 72.6 & 39.1 & 49.8 & 57.2 & & &\\
\bottomrule
\end{tabular}%
}
\caption{Comparison of models on CVQA across different country-language pairs (in local languages). Includes Macro-Acc.}
\label{tab:breakdown:CVQA}
\end{table}


\begin{table}[htbp]
\centering
\resizebox{\textwidth}{!}{%
\begin{tabular}{l|cccccccccc}
\toprule
\textbf{Models} & \textbf{ar-es} & \textbf{br-pt} & \textbf{bu-bg} & \textbf{ch-es} & \textbf{ch-zh} & \textbf{co-es} & \textbf{ec-es} & \textbf{eg-ar} & \textbf{et-am} & \textbf{et-or} \\
\midrule
Llava-1.5-7B & 56.2 & 61.6 & 52.3 & 60.2 & 54.0 & 55.6 & 55.5 & 50.2 & 51.3 & 53.3\\
Llava-Next-7B & 53.9 & 61.3 & 50.9 & 59.8 & 58.8 & 60.2 & 52.8 & 54.7 & 52.9 & 58.9\\
Phi-3.5-Vision & 59.2 & 61.9 & 54.9 & 64.1 & 58.2 & 59.3 & 57.5 & 50.7 & 54.7 & 58.4\\
Cambrian-8B & 57.7 & 66.5 & 56.1 & 65.4 & 64.3 & 59.3 & 60.2 & 56.7 & 60.3 & 56.5\\
Llava-OV-7B & 63.0 & 73.9 & 59.3 & 65.8 & 68.8 & 65.1 & 63.3 & 62.1 & 59.8 & 59.3 \\
Molmo-7B-D & 57.7 & 65.8 & 45.6 & 63.7 & 68.5 & 57.3 & 55.0 & 43.8 & 31.6 & 38.8\\
Llama3.2-11B & 66.8 & 72.9 & 54.4 & 72.6 & 72.0 & 66.4 & 65.2 & 56.7 & 41.9 & 32.2\\
PaliGemma-3B & 51.7 & 59.5 & 49.3 & 51.7 & 54.9 & 54.8 & 47.2 & 51.2 & 52.6 & 51.4\\
PALO-7B & 50.2 & 57.0 & 48.8 & 53.4 & 52.1 & 51.9 & 53.0 & 48.3 & 47.0 & 52.3\\
mBLIP mT0-XL & 38.1 & 45.4 & 39.1 & 42.7 & 43.7 & 41.1 & 40.9 & 42.9 & 34.2 & 42.1\\
mBLIP BLOOMZ-7B & 46.0 & 51.4 & 41.5 & 44.4 & 48.9 & 49.0 & 45.0 & 45.3 & 38.9 & 46.3\\
\model-7B (Ours) & 67.2 & 72.9 & 60.1 & 68.8 & 67.2 & 64.7 & 61.6 & 59.1 & 60.7 & 56.0\\
\midrule
\textbf{Models} & \textbf{fr-br} & \textbf{in-bn} & \textbf{in-ta} & \textbf{in-te} & \textbf{ind-id} & \textbf{ind-jv} & \textbf{ind-mi} & \textbf{ind-sv} & \textbf{ir-ir} & \textbf{ja-jp} \\
\midrule
Llava-1.5-7B & 37.3 & 52.1 & 61.4 & 63.5 & 47.8 & 50.8 & 49.0 & 44.0 & 61.3 & 41.9\\
Llava-Next-7B & 37.5 & 60.8 & 61.4 & 60.5 & 48.5 & 48.1 & 51.4 & 49.0 & 66.6 & 40.9\\
Phi-3.5-Vision & 41.7 & 58.7 & 60.5 & 60.0 & 51.7 & 45.5 & 51.4 & 47.5 & 62.6 & 41.4\\
Cambrian-8B & 40.7 & 68.5 & 65.6 & 63.0 & 55.1 & 50.2 & 58.2 & 56.0 & 66.6 & 42.4\\
Llava-OV-7B & 44.2 & 69.6 & 72.0 & 70.5 & 59.0 & 55.9 & 59.4 & 58.5 & 76.4 & 47.3 \\
Molmo-7B-D & 29.6 & 47.9 & 36.4 & 41.5 & 50.5 & 45.1 & 43.4 & 39.5 & 43.6 & 44.8\\
Llama3.2-11B & 36.3 & 62.9 & 66.4 & 66.5 & 63.6 & 48.8 & 58.2 & 54.0 & 57.4 & 58.1\\
PaliGemma-3B & 37.3 & 59.1 & 66.0 & 62.5 & 49.3 & 48.1 & 43.4 & 46.0 & 58.3 & 44.8\\
PALO-7B & 36.8 & 52.4 & 53.5 & 56.5 & 45.1 & 45.8 & 44.2 & 42.0 & 55.6 & 37.4\\
mBLIP mT0-XL & 30.4 & 43.0 & 46.0 & 41.0 & 38.1 & 39.1 & 38.6 & 32.5 & 37.4 & 34.0\\
mBLIP BLOOMZ-7B & 34.6 & 43.4 & 52.6 & 49.5 & 41.0 & 44.8 & 38.2 & 30.5 & 42.3 & 36.5\\
\model-7B (Ours) & 45.2 & 67.1 & 71.0 & 68.0 & 60.4 & 57.2 & 56.9 & 56.0 & 72.7 & 45.8\\
\midrule
\textbf{Models} & \textbf{ke-sw} & \textbf{ma-my} & \textbf{me-es} & \textbf{mo-mg} & \textbf{ni-ig} & \textbf{no-ng} & \textbf{pk-ur} & \textbf{ph-fi} & \textbf{ro-ro} & \textbf{ru-ru} \\
\midrule
Llava-1.5-7B & 68.9 & 52.1 & 47.9 & 45.8 & 51.0 & 58.5 & 63.9 & 52.7 & 55.6 & 59.0\\
Llava-Next-7B & 71.1 & 54.9 & 51.1 & 44.2 & 53.0 & 57.2 & 67.1 & 56.7 & 62.6 & 58.5\\
Phi-3.5-Vision & 72.9 & 57.1 & 46.3 & 50.7 & 53.0 & 56.2 & 60.6 & 57.6 & 61.9 & 58.5\\
Cambrian-8B & 74.4 & 61.9 & 56.7 & 48.7 & 56.5 & 60.5 & 73.1 & 60.1 & 66.6 & 61.5\\
Llava-OV-7B & 79.1 & 65.1 & 63.2 & 52.6 & 57.5 & 64.2 & 75.0 & 64.0 & 72.5 & 72.5 \\
Molmo-7B-D & 47.6 & 51.7 & 55.1 & 35.9 & 36.0 & 49.2 & 46.8 & 43.3 & 52.0 & 63.5\\
Llama3.2-11B & 61.5 & 69.2 & 64.7 & 41.0 & 39.5 & 65.9 & 65.7 & 66.0 & 75.5 & 74.5\\
PaliGemma-3B & 59.7 & 54.9 & 51.7 & 43.4 & 46.0 & 55.2 & 67.6 & 48.8 & 60.9 & 56.0\\
PALO-7B & 65.9 & 49.2 & 53.4 & 42.9 & 49.0 & 54.5 & 60.6 & 52.7 & 55.0 & 53.5\\
mBLIP mT0-XL & 50.2 & 41.6 & 34.7 & 33.9 & 39.5 & 43.1 & 45.4 & 36.9 & 43.7 & 41.0\\
mBLIP BLOOMZ-7B & 54.6 & 45.7 & 39.3 & 38.1 & 45.0 & 47.2 & 60.6 & 36.9 & 50.3 & 44.0\\
\model-7B (Ours) & 77.2 & 62.5 & 61.6 & 52.9 & 59.5 & 64.9 & 72.2 & 64.0 & 71.9 & 68.5\\
\midrule
\textbf{Models} & \textbf{rw-ki} & \textbf{sg-zh} & \textbf{sk-ko} & \textbf{sp-es} & \textbf{sr-si} & \textbf{ur-es} & \textbf{macro} & & & \\
\midrule
Llava-1.5-7B & 51.1 & 60.8 & 56.9 & 66.0 & 58.7 & 42.5 & 54.2 & & & \\
Llava-Next-7B & 52.8 & 62.3 & 60.0 & 67.6 & 59.1 & 38.7 & 55.7 & & & \\
Phi-3.5-Vision & 52.3 & 59.4 & 66.5 & 66.7 & 61.3 & 46.3 & 56.3 & & & \\
Cambrian-8B & 56.2 & 66.0 & 63.1 & 71.7 & 63.1 & 47.0 & 59.7 & & & \\
Llava-OV-7B & 55.7 & 73.6 & 67.9 & 80.2 & 72.9 & 48.9 & 65.2\\
Molmo-7B-D & 34.9 & 66.0 & 56.9 & 66.7 & 31.6 & 44.8 & 48.3 & & & \\
Llama3.2-11B & 40.4 & 73.6 & 73.1 & 83.3 & 51.1 & 56.2 & 61.2 & & & \\
PaliGemma-3B & 44.7 & 59.4 & 58.3 & 61.0 & 62.2 & 40.6 & 52.9 & & & \\
PALO-7B & 51.9 & 56.1 & 55.9 & 62.9 & 54.2 & 42.2 & 50.9 & & & \\
mBLIP mT0-XL & 38.3 & 43.9 & 41.4 & 51.9 & 48.0 & 34.9 & 40.5 & & & \\
mBLIP BLOOMZ-7B & 45.1 & 53.8 & 46.9 & 58.5 & 46.7 & 34.0 & 44.9 & & & \\
\model-7B (Ours) & 56.6 & 71.7 & 66.6 & 75.2 & 70.6 & 52.7 & 64.4 & & & \\
\bottomrule
\end{tabular}%
}
\caption{Comparison of models on CVQA across different country-language pairs (in English). Includes Macro-Acc.}
\label{tab:cvqa_en}
\end{table}

\subsection{MaRVL}
We show the performance of different models on the MaRVL benchmark in \autoref{tab:breakdown:MaRVL}.
\begin{table}[!h]
\centering
\small
\begin{tabular}{@{}l|cc|ccccc@{}}
\toprule
\textbf{Models} & \textbf{English} & \textbf{Multi} & \textbf{Indonesian} & \textbf{Swahili} & \textbf{Tamil} & \textbf{Turkish} & \textbf{Chinese} \\ \midrule
GPT4o & 81.8 & 82.3 & 81.9 & 80.8 & 80.2 & 86.4 & 82.1 \\
Gemini-1.5-Pro & 76.4 & 72.0 & 71.2 & 67.8 & 70.0 & 75.4 & 75.8 \\ \midrule
Llava-1.5-7B & 56.2 & 53.7 & 56.1 & 49.8 & 49.7 & 55.4 & 57.5 \\
Llava-Next-7B & 62.8 & 50.9 & 52.2 & 50.6 & 50.5 & 50.4 & 50.6 \\
Phi-3.5-Vision & 72.1 & 56.5 & 58.6 & 51.4 & 52.0 & 58.6 & 61.7 \\
Cambrian-8B & 75.4 & 61.8 & 64.7 & 53.6 & 56.7 & 65.2 & 68.9 \\ 
Llava-OV-7B & 72.7 & 57.5 & 60.9 & 51.2 & 51.9 & 63.5 & 60.0 \\
Molmo-7B-D & 65.3 & 54.9 & 61.1 & 49.6 & 49.6 & 52.2 & 62.2 \\
Llama3.2-11B & 64.5 & 58.1 & 62.7 & 52.4 & 54.0 & 61.6 & 59.5 \\
PaliGemma-3b & 56.5 & 52.2 & 53.4 & 49.6 & 50.5 & 56.3 & 51.3 \\
PALO-7B & 63.3 & 54.2 & 58.3 & 50.6 & 51.9 & 54.9 & 55.3 \\
mBLIP mT0-XL & 67.3 & 66.7 & 64.9 & 64.8 & 69.7 & 68.1 & 65.9 \\
mBLIP BLOOMZ-7B & 62.3 & 58.6 & 59.1 & 56.2 & 60.3 & 57.7 & 59.7 \\ \midrule
\model-7B & 87.0 & 79.0 & 81.3 & 75.1 & 69.4 & 84.8 & 84.3 \\ \bottomrule
\end{tabular}
\caption{Comparison of models on the MaRVL dataset across different languages.}
\label{tab:breakdown:MaRVL}
\end{table}

\subsection{XM100}
We show the performance of different models on the XM100 benchmark in \autoref{tab:breakdown_XM100}.

\begin{table}[!h]
\centering
\resizebox{0.95\textwidth}{!}{%
\begin{tabular}{l|cccccccc}
\toprule
\textbf{Models} & \textbf{English} & \textbf{Multi} & \textbf{Arabic} & \textbf{Bengali} & \textbf{Czech} & \textbf{Danish} & \textbf{German} & \textbf{Greek} \\
\midrule
Gemini-1.5-Pro & 27.6 & 19.1 & 1.7 & 7.5 & 25.9 & 32.8 & 27.6 & 5.0 \\
GPT4o & 27.7 & 19.1 & 15.8 & 13.5 & 21.1 & 25.3 & 19.3 & 21.1 \\
Llava-1.5-7B & 28.6 & 1.1 & 0.0 & 0.0 & 2.1 & 1.0 & 3.1 & 0.0 \\
Llava-Next-7B & 29.3 & 9.4 & 5.6 & 0.1 & 12.1 & 15.7 & 14.4 & 4.2 \\
Phi-3.5-Vision & 30.2 & 5.2 & 0.4 & 2.4 & 16.6 & 16.2 & 0.0 & 20.7 \\
Cambrian-8B & 20.6 & 9.9 & 1.4 & 6.6 & 7.4 & 15.1 & 15.5 & 4.4 \\
Llava-OV-7B & 30.6 & 7.0 & 0.2 & 0.6 & 5.2 & 16.8 & 14.0 & 0.4 \\
Molmo-7B-D & 22.1 & 9.1 & 5.4 & 7.9 & 5.7 & 13.8 & 12.2 & 4.2 \\
Llama3.2-11B & 27.6 & 4.5 & 0.0 & 0.0 & 1.5 & 11.8 & 4.6 & 1.2 \\
PaliGemma-3B & 18.7 & 0.8 & 0.0 & 0.0 & 1.1 & 3.1 & 2.7 & 0.0 \\
PALO-7B & 30.4 & 0.8 & 0.0 & 0.0 & 2.0 & 1.0 & 2.7 & 0.0 \\
mBLIP mT0-XL & 31.9 & 3.1 & 3.2 & 1.6 & 3.7 & 2.1 & 2.9 & 3.1 \\
mBLIP BLOOMZ & 22.5 & 10.3 & 9.5 & 6.4 & 11.5 & 15.9 & 14.5 & 10.9 \\
\model-7B (Ours) & 30.4 & 14.2 & 18.1 & 16.4 & 16.2 & 20.7 & 20.6 & 11.2 \\
\midrule
\textbf{Models} & \textbf{Spanish} & \textbf{Persian} & \textbf{Finnish} & \textbf{Filipino} & \textbf{French} & \textbf{Hebrew} & \textbf{Hindi} & \textbf{Croatian} \\
\midrule
Gemini-1.5-Pro & 39.5 & 4.2 & 29.0 & 28.7 & 42.4 & 4.3 & 2.2 & 33.8 \\
GPT4o & 28.3 & 26.6 & 13.1 & 26.4 & 23.1 & 20.4 & 17.0 & 19.4 \\
Llava-1.5-7B & 3.7 & 0.0 & 0.4 & 1.1 & 2.0 & 0.1 & 0.0 & 0.3 \\
Llava-Next-7B & 23.6 & 9.4 & 5.5 & 9.3 & 23.0 & 2.7 & 10.2 & 7.5 \\
Phi-3.5-Vision & 20.7 & 0.0 & 1.0 & 1.7 & 21.2 & 0.3 & 0.0 & 0.5 \\
Cambrian-8B & 18.6 & 9.6 & 5.1 & 19.6 & 18.3 & 5.8 & 6.8 & 7.2 \\
Llava-OV-7B & 24.9 & 3.8 & 1.5 & 4.2 & 22.0 & 0.0 & 4.4 & 7.2 \\
Molmo-7B-D & 19.8 & 11.3 & 3.1 & 13.0 & 19.8 & 8.3 & 9.4 & 6.9 \\
Llama3.2-11B & 10.2 & 0.0 & 2.4 & 8.4 & 12.0 & 0.0 & 0.2 & 0.7 \\
PaliGemma-3B & 0.7 & 0.0 & 0.1 & 0.1 & 0.6 & 0.0 & 0.0 & 1.3 \\
PALO-7B & 1.5 & 0.0 & 0.4 & 0.9 & 2.1 & 0.0 & 0.0 & 0.2 \\
mBLIP mT0-XL & 8.3 & 5.5 & 1.7 & 2.8 & 6.4 & 4.0 & 1.8 & 0.9 \\
mBLIP BLOOMZ & 18.9 & 13.8 & 4.8 & 7.7 & 19.1 & 7.5 & 10.1 & 3.2 \\
\model-7B (Ours) & 26.2 & 19.3 & 3.8 & 18.9 & 26.7 & 18.2 & 17.4 & 10.8 \\
\midrule
\textbf{Models} & \textbf{Hungarian} & \textbf{Indonesian} & \textbf{Italian} & \textbf{Japanese} & \textbf{Korean} & \textbf{Maori} & \textbf{Dutch} & \textbf{Norwegian} \\
\midrule
Gemini-1.5-Pro & 37.2 & 55.4 & 27.6 & 1.2 & 8.2 & 3.8 & 27.7 & 36.7 \\
GPT4o & 21.8 & 28.4 & 21.0 & 0.0 & 11.1 & 26.8 & 26.4 & 24.7 \\
Llava-1.5-7B & 3.3 & 0.9 & 4.3 & 0.0 & 0.0 & 0.2 & 2.9 & 3.7 \\
Llava-Next-7B & 9.3 & 14.7 & 17.6 & 4.2 & 5.2 & 9.2 & 23.8 & 16.3 \\
Phi-3.5-Vision & 3.4 & 3.2 & 17.5 & 1.6 & 0.3 & 0.2 & 17.2 & 14.1 \\
Cambrian-8B & 6.6 & 15.7 & 15.5 & 7.2 & 2.0 & 3.2 & 20.3 & 16.0 \\
Llava-OV-7B & 3.6 & 16.4 & 12.8 & 0.6 & 0.0 & 1.7 & 24.7 & 13.9 \\
Molmo-7B-D & 3.5 & 17.2 & 17.8 & 5.2 & 2.4 & 7.5 & 15.7 & 13.8 \\
Llama3.2-11B & 12.7 & 1.2 & 16.0 & 0.0 & 0.0 & 9.3 & 22.0 & 1.1 \\
PaliGemma-3B & 2.0 & 0.2 & 1.8 & 0.0 & 0.0 & 4.0 & 2.6 & 2.3 \\
PALO-7B & 3.4 & 1.1 & 3.2 & 0.0 & 0.0 & 0.1 & 3.5 & 0.7 \\
mBLIP mT0-XL & 2.8 & 6.0 & 2.8 & 0.3 & 2.1 & 1.5 & 3.4 & 3.1 \\
mBLIP BLOOMZ & 11.8 & 16.0 & 16.5 & 0.0 & 4.5 & 0.1 & 18.2 & 14.5 \\
\model-7B (Ours) & 7.7 & 27.9 & 22.9 & 2.1 & 8.1 & 0.7 & 26.6 & 24.9 \\
\midrule
\textbf{Models} & \textbf{Polish} & \textbf{Portuguese} & \textbf{Quechua} & \textbf{Romanian} & \textbf{Russian} & \textbf{Swedish} & \textbf{Swahili} & \textbf{Telugu} \\
\midrule
Gemini-1.5-Pro & 35.5 & 35.7 & 0.7 & 31.2 & 32.4 & 37.8 & 10.7 & 0.0 \\
GPT4o & 22.2 & 28.0 & 4.4 & 19.1 & 20.7 & 26.0 & 20.0 & 12.5 \\
Llava-1.5-7B & 0.8 & 2.5 & 0.0 & 1.6 & 0.5 & 2.0 & 0.1 & 0.0 \\
Llava-Next-7B & 13.5 & 21.3 & 0.0 & 11.5 & 13.5 & 16.0 & 3.2 & 0.0 \\
Phi-3.5-Vision & 1.0 & 21.0 & 0.4 & 3.2 & 0.7 & 12.5 & 0.4 & 0.0 \\
Cambrian-8B & 9.3 & 17.5 & 0.0 & 13.4 & 11.3 & 17.9 & 3.7 & 2.3 \\
Llava-OV-7B & 7.4 & 24.6 & 0.0 & 6.8 & 5.5 & 15.0 & 2.0 & 0.0 \\
Molmo-7B-D & 8.2 & 16.2 & 0.6 & 11.6 & 12.3 & 14.1 & 3.8 & 0.4 \\
Llama3.2-11B & 1.0 & 18.6 & 0.0 & 10.1 & 0.6 & 7.4 & 5.8 & 0.0 \\
PaliGemma-3B & 0.9 & 1.3 & 0.1 & 0.8 & 0.0 & 2.0 & 0.0 & 0.0 \\
PALO-7B & 0.8 & 1.7 & 0.0 & 1.1 & 0.5 & 0.9 & 0.2 & 0.0 \\
mBLIP mT0-XL & 3.5 & 5.8 & 0.2 & 2.3 & 3.1 & 3.7 & 3.8 & 2.7 \\
mBLIP BLOOMZ & 11.8 & 16.5 & 0.1 & 13.7 & 14.5 & 14.5 & 8.4 & 3.0 \\
\model-7B (Ours) & 16.2 & 28.1 & 0.0 & 21.4 & 20.9 & 19.4 & 18.7 & 0.1 \\
\midrule
\midrule
\textbf{Models} & \textbf{Thai} & \textbf{Turkish} & \textbf{Ukrainian} & \textbf{Vietnamese} & \textbf{Chinese} & \textbf{} & \textbf{} & \textbf{} \\
\midrule
Gemini-1.5-Pro & 0.0 & 0.9 & 0.0 & 0.0 & 0.9 & & & \\
GPT4o & 0.0 & 17.6 & 16.9 & 30.9 & 0.4 & & & \\
Llava-1.5-7B & 0.0 & 0.0 & 0.0 & 0.0 & 0.0 & & & \\
Llava-Next-7B & 0.0 & 0.0 & 0.3 & 0.0 & 6.3 & & & \\
Phi-3.5-Vision & 0.5 & 1.9 & 0.0 & 2.2 & 0.0 & & & \\
Cambrian-8B & 0.4 & 9.3 & 5.9 & 17.8 & 11.3 & & & \\
Llava-OV-7B & 0.0 & 0.0 & 0.0 & 0.0 & 2.9 & & & \\
Molmo-7B-D & 0.0 & 0.0 & 0.0 & 0.0 & 0.0 & & & \\
Llama3.2-11B & 0.0 & 0.0 & 0.0 & 0.0 & 2.9 & & & \\
PaliGemma-3B & 0.5 & 0.0 & 0.0 & 0.2 & 0.0 & & & \\
PALO-7B & 0.2 & 0.0 & 0.0 & 0.1 & 0.0 & & & \\
mBLIP mT0-XL & 0.0 & 3.9 & 2.0 & 7.1 & 0.0 & & & \\
mBLIP BLOOMZ & 0.5 & 1.9 & 0.0 & 2.2 & 0.0 & & & \\
\model-7B (Ours) & 0.0 & 0.0 & 0.3 & 0.0 & 4.9 & & & \\
\bottomrule
\end{tabular}%
}
\caption{Comparison of models on the XM100 dataset across different languages.}
\label{tab:breakdown_XM100}
\end{table}

\subsection{xGQA}
We show the performance of different models on the xGQA benchmark in \autoref{tab:breakdown:xGQA}.

\begin{table}[htbp]
\centering
\resizebox{\textwidth}{!}{%
\begin{tabular}{l|cc|ccccccc}
\toprule
\textbf{Models} & \textbf{English} & \textbf{Multi} & \textbf{Bengali} & \textbf{German} & \textbf{Indonesian} & \textbf{Korean} & \textbf{Portuguese} & \textbf{Russian} & \textbf{Chinese} \\
\midrule
Gemini-1.5-Pro & 54.2 & 48.7 & 49.4 & 50.2 & 48.6 & 46.4 & 51.2 & 44.8 & 50.2 \\
GPT4o & 55.8 & 51.0 & 49.4 & 52.6 & 50.4 & 51.0 & 52.2 & 50.0 & 51.4 \\\midrule
Llava-1.5-7B & 62.0 & 30.7 & 15.6 & 28.4 & 33.4 & 38.2 & 27.5 & 33.1 & 38.4 \\
Llava-Next-7B & 64.8 & 37.8 & 11.5 & 41.5 & 37.3 & 42.5 & 39.8 & 43.5 & 48.2 \\
Phi-3.5-Vision & 64.7 & 38.4 & 7.7 & 51.4 & 36.0 & 36.3 & 49.6 & 46.2 & 41.4 \\
Cambrian-8B & 64.6 & 39.8 & 32.3 & 44.6 & 36.0 & 43.6 & 41.6 & 44.2 & 36.2 \\
Llava-OV-7B & 64.4 & 48.2 & 41.8 & 49.2 & 48.8 & 45.3 & 52.4 & 54.0 & 45.9 \\
Molmo-7B-D & 51.5 & 43.0 & 25.6 & 45.9 & 44.9 & 44.2 & 46.5 & 45.6 & 48.1 \\
Llama3.2-11B & 55.6 & 45.4 & 42.9 & 46.7 & 46.2 & 44.5 & 46.5 & 44.7 & 46.1 \\
PaliGemma-3B & 59.7 & 30.5 & 13.3 & 44.5 & 21.3 & 22.8 & 34.7 & 35.8 & 41.2 \\
PALO-7B & 60.5 & 37.8 & 42.2 & 39.1 & 36.8 & 41.7 & 31.7 & 27.0 & 46.5 \\
mBLIP mT0-XL & 44.2 & 39.9 & 39.1 & 41.1 & 39.1 & 39.7 & 40.7 & 40.2 & 39.4 \\
mBLIP BLOOMZ-7B & 43.3 & 36.9 & 37.7 & 36.3 & 39.3 & 28.5 & 40.7 & 36.6 & 39.1 \\ \midrule
\model-7B (Ours) & 64.7 & 60.2 & 58.9 & 61.6 & 60.1 & 58.9 & 61.8 & 60.4 & 59.6 \\
\bottomrule
\end{tabular}%
}
\caption{Comparison of models on the xGQA dataset across different languages}
\label{tab:breakdown:xGQA}
\end{table}

\subsection{MAXM}
We show the performance of different models on the MAXM benchmark in \autoref{tab:breakdown:MAXM}.
\begin{table}[!h]
\centering
\resizebox{\textwidth}{!}{%
\begin{tabular}{l|rr|rrrrrr}
\toprule
\textbf{Models} & \textbf{English} & \textbf{Multi} & \textbf{French} & \textbf{Hindi} & \textbf{Hebrew} & \textbf{Romanian} & \textbf{Thai} & \textbf{Chinese} \\
\midrule
Gemini-1.5-Pro & 56.4 & 63.5 & 60.2 & 66.5 & 65.7 & 57.4 & 73.9 & 57.4 \\
GPT4o & 60.7 & 65.4 & 59.8 & 68.8 & 70.0 & 61.3 & 76.5 & 56.3 \\\midrule
Llava-1.5-7B & 49.8 & 20.4 & 32.2 & 17.3 & 12.9 & 15.1 & 17.2 & 27.8 \\
Llava-Next-7B & 54.9 & 21.4 & 33.7 & 16.2 & 10.7 & 15.5 & 18.3 & 33.9 \\
Phi-3.5-Vision & 55.3 & 25.0 & 38.3 & 31.9 & 17.5 & 10.9 & 24.3 & 27.4 \\
Cambrian-8B & 55.3 & 28.7 & 41.7 & 23.8 & 17.1 & 32.0 & 25.7 & 31.8 \\
Llava-OV-7B & 54.9 & 34.8 & 37.9 & 31.9 & 17.8	& 30.2 & 53.0 & 37.9 \\
Molmo-7B-D & 52.9 & 37.5	& 45.5	& 33.5	& 30.7	& 28.9	& 46.3	& 40.4 \\
Llama3.2-11B & 55.3 &43.9 & 48.1 & 50.4 & 41.8	& 36.6 & 56.7 &30.0 \\
PaliGemma-3B & 47.9 & 19.9 & 8.0 & 36.5 & 19.3 & 13.4 & 31.3 & 10.8 \\
PALO-7B & 51.4 & 16.3 & 33.7 & 15.8 & 12.1 & 11.3 & 14.6 & 10.5 \\
mBLIP mT0-XL & 44.7 & 36.8 & 36.0 & 42.7 & 28.9 & 30.3 & 56.3 & 26.4 \\
mBLIP BLOOMZ-7B & 44.7 & 24.8 & 33.0 & 47.3 & 8.9 & 16.9 & 9.7 & 33.2 \\\midrule
\model-7B (Ours) & 55.3 & 53.3 & 43.6 & 53.5 & 59.3 & 45.8 & 67.2 & 50.2 \\
\bottomrule
\end{tabular}%
}
\caption{Comparison of models on the MAXM dataset across different languages.}
\label{tab:breakdown:MAXM}
\end{table}

\subsection{xMMMU}
We show the performance of different models on the xMMMU benchmark in \autoref{tab:breakdown:xMMMU}.

\begin{table}[!h]
\centering
\resizebox{\textwidth}{!}{%
\begin{tabular}{l|cc|cccccc}
\toprule
\textbf{Models} & \textbf{English} & \textbf{Multi} & \textbf{Arabic} & \textbf{French} & \textbf{Hindi} & \textbf{Indonesian} & \textbf{Japanese} & \textbf{Portuguese} \\
\midrule
Gemini-1.5-Pro (0801) & 65.8 & 57.7 & 57.7 & 58.1 & 55.5 & 60.2 & 55.0 & 59.6 \\
GPT4o (0513) & 69.1 & 58.3 & 56.7 & 58.1 & 58.1 & 59.9 & 58.0 & 58.9 \\\midrule
Llava-1.5-7B & 36.2 & 31.5 & 29.5 & 34.9 & 27.5 & 31.6 & 32.0 & 33.7 \\
Llava-Next-7B & 36.7 & 34.3 & 30.5 & 35.6 & 30.9 & 37.0 & 34.9 & 37.0 \\
Phi-3.5-Vision & 42.6 & 38.8 & 35.6 & 44.0 & 30.9 & 36.7 & 37.9 & 47.8 \\
Cambrian-8B & 41.8 & 33.2 & 32.6 & 34.6 & 30.9 & 31.3 & 33.5 & 36.0 \\
Llava-OV-7B & 46.3 & 41.0 & 41.6 & 43.0 & 34.7 & 43.4 & 40.1 & 43.4 \\
Molmo-7B-D &  42.9 & 40.4 & 40.6 & 42.6 & 32.6 & 40.7 & 43.9 & 42.1 \\
Llama3.2-11B & 39.2 & 34.0 & 33.6 & 39.6 & 32.3 & 36.7 & 29.0 & 33.0 \\
PaliGemma-3B & 26.3 & 25.2 & 29.2 & 23.8 & 21.6 & 24.2 & 24.5 & 27.6 \\
PALO-7B & 33.1 & 30.5 & 30.5 & 33.2 & 28.9 & 34.0 & 27.1 & 33.3 \\
mBLIP mT0-XL & 29.3 & 30.4 & 30.2 & 33.2 & 28.2 & 26.9 & 31.6 & 32.3 \\
mBLIP BLOOMZ-7B & 29.2 & 30.8 & 28.5 & 33.9 & 27.8 & 33.3 & 31.6 & 29.6 \\\midrule
\model-7B (Ours) & 45.7 & 43.7 & 42.3 & 45.3 & 41.6 & 46.5& 40.5 & 46.1 \\
\bottomrule
\end{tabular}%
}
\caption{Comparison of models on the xMMMU dataset across different languages.}
\label{tab:breakdown:xMMMU}
\end{table}

\subsection{M3Exam}
We show the performance of different models on the M3Exam benchmark in \autoref{tab:breakdown:m3exam}.


\begin{table}[htbp]
\centering
\resizebox{\textwidth}{!}{%
\begin{tabular}{l|cc|cccccc}
\toprule
\textbf{Models} & \textbf{English} & \textbf{Multi} & \textbf{Afrikaans} & \textbf{Chinese} & \textbf{Italian} & \textbf{Portuguese} & \textbf{Thai} & \textbf{Vietnamese} \\
\midrule
Gemini-1.5-Pro & 77.4 & 64.7 & 80.4 & 74.1 & 76.3 & 61.8 & 49.9 & 46.0 \\
GPT4o & 68.0 & 61.0 & 73.0 & 68.0 & 67.0 & 58.0 & 52.0 & 48.3 \\\midrule
Llava-1.5-7B & 32.3 & 29.0 & 28.2 & 24.3 & 40.1 & 28.2 & 23.7 & 29.3 \\
Llava-Next-7B & 36.5 & 28.4 & 28.2 & 25.4 & 37.8 & 27.0 & 23.7 & 28.4 \\
Phi-3.5-Vision & 55.8 & 37.2 & 44.2 & 40.8 & 51.4 & 40.3 & 25.2 & 21.6 \\
Cambrian-8B & 34.7 & 33.4 & 36.8 & 34.2 & 45.2 & 30.3 & 28.9 & 25.0 \\
Llava-OV-7B & 60.4 & 45.8 & 50.3 & 58.0 & 57.2 & 43.8 & 30.9 & 34.5 \\
Molmo-7B-D & 57.1 & 39.1 & 35.6 & 56.4 & 49.4 & 40.2 & 27.4 & 25.9 \\
Llama3.2-11B & 51.8 & 36.6 & 42.3 & 46.4 & 45.8 & 28.4 & 26.4 & 30.2 \\
PaliGemma-3B & 36.0 & 25.6 & 26.4 & 24.7 & 32.2 & 24.3 & 27.2 & 19.0 \\
PALO-7B & 30.8 & 27.8 & 31.9 & 22.1 & 36.9 & 32.3 & 22.7 & 20.7 \\
mBLIP mT0-XL & 22.8 & 25.0 & 16.0 & 25.6 & 33.7 & 21.2 & 22.4 & 31.0 \\
mBLIP BLOOMZ-7B & 30.3 & 29.5 & 28.2 & 29.8 & 37.3 & 28.3 & 22.9 & 30.2 \\\midrule
\model-7B (Ours) & 61.4 & 42.1 & 52.1& 49.2 & 54.9 & 43.3 & 32.9 & 19.8 \\
\bottomrule
\end{tabular}%
}
\caption{Comparison of models on the M3Exam dataset across different languages.}
\label{tab:breakdown:m3exam}
\end{table}

\subsection{TyDiQA}

We show the performance of different models on the TyDiQA benchmark in \autoref{tab:breakdown:TyDiQA}.

\begin{table}[!h]
\centering
\resizebox{\textwidth}{!}{%
\begin{tabular}{l|cc|rrrrrrrrrr}
\toprule
\textbf{Models} & \textbf{English} & \textbf{Multi} & \textbf{Arabic} & \textbf{Bengali} & \textbf{Finnish} & \textbf{Indonesian} & \textbf{Korean} & \textbf{Russian} & \textbf{Swahili} & \textbf{Telugu} \\
\midrule
Vicuna-1.5-7B       &59.7 &52.7 &32.3 &68.1 &63.0 &72.6 &58.8 &57.6 &51.3 &18.1 \\
Qwen2-7B-Instruct   &72.2 &71.2 &67.6 &75.9 &67.1 &78.0 &64.9 &67.2 &75.3 &73.8 \\
\midrule
Llava-1.5-7B &66.8 &52.8 &61.8 &33.4 &60.2 &72.8 &63.3 &55.0 &55.0 &20.6 \\
Llava-Next-7B &68.3 &52.1 &64.5 &24.9 &63.0 &74.3 &61.9 &58.4 &53.1 &17.0 \\
Phi-3.5-Vision      &75.9 &51.3 &63.1 &24.8 &57.3 &70.6 &60.2 &57.5 &48.7 &28.3 \\
PALO-7B             &69.4 &50.8 &60.9 &46.0 &61.8 &70.6 &56.8 &56.7 &42.5 &10.8 \\
\midrule
\model-7B (Ours)    &73.7 &66.0 &55.5 &65.3 &66.3 &74.5 &69.4 &60.1 &76.6 &60.0 \\
\bottomrule
\end{tabular}%
}
\caption{Comparison of models on the TyDiQA dataset across different languages.}
\label{tab:breakdown:TyDiQA}
\end{table}

\subsection{XStoryCloze}
We show the performance of different models on the XStoryCloze benchmark in \autoref{tab:breakdown:XStoryCloze}.

\begin{table}[!h]
\centering
\resizebox{\textwidth}{!}{%
\begin{tabular}{l|rr|rrrrrrrrrr}
\toprule
\textbf{Models} & \textbf{English} & \textbf{Multi} & \textbf{Arabic} & \textbf{Spanish} & \textbf{Basque} & \textbf{Hindi} & \textbf{Ind.} & \textbf{Burmese} & \textbf{Russian} & \textbf{Swahili} & \textbf{Telugu} & \textbf{Chinese} \\
\midrule
Vicuna-1.5-7B & 78.1 & 57.4 & 52.7 & 69.4 & 50.8 & 54.5 & 61.0 & 48.4 & 66.5 & 52.1 & 54.5 & 63.5 \\
Qwen2-7B-Instruct & 80.3 & 61.9 & 64.0 & 71.6 & 51.6 & 59.6 & 68.5 & 50.7 & 72.7 & 53.2 & 55.3 & 72.1 \\\midrule
Llava-1.5-7B & 79.1 & 57.6 & 52.7 & 69.2 & 50.9 & 54.9 & 62.6 & 49.0 & 65.9 & 51.7 & 55.8 & 63.9 \\
Llava-Next-7B & 79.1 & 57.1 & 51.7 & 68.8 & 50.3 & 54.5 & 62.0 & 46.7 & 65.5 & 52.1 & 55.2 & 63.8 \\
Phi-3.5-Vision & 77.9 & 54.8 & 53.7 & 67.2 & 50.4 & 54.9 & 51.7 & 47.8 & 61.3 & 49.3 & 52.5 & 59.5 \\
PALO-7B & 77.4 & 57.2 & 56.5 & 68.4 & 49.8 & 58.6 & 58.5 & 47.4 & 65.6 & 51.2 & 53.1 & 62.8 \\\midrule
\model-7B (Ours) & 79.1 & 61.2 & 60.5 & 67.8 & 50.0 & 61.8 & 66.4 & 48.7 & 69.4 & 58.9 & 60.4 & 68.2 \\
\bottomrule
\end{tabular}%
}
\caption{Comparison of models on the XStoryCloze dataset across different languages.}
\label{tab:breakdown:XStoryCloze}
\end{table}

\subsection{MGSM}


We show the performance of different models on the MGSM benchmark in \autoref{tab:breakdown:MGSM}.
\begin{table}[!h]
\centering
\resizebox{\textwidth}{!}{%
\begin{tabular}{l|rr|rrrrrrrrrr}
\toprule
\textbf{Models} & \textbf{English} & \textbf{Multi} & \textbf{Bengali} & \textbf{German} & \textbf{Spanish} & \textbf{French} & \textbf{Japanese} & \textbf{Russian} & \textbf{Swahili} & \textbf{Telugu} & \textbf{Thai} & \textbf{Chinese} \\
\midrule
Vicuna-1.5-7B       &17.6  &6.4 &0.0 &14.4 &9.6  &14.4 &2.8  &10.8 &3.6  &0.0   &2.0    &14.8 \\
Qwen2-7B-Instruct   &48.8 &40.4 &0.0 &67.2 &67.6 &68.8 &11.2 &71.2 &10.8 &2.4 &45.6 &59.2 \\\midrule
Llava-1.5-7B &14.8 & 7.6 & 0.0 & 15.2 & 10.8 & 18.0 & 2.8 & 11.2 & 0.4 & 0.0 & 1.6 & 15.6 \\
Llava-Next-7B &15.6 & 7.5 & 0.0 & 13.6 & 13.2 & 16.0 & 1.6 & 12.8 & 2.0 & 0.0 & 1.6 & 14.0 \\
Phi-3.5-Vision      &59.2 & 33.1 & 0.0 & 64.0 & 59.6 & 58.0 & 20.0 & 54.0 & 4.0 & 0.0 & 18.8 & 52.4 \\
PALO-7B             &13.6 & 5.8 & 0.0 & 11.6 & 9.6 & 13.2 & 1.6 & 8.8 & 0.4 & 0.0 & 0.0 & 12.4 \\\midrule
\model-7B (Ours)    & 82.0 & 47.4 &0.0 &68.4 &74.8 &63.2 &22.0 &68.0 &54.0 &5.6 &49.6 &68.0 \\
\bottomrule
\end{tabular}%
}
\caption{Comparison of models on the MGSM dataset across different languages.}
\label{tab:breakdown:MGSM}
\end{table}

\subsection{MMMLU}
We show the performance of different models on the MMMLU benchmark in \autoref{tab:breakdown:xMMLU}.
\begin{table}[!h]
\centering
\resizebox{\textwidth}{!}{%
\begin{tabular}{l|cccccccc}
\toprule

\textbf{Models} & \textbf{English} & \textbf{Multi} & \textbf{Arabic} & \textbf{Bengali} & \textbf{ Portuguese} & \textbf{Chinese} & \textbf{French} & \textbf{German} \\
\midrule
Vicuna-1.5-7B       &49.5 &34.7 &30.3 &28.5 &39.6 &36.9 &40.4 &39.8 \\
Qwen2-7B-Instruct   &70.1 &53.1 &51.0 &43.4 &60.7 &63.8 &61.5 &57.7 \\
\midrule
Llava-1.5-7B &50.2 &34.9 &29.7 &28.5 &40.3 &36.8 &40.1 &39.8 \\
Llava-Next-7B &52.1 &35.6 &30.0 &28.8 &40.7 &37.3 &41.4 &41.4 \\
Phi-3.5-Vision      &62.0 &39.1 &34.9 &27.9 &47.6 &41.5 &49.2 &45.8 \\
PALO-7B             &46.7 &32.6 &30.3 &29.5 &36.0 &34.2 &36.9 &35.8 \\
\midrule
\model-7B (Ours)    &68.4 &52.2 &49.3 &44.4 &58.9 &60.5 &58.9 &56.7  \\
\midrule
\midrule
\textbf{Models} & \textbf{Hindi} & \textbf{Indonesian} & \textbf{Italian} & \textbf{Japanese} & \textbf{Korean} & \textbf{Spanish} & \textbf{Swahili} & \textbf{Yoruba} \\\midrule
Vicuna-1.5-7B       &29.8 &36.5 &39.5 &35.9 &34.1 &40.3 &27.9 &26.8 \\
Qwen2-7B-Instruct   &45.7 &57.1 &60.8 &58.0 &54.6 &61.9 &36.0 &31.8 \\
\midrule
Llava-1.5-7B &29.2 &37.1 &41.0 &35.1 &34.1 &41.6 &28.0 &27.3 \\
Llava-Next-7B &29.6 &37.5 &41.2 &36.0 &34.2 &42.7 &28.5 &28.7 \\
Phi-3.5-Vision      &32.9 &38.3 &47.0 &40.0 &36.6 &49.6 &28.9 &27.8 \\
PALO-7B             &29.6 &33.7 &36.4 &32.7 &30.6 &37.0 &26.4 &27.1 \\
\midrule
\model-7B (Ours)    &45.7 &55.4 &58.8 &55.3 &52.7 &59.7 &42.8 &31.3 \\
\bottomrule
\end{tabular}%
}
\caption{Comparison of models on the MMMLU dataset across different languages.}
\label{tab:breakdown:xMMLU}
\end{table}

\newpage
\section{A Preliminary Exploration of Constructing Multilingual OCR Instructions}
\label{sec:multingual_ocr_details}

Optical Character Recognition (OCR) is a critical capability for multimodal LLMs, enabling them to interpret and process textual information embedded within images. However, most existing OCR training datasets are predominantly English-centric, which limits the models' performance in non-English contexts. To address this gap, we have curated a comprehensive set of 500K multilingual OCR training samples from web user interfaces, spanning 10 languages, with 50K examples per language, sourced from web user interfaces. Webpages naturally serve as image-rich environments containing abundant text, and by capturing screenshots of websites from various countries in different languages, we were able to gather a substantial number of OCR images. 


We utilize URLs from the CC-News-Multilingual\footnote{https://huggingface.co/datasets/intfloat/multilingual\_cc\_news}dataset~\citep{Hamborg2017} to obtain a diverse set of multilingual web pages. Using Playwright\footnote{https://github.com/microsoft/playwright}, we render each website and automatically capture screenshots under various device settings and resolutions to achieve a wide range of image dimensions and aspect ratios. Each screenshot includes a red bounding box that highlights a specific element targeted for OCR extraction. We focus on ten languages for this dataset: English, Chinese, Japanese, Korean, Indonesian, Hindi, Spanish, French, Portuguese, and Arabic. We totally have 1M samples (50K for each language). 

\begin{wrapfigure}{r}{0.4\linewidth}
    \centering
    \vspace{-10pt}
    \includegraphics[width=\linewidth]{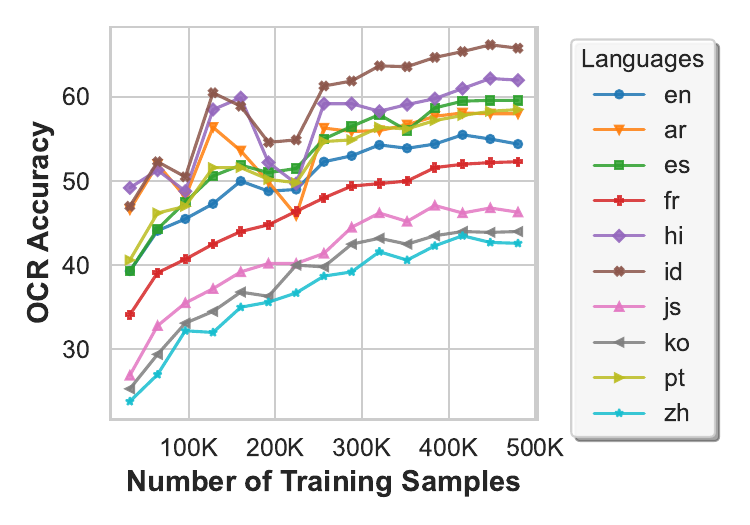}
    \vspace{-20pt}
    \caption{A preliminary exploration of multilingual OCR.}
    \vspace{-10pt}
    \label{fig:ocr_accuracy}
\end{wrapfigure}
We employed the same model architecture as \model but trained it exclusively on these OCR images, reserving a portion of the data as a test set. As shown in \autoref{fig:ocr_accuracy}
, the results indicate that improving multilingual OCR performance is feasible with an increase in training samples. However, the OCR accuracy for non-Latin scripts (e.g., Chinese, Japanese, and Korean) remains lower than for Latin-based languages. Looking ahead, we aim to further expand the multilingual OCR training dataset to include more languages and integrate this data into \traindata.

\end{document}